\documentclass[11pt]{article}

\usepackage[preprint]{acl}

\usepackage{times}
\usepackage{latexsym}

\usepackage[T1]{fontenc}

\usepackage[utf8]{inputenc}

\usepackage{microtype}

\usepackage{inconsolata}

\usepackage{graphicx}
\usepackage{amsmath}

\usepackage{graphicx}
\usepackage{enumerate}
\usepackage{enumitem}
\usepackage{wrapfig}
\usepackage{booktabs}
\usepackage{multirow}
\usepackage{pifont}
\usepackage{longtable}
\usepackage{caption}
\usepackage{subcaption}
\usepackage{tabularx}
\usepackage{booktabs}
\usepackage{caption}
\usepackage[export]{adjustbox}
\usepackage{stfloats}
\usepackage{balance}
\usepackage{flushend}

\definecolor{myred}{RGB}{255,0,0}
\definecolor{mygreen}{RGB}{0,128,0}
\definecolor{mygrey}{RGB}{128,128,128}

\title{V-MAGE: A Game Evaluation Framework for Assessing 
        Vision-Centric Capabilities in Multimodal Large Language Models}

\author{
 \textbf{Xiangxi Zheng\textsuperscript{1}},
 \textbf{Linjie Li\textsuperscript{2}},
 \textbf{Zhengyuan Yang\textsuperscript{2}},
 \textbf{Ping Yu\textsuperscript{1}},
\\
 \textbf{Alex Jinpeng Wang\textsuperscript{3}$^\dagger$},
 \textbf{Rui Yan\textsuperscript{4}$^\dagger$},
 \textbf{Yuan Yao\textsuperscript{1}}, 
 \textbf{Lijuan Wang\textsuperscript{2}}
\\
\\
 \textsuperscript{1}Nanjing University, 
 \textsuperscript{2}Microsoft Research, 
\\
 \textsuperscript{3}Central South University,
 \textsuperscript{4}Nanjing University of Science and Technology
 \\
 \texttt{zhengxx@nju.edu.cn, jinpengwang@csu.edu.cn, ruiyan@njust.edu.cn} 
}

\begin{document}
\maketitle

\renewcommand{\thefootnote}{\fnsymbol{footnote}}
\footnotetext[2]{Corresponding authors: Alex Jinpeng Wang, Rui Yan}

\begin{abstract}
\label{abstract}

Recent advancements in Multimodal Large Language Models (MLLMs) have demonstrated impressive capabilities in visual-text processing. 
However, existing static image-text benchmarks are insufficient for evaluating their dynamic perception and interactive reasoning abilities. 
We introduce \textbf{V}ision-centric \textbf{M}ultiple \textbf{A}bilities \textbf{G}ame \textbf{E}valuation (\textbf{V-MAGE}), a novel game-based evaluation framework designed to systematically assess MLLMs’ visual reasoning in interactive, continuous-space environments. 
V-MAGE features five distinct video games comprising over 30 carefully constructed evaluation scenarios.  
These scenarios are set in free-form, visually complex environments that require models to interpret dynamic game states and make decisions based solely on visual input, thereby closely reflecting the conditions encountered by human players.
To ensure robust and interpretable comparisons across models, V-MAGE employs a dynamic ELO-based ranking system that accounts for varying difficulty levels and task diversity.
Benchmarking state-of-the-art MLLMs against human baselines reveals that while leading models approach human-level performance in simple tasks, their performance drops significantly in complex scenarios requiring advanced reasoning and task orchestration. 
This persistent performance gap highlights fundamental limitations in current MLLMs' ability to perform vision-grounded, interactive frame-by-frame control in simulated continuous-time environments.
Through extensive analyses, we demonstrate the utility of V-MAGE in uncovering these limitations and providing actionable insights for improving the visual and reasoning capabilities of MLLMs in dynamic, interactive settings.
Code is publicly available at \url{https://github.com/CSU-JPG/V-MAGE}.

\end{abstract}

\section{Introduction}
\label{sec: introduction}


Building on the success of Large Language Models (LLMs) in text-based tasks~\cite{bai2023qwen, cai2024internlm2, openai2023gpt4}, researchers have extended their capabilities to visual-text multimodal tasks through Multimodal Large Language Models (MLLMs)~\cite{gpt4v,llava,geminiteam2023gemini,yang2023dawn,li2024multimodal, wang2024qwen2vlenhancingvisionlanguagemodels, Qwen2.5-VL, zhu2025internvl3exploringadvancedtraining}. Various multimodal evaluation benchmarks, such as MME~\cite{fu2023mme}, MMBench~\cite{liu2023mmbench}, SEED-Bench~\cite{li2023seedbench} have driven improvements in MLLM performance. 
With improving model capabilities, researchers are shifting toward open-world, dynamic, multi-round tasks beyond static benchmarks with fixed image-text inputs, as these better reflect real-world interaction and reasoning challenges. Among the promising approaches for evaluating models in such dynamic settings, game-based evaluation has emerged as a promising alternative, offering a more natural and interactive assessment of a model’s perception and reasoning abilities.

While progress has been made in game-based MLLM benchmarks, current approaches predominantly rely on text-based ~\cite{costarelli2024gamebenchevaluatingstrategicreasoning, hu2024gamearenaevaluatingllmreasoning, duan2024gtbenchuncoveringstrategicreasoning} or grid-based~\cite{zhang2024ingvpmllmsplayeasy, wang2025large, paglieri2024balrogbenchmarkingagenticllm} games. 
In such settings, limited visual reasoning demands and static, fully textually renderable content restrict evaluation of spatial, temporal, and dynamic complexities crucial for real-world problem-solving.
In contrast, the rich visual information inherent in video games presents a valuable opportunity to assess MLLMs' genuine visual reasoning capabilities, potentially addressing current methodological limits.

\begin{figure*}[h]
\vspace{-10pt}
    \centering
    \includegraphics[width = \textwidth]{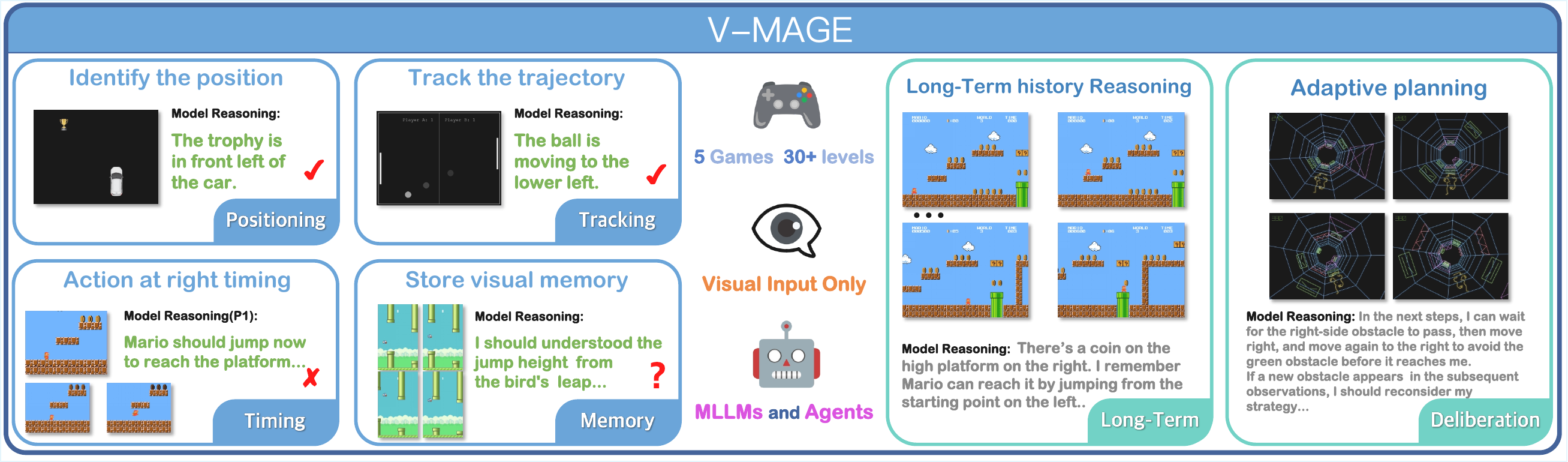}
    \fontsize{8pt}{8pt}\selectfont
\setlength{\tabcolsep}{2pt}
\centering
\renewcommand{\arraystretch}{1.15}

\caption{\small The overview of the V-MAGE benchmark, designed to evaluate vision-centric capabilities and higher-level reasoning of MLLMs across 5 free-form games with 30+ levels. V-MAGE assesses critical abilities in visual reasoning, providing a comprehensive evaluation of model performance in complex, dynamic environments. 
}
\label{fig:V-MAGE overview}
\vspace{-10pt}
\end{figure*}

To address the lack of vision-centric video game benchmarks, we present \textbf{V}ision-centric \textbf{M}ultiple \textbf{A}bilities \textbf{G}ame \textbf{E}valuation (\textbf{V-MAGE}), which allows a thorough assessment of diverse model and agent abilities within dynamic, interactive game environments and addresses key limitations in current game-based evaluations of MLLM capabilities. 


With V-MAGE, we evaluate leading MLLMs across five interactive games across 30+ levels. Results highlight significant challenges posed by the dynamic visual interaction environment for MLLMs. 
The results reveal that current MLLMs, despite excelling in static benchmarks, lack perception, multi-step reasoning, and task orchestration for human-level play in dynamic games.

Our contributions are summarized as follows:
\begin{itemize}[leftmargin=2mm,itemsep=0.1pt,topsep=0.1pt]
    \item We established V-MAGE, an interactive and visually rich evaluation framework focused on dynamic interaction and vision-centric reasoning. It also serves as a sandbox environment conducive to vision agent development.
    \item We evaluated various publicly available MLLMs with V-MAGE, measuring model performance with ELO scores and highlighting the significant gap between model performance and human-level proficiency on complex tasks.
    \item Through the evaluation results of V-MAGE, we further analyzed the reasons for the suboptimal performance of current MLLMs on video game tasks, including deficiencies in several fundamental visual capabilities, challenges in reasoning during prolonged interactions, and issues such as anchoring bias, among others.
\end{itemize}

\section{Related work}

\begin{figure*}[ht]
\vspace{-10pt}
\begin{center}
\centerline{\includegraphics[width=\textwidth]{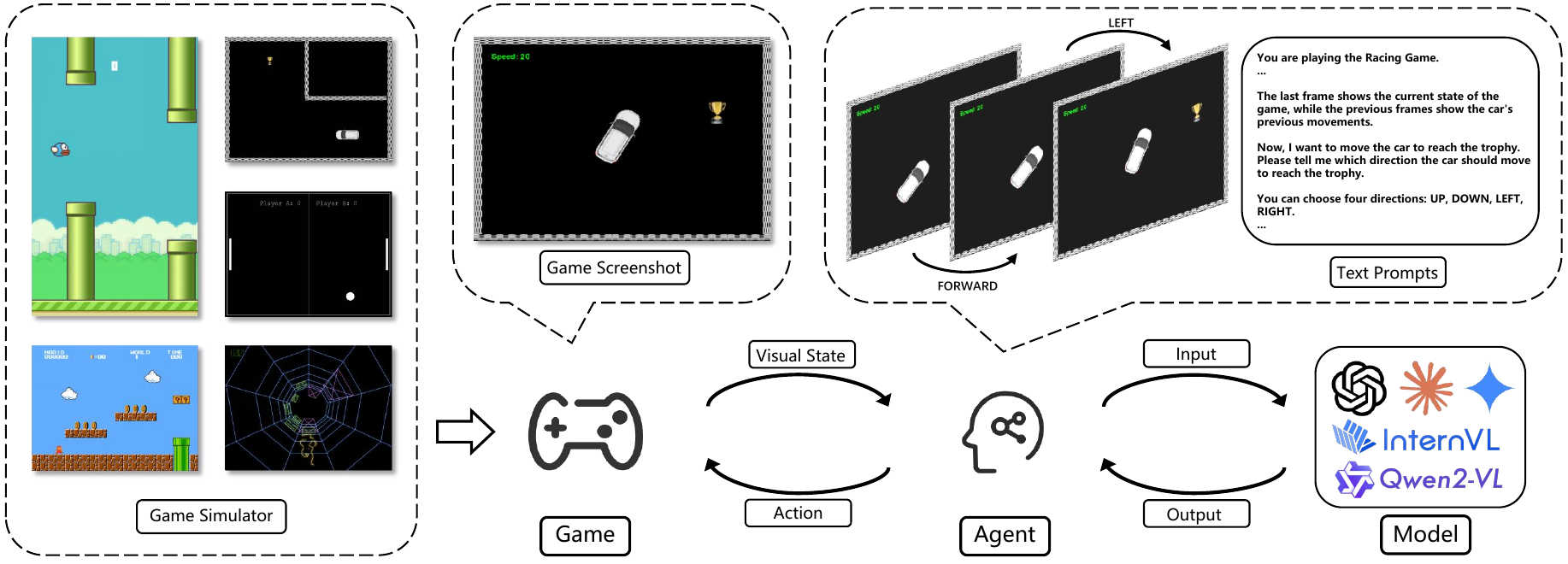}}
\caption{\small
V-MAGE games and evaluation pipeline.
V-MAGE employs five distinct games, each with several levels, to facilitate a decomposed evaluation of model performance.
These games include \textbf{FlappyBird}, \textbf{Race}, \textbf{SuperMario}, \textbf{Pong} and \textbf{TempestRun}.
During the evaluation process, the \textbf{Agent} module receives visual game state information directly from the \textbf{Game} module, primarily in the form of screenshots. The \textbf{Agent} module then structures these screenshots, combined with prompts containing the game rules, into the appropriate input format for MLLMs.
Subsequently, the model's output is processed by the \textbf{Agent} module to generate executable actions, which are then transmitted back to the \textbf{Game} module to update the environment state.
}
\label{fig:pipeline}
\end{center}
\vspace{-10pt}
\end{figure*}

\noindent
\textbf{MLLMs and Multimodal Agents.} As LLMs~\cite{qwen2025qwen25technicalreport, cai2024internlm2, openai2023gpt4} advance, MLLMs have emerged to handle multimodal tasks by integrating text and visual inputs~\cite{zhu2025internvl3exploringadvancedtraining,bai2025qwen25vltechnicalreport,chen2024internvl, llava}. Open-source models like InternVL and QwenVL are narrowing the gap ~\cite{chen2024far} with closed-source models such as GPT-4o~\cite{gpt4o}, and Gemini~\cite{geminiteam2023gemini}. 

MLLMs are evolving into interactive multimodal agents, finding applications in areas such as robotics~\cite{driess2023palme}, virtual assistants~\cite{rt22023arxiv, rt12022arxiv}, GUI automation~\cite{xu2024crab, bonatti2024windows, zhang2023appagentmultimodalagentssmartphone}, and game agents~\cite{tan2024cradle, chen2024vlmsplayactionroleplaying}.
These domains require sequential reasoning, memory, and adaptability, which static benchmarks inadequately capture.

\noindent
\textbf{MLLM Benchmarks.} 
Classic MLLM benchmarks have focused on tasks like Visual Question Answering (VQA)~\cite{antol2015vqa, vqav2, li2018vqa, okvqa} and image captioning~\cite{coco_captions, agrawal2019nocaps, sidorov2020textcaps}. More comprehensive benchmarks, such as MME~\cite{fu2023mme}, MMBench~\cite{liu2023mmbench}, SEED-Bench~\cite{li2023seedbench}, MMMU~\cite{yue2024mmmu}, and MM-Vet~\cite{yu2023mm,yu2024mm}, introduce broader assessments across multiple domains. 

Most of these evaluations rely on structured multiple-choice and VQA-style tasks, limiting their ability to measure real-world problem-solving and interactive reasoning. 
Recent multimodal agent benchmarks like OSWorld~\cite{OSWorld}, Windows Agent Arena~\cite{bonatti2024windows}, and COMMA~\cite{ossowski2024comma} assess broader capabilities: open-ended tasks in real environments, OS interaction, and multi-agent collaboration.

\noindent

\textbf{Evaluating MLLMs in Games.} 
Recent work~\cite{tan2024cradle,chen2024vlmsplayactionroleplaying,ruoss2024lmact} has explored MLLMs in interactive gaming environments. 
Existing game-based benchmarks span text-only settings~\cite{costarelli2024gamebenchevaluatingstrategicreasoning, hu2024gamearenaevaluatingllmreasoning, duan2024gtbenchuncoveringstrategicreasoning}, grid- or board-style visual settings~\cite{zhang2024ingvpmllmsplayeasy, wang2025large, paglieri2024balrogbenchmarkingagenticllm}, escape-room style multimodal reasoning environments~\cite{wang2025multimodallargelanguagemodels}, and GUI/adventure-game agents with long story arcs~\cite{ahn2025flashadventure}. 
These benchmarks are valuable and complementary, but they emphasize different capability profiles. In particular, many settings either remain text-dominant, expose substantial symbolic structure, or involve UI elements that can be mapped to discrete states. 
V-MAGE instead targets continuous-space, arcade-style environments in which key information lies in pixel-level dynamics such as motion, collision, relative position, and timing. 
This makes V-MAGE a complementary benchmark for evaluating vision-centric control and temporally grounded reasoning rather than a replacement for prior game benchmarks.

\section{V-MAGE Benchmark}

We present V-MAGE, a benchmark built on video game environments designed to evaluate the comprehensive performance of MLLMs, with a focus on vision-centric capabilities. 
Its defining features are as follows:
\begin{itemize}[leftmargin=2mm,itemsep=0.05pt,topsep=0.05pt]
    \item \textbf{Vision Centric Gameplay.} 
    Models receive only visual input, requiring pixel-level scene understanding, object tracking, and spatial-temporal reasoning. V-MAGE features continuous-space environments, allowing models to explore the almost infinite state space. Each game is designed with different difficulty levels that target various skill dimensions.
    \item \textbf{Extensible Evaluation Framework.} 
    V-MAGE extends beyond model evaluation to assess agentic skills that are out-of-scope for current MLLMs. Our game-agent-model three-module evaluation pipeline allows optimizations in both MLLMs and their agent strategies.  
    \item \textbf{Adaptive ELO-based Ranking.} 
    V-MAGE uses a dynamic ELO system to provide a unified and interpretable metric across diverse games and difficulty levels. 
    Unlike raw scores varying in scale across tasks, ELO captures relative skill via win–loss dynamics between model performances on shared levels. 
    
\end{itemize}

\subsection{Evaluation Pipeline}
\label{subsec: pipeline}

V-MAGE separates the game environment from the MLLM, ensuring that all information is conveyed solely through visual input. 
Rather than evaluating raw inference latency, V-MAGE evaluates interactive frame-by-frame control in a simulated continuous-time environment. 
Concretely, when an action is requested, the game is paused while the model processes the visual inputs and generates its response. 
This frame-pausing mechanism decouples temporal reasoning from infrastructure-related delays and enables fair comparison across models with different serving conditions.

As depicted in Figure~\ref{fig:pipeline}, the system operates through iterative action cycles composed of three sequential components. 
The Game Module serves as the environment interface, executing game logic, capturing screenshots of the current state, and transmitting these visual frames to subsequent modules. 
The Agent Module integrates three critical data streams: (1) current visual inputs, (2) short-term temporal context from past observations, and (3) task-specific textual prompts such as game rules. 
This synthesized input is structured into a multimodal format compatible with the MLLM. 
The Model Execution Phase completes the cycle, wherein the MLLM generates an action command that undergoes semantic validation by the Agent Module before being relayed back to the Game Module for state updates.



To prioritize unbiased evaluation of core MLLM capabilities, V-MAGE’s architecture adopts a deliberately minimalistic design, avoiding auxiliary subsystems that might obscure model performance.  
The framework retains modular extensibility, allowing strategy modifications without altering core protocols, thereby ensuring benchmarking rigor while accommodating specialized needs.

V-MAGE incorporates five human-playable video games (Figure~\ref{fig:pipeline}), each featuring 3 to 10 levels, culminating in over 30 distinct evaluation environments. 
In contrast to traditional grid-based evaluation setups, V-MAGE selects games based on specific principles. The games feature free-form or continuous-space visual environments, facilitating more nuanced and flexible model movement and interaction. 
Crucially, to effectively assess vision-centric capabilities, the game environments are designed to be \textbf{visually irreducible}. 
This characteristic ensures that the system state cannot be fully discretized or textually summarized without significant information loss, thereby necessitating continuous visual grounding throughout the reasoning process.
Appendix~\ref{sec: Games in V-MAGE} details game selection criteria and sources.

\subsection{Games and Levels}


Existing game-based benchmarks indicate that MLLMs frequently struggle to achieve meaningful scores at standard human-level difficulties in conventional game-based benchmarks~\cite{zhang2024ingvpmllmsplayeasy, wang2025large}.

\begin{figure}[t]
\centering
\includegraphics[width=\columnwidth]{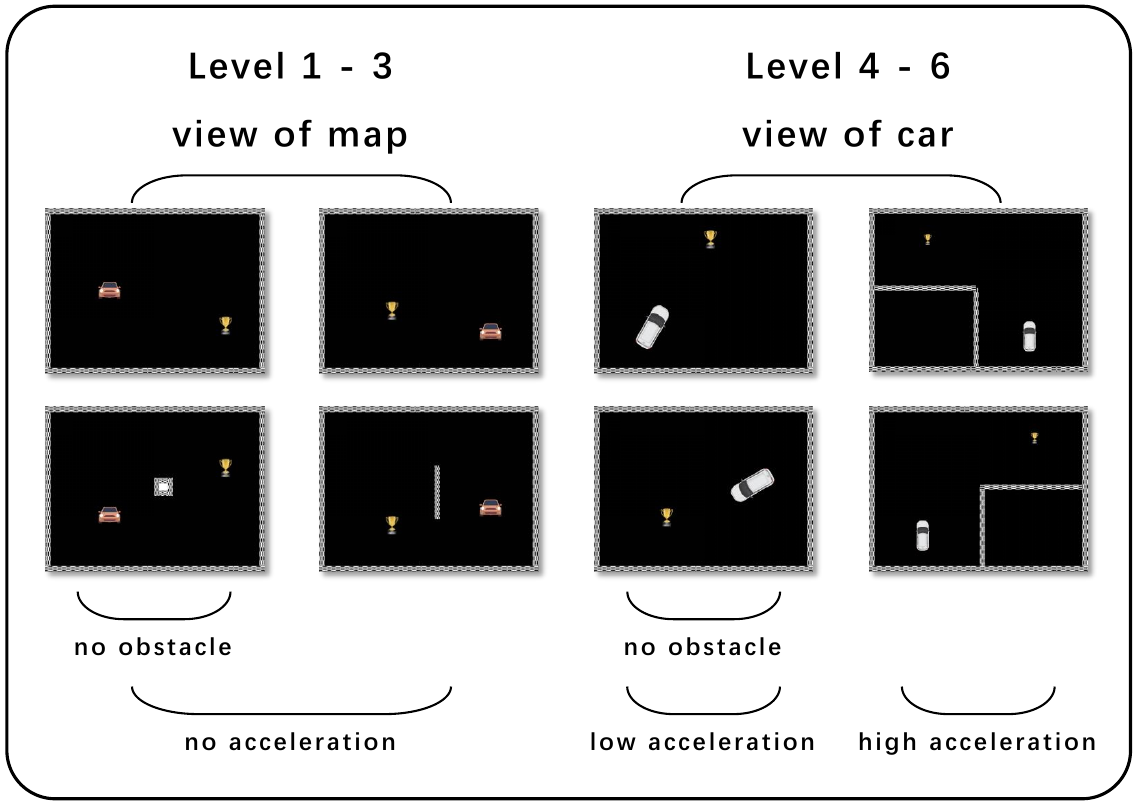}
\caption{\small
Race level design. Six levels progressively increase in difficulty while sharing the core objective: navigating a car to a trophy. Detailed Race level configurations are provided in Appendix Table\ref{tab:race level configs}.
}
\label{fig:race level design}
\end{figure}

This limits their discriminative power for fine-grained capability assessment and inter-model comparisons in complex tasks. 
To address this, V-MAGE introduces a multi-level assessment framework that evaluates models across various skill dimensions and provides granular performance diagnostics through difficulty-stratified tasks. 
Specifically, levels are designed for each game with gradually increasing complexity, varying control paradigms and perceptual challenges. 
For instance, Figure~\ref{fig:race level design} illustrates the level design in Race. Detailed information on the level design for all games can be found in Appendix~\ref{appendix: game design and implementation}.

\subsection{Evaluation Metrics}

V-MAGE employs a robust implementation of the ELO rating system to evaluate MLLMs, building on recent advancements in LLM benchmarking~\cite{duan2024gtbenchuncoveringstrategicreasoning, jiang2024genaiarenaopenevaluation}. 
This competitive evaluation framework ensures reliable model rankings by leveraging competitive evaluation mechanisms, ensuring robustness through dynamically balanced interactions. 

Games demonstrate diverse scales and difficulty thresholds.
The ELO system mitigates this variability by providing a standardized metric for comparing model performance across heterogeneous environments. 
Moreover, ELO is inherently sensitive to performance consistency. 
A model that achieves a high average score through a mix of exceptional successes and frequent failures may still be ranked lower, as its instability would likely lead to more losses in direct, pairwise matchups. 
This allows the ELO system to reward reliable performance over erratic, high-variance gameplay, which simple score averaging might otherwise obscure. 

Furthermore, it measures incremental advancement in games with non-linear scoring plateaus, where minor improvements vary by performance range. For instance, it differentiates progress from 80 to 85 and breakthroughs from 95 to 100.

We introduce an ELO-based ranking system to assess model performance by means of competitive pairwise comparisons.
In each game level, models are randomly matched in pairs for up to 100 evaluation rounds. The outcomes are determined based on their gameplay scores and the rates of valid actions taken.  
Detailed mathematical formulations of the pairing mechanism, rating updates, and stabilization process are provided in Appendix~\ref{sec: elo details}.

\subsection{Comparison to Existing Works}

\begin{table*}[htbp]
    \centering
    \caption{The comparison of V-MAGE with existing game-based evaluation benchmarks. 
    *Text in V-MAGE only represents the instructions for game rules and output format.}
    \label{tab:comparison-LVLM-bench}
    \renewcommand{\arraystretch}{1.3}
    \resizebox{\textwidth}{!}{
        \begin{tabular}{ccccc}
            \toprule
            Game Benchmarks & Game Type& Input & Reasoning Type & Level Design \\
            \cmidrule{1-1}\cmidrule{2-5}
            GameBench\cite{costarelli2024gamebenchevaluatingstrategicreasoning}  & Word & Text-Only & Text Reasoning & \textcolor{red}{\textbf{\ding{55}}} \\
            GameArena\cite{hu2024gamearenaevaluatingllmreasoning} & Word& Text-Only & Text Reasoning & \textcolor{red}{\textbf{\ding{55}}} \\
            GTBench\cite{duan2024gtbenchuncoveringstrategicreasoning}  & Word & Text-Only & Text Reasoning & \textcolor{red}{\textbf{\ding{55}}} \\
            ING-VP\cite{zhang2024ingvpmllmsplayeasy} & Grid Based & Single-Image-Text & Visual Aid & \textcolor{green}{\textbf{\ding{51}}} \\
            LVLM-Playground\cite{wang2025large} & Grid Based & Single-Image-Text & Visual Aid & \textcolor{green}{\textbf{\ding{51}}}\\
            BALROG\cite{paglieri2024balrogbenchmarkingagenticllm}  & Word / Grid Based & Single-Image-Text & Text / Visual Aid & \textcolor{red}{\textbf{\ding{55}}} \\
            Orak\cite{park2025orak} & Video & Single-Image-Text & Text / Visual Aid & \textcolor{red}{\textbf{\ding{55}}} \\
            \cmidrule{1-1}\cmidrule{2-5}
            \textbf{V-MAGE} & \textbf{Video} & \textbf{Multi-Images-Text*} & \textbf{Vision-Centric Reasoning} & \textcolor{green}{\textbf{\ding{51}}} \\
            \bottomrule
        \end{tabular}%
    }
\end{table*}

Humans play dynamic games using visual perception and intuitive reasoning, a process largely uncaptured by most existing MLLM game benchmarks. 
Many of these rely on grid-based games~\cite{wang2025large, zhang2024ingvpmllmsplayeasy} where states are textually representable. 
While such benchmarks assess text-based reasoning, similar to traditional LLM tasks~\cite{paglieri2024balrogbenchmarkingagenticllm}, they offer limited insights into MLLMs' visual intuitive reasoning. 
Models often bypass genuine visual perception here, acting as OCR converters, which hinders assessing and improving vision-centric abilities.~\cite{park2025orak} also employed video games as a testbed, wherein visual information remained auxiliary, and the game state was accessed chiefly through text-based inputs.

In contrast to this paradigm, V-MAGE shifts the evaluation focus by embedding models in dynamic visual environments that fundamentally require temporally grounded perception and action based on visual input. 
V-MAGE deliberately adopts environments lacking rigid grid structures, where the states of characters and objects cannot be easily simplified into sparse, coordinate-based textual descriptions. 
This design compels models to continuously leverage the visual modality during reasoning, rather than discarding it after an initial conversion. 

Furthermore, unlike benchmarks where decisions are made based on a single static frame, such as in many chess-like environments, V-MAGE requires sophisticated temporal reasoning across sequences of frames to make informed decisions, more closely mirroring human gameplay dynamics.

By shifting evaluation to more naturalistic and visually complex dynamic game environments, V-MAGE provides a more rigorous and representative test of MLLM capabilities, particularly in assessing their visual intuitive reasoning. A holistic comparison between V-MAGE and existing game benchmarks is presented in Table~\ref{tab:comparison-LVLM-bench}.

\section{Experiments}

\begin{table*}[htbp]
  \centering
  \caption{Performance comparison across different games based on the ELO ranking system.  The Random baseline refers to randomly selecting actions from the predefined action space during decision-making phases. Average performance ratio, abbreviated as \textbf{Avg. Ratio}, refers to the average percentage of the model's score compared to the human baseline score. }
  \label{tab:model_performance}
  \begin{adjustbox}{width=\textwidth}
  \begin{tabular}{@{}l|ccccc|c|c@{}}
    \toprule
    Model & Flappybird & Pong & Race & Supermario & Tempestrun & Avg. ELO Score & Avg. Ratio (\%) \\
    \midrule
    GPT‑5‑2025‑08‑07  & 1572 & \textbf{1939} & \textbf{1710} & 1584 & \textbf{1743} & \textbf{1710} & \textbf{43.4}\\
    Gemini‑2.5‑Pro  & 1526 & 1602 & 1660 & \textbf{1758} & 1474 & 1604 & 36.3\\
    Claude‑3.7‑Sonnet  & 1560 & 1570 & 1633 & 1582 & 1369 & 1543 & 30.8\\
    Gemini‑2.5‑Flash  & \textbf{1578} & 1524 & 1520 & 1531 & 1489 & 1528 & 23.8\\
    GPT-4o & 1557 & 1449 & 1581 & 1518 & 1527 & 1526 & 26.6\\
    Gemini-2.0-Flash-Thinking & 1517 & 1479 & 1503 & 1564 & 1516 & 1516 & 22.6 \\
    GPT‑5.1‑2025‑11‑13  & 1552 & 1514 & 1507 & 1449 & 1411 & 1486 & 20.1\\
    Gemini-2.0-Flash & 1494 & 1461 & 1437 & 1499 & 1530 & 1484 & 16.7 \\
    \midrule
    
    Qwen3‑VL‑235B‑A22B‑Instruct	 & 1567 & 1441 & 1517 & 1556 & 1496 & 1515 & 24.3\\
    Qwen2.5-VL-72B-Instruct & 1556 & 1442 & 1506 & 1541 & 1530 & 1515 & 22.8 \\
    InternVL2.5-78B & 1463 & 1462 & 1465 & 1543 & 1528 & 1492 & 19.2 \\
    Qwen2-VL-72B-Instruct & 1426 & 1445 & 1442 & 1505 & 1547 & 1473 & 16.5 \\
    InternVL2.5-8B & 1459 & 1448 & 1431 & 1373 & 1495 & 1441 & 12.9 \\
    Qwen2.5-VL-7B-Instruct & 1457 & 1446 & 1423 & 1354 & 1517 & 1439 & 12.1 \\
    Qwen2-VL-7B-Instruct & 1470 & 1447 & 1408 & 1362 & 1501 & 1438 & 11.4 \\
    Keye-VL-8B-Preview & 1419 & 1444 & 1428 & 1381 & 1499 & 1434 & 13.1\\
    Phi-4-multimodal-instruct & 1404 & 1454 & 1420 & 1482 & 1385 & 1429 & 13.7 \\
    Random & 1422 & 1434 & 1410 & 1417 & 1445 & 1426 & 11.0 \\

    \bottomrule
  \end{tabular}
  \end{adjustbox}
  \vspace{-15pt}
\end{table*}

As the baseline setting for the V-MAGE benchmark, we evaluate state-of-the-art MLLMs using full-precision models under a minimal naive agent strategy (Appendix~\ref{appedix agent module details}) to ensure a fair comparison.
The naive agent uses the most recent $k$ frames (typically $k=3$) for reasoning, together with reasoning history, prior actions, and game rules.
Detailed experimental settings and prompts can be found in Appendix~\ref{sec: experiment details} and Appendix~\ref{subsec: Games and Prompts}.

\subsection{Main Result}

\noindent
\textbf{Scores and Rankings.} 
The evaluation results clearly demonstrate a performance gradient across models ranging from 7B to 70B+ parameters. 
This also highlights that the dynamic visual reasoning tasks we propose represent a universal challenge for current MLLMs. 
Rankings from ELO scores and the Average Ratio may occasionally differ. 
This discrepancy arises because the ELO system rewards performance consistency (penalizing unstable, high-variance results) and provides a more balanced, holistic assessment across games with varying score scales. 
In contrast, the Avg Ratio metric can be skewed when averaging across tasks with imbalanced performance levels. More detailed analyses are provided in Appendix \ref{detailed statistics} and \ref{inconsistency analysis}.


\noindent
\textbf{Significant Performance Gap Between MLLMs and Humans in Complex Scenes.} 

We recruited five novice human participants to play the games under the same evaluation interface as the MLLMs and used their average scores as the human baseline. 
Participants were not specifically trained on the V-MAGE games; they received the same high-level natural-language rules as the models, observed the same screenshot-based inputs, and played under the same frame-pausing protocol. 
Appendix Figure~\ref{fig:model analysis (human)} compares leading MLLMs and human players across different levels.

These results should be interpreted as evidence about vision-driven decision-making under controlled dynamic tasks, rather than as a blanket equivalence between arbitrary game scores and general visual reasoning. 
Under the paused protocol and simplified low-difficulty settings, human novices remain near ceiling, while model performance drops sharply as tasks require more temporal understanding and strategic adaptation. 
Together with the text-state ablation in Figure~\ref{fig:perception skip}, this suggests that the gap reflects a combination of visual perception, temporal reasoning, and action-planning limitations rather than mere reaction-speed constraints.

\subsection{Further Analysis}

\begin{figure}[!t]
    \centering
    \begin{subfigure}[b]{0.47\columnwidth}
        \centering
        \includegraphics[width=\linewidth, valign=t]{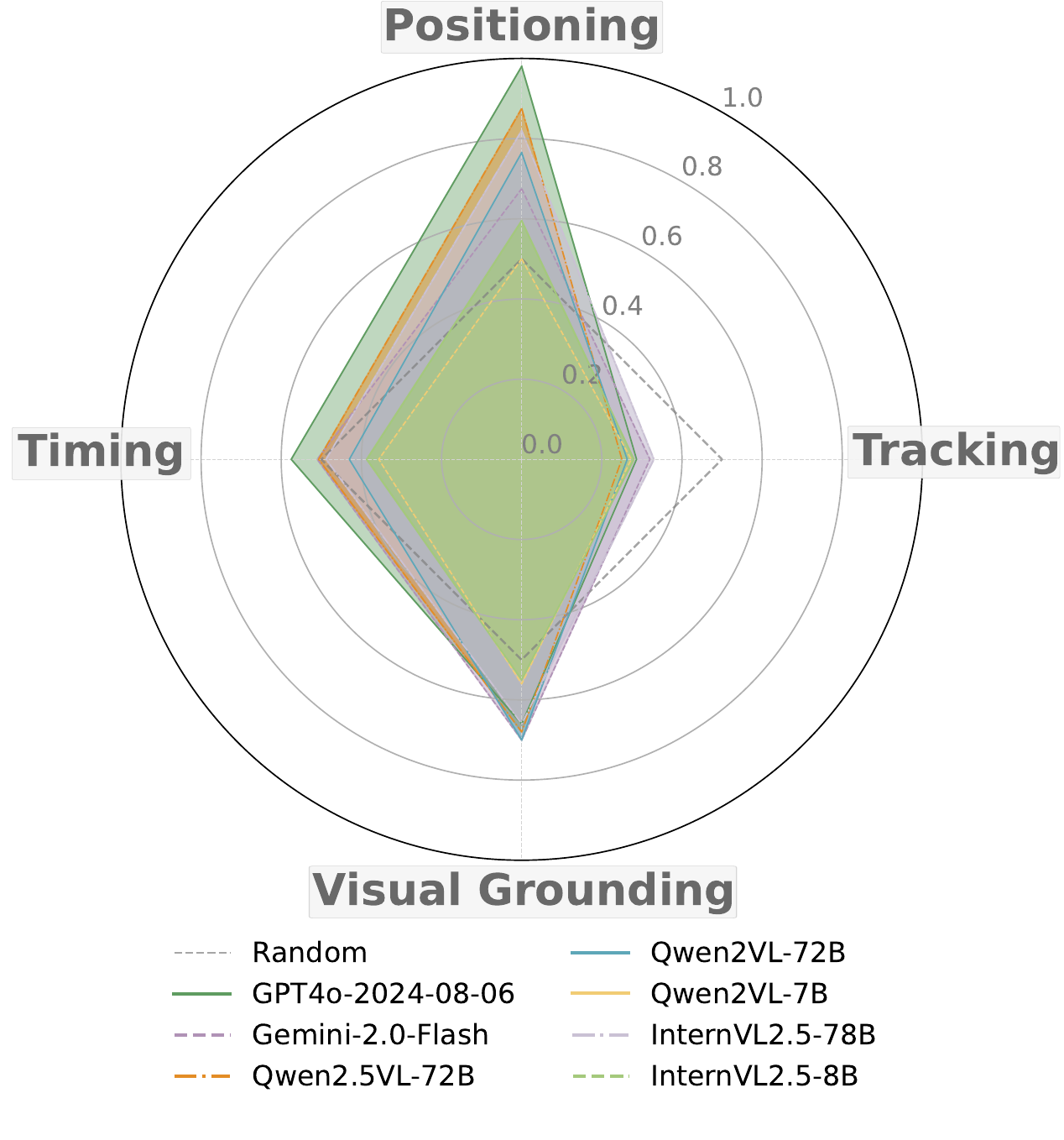}
        \caption{Capability maps of the underlying visual capabilities of each model.}
        \label{fig:abilities}
    \end{subfigure}
    \hfill
    \begin{subfigure}[b]{0.47\columnwidth}
        \centering
        \includegraphics[width=\linewidth, valign=t]{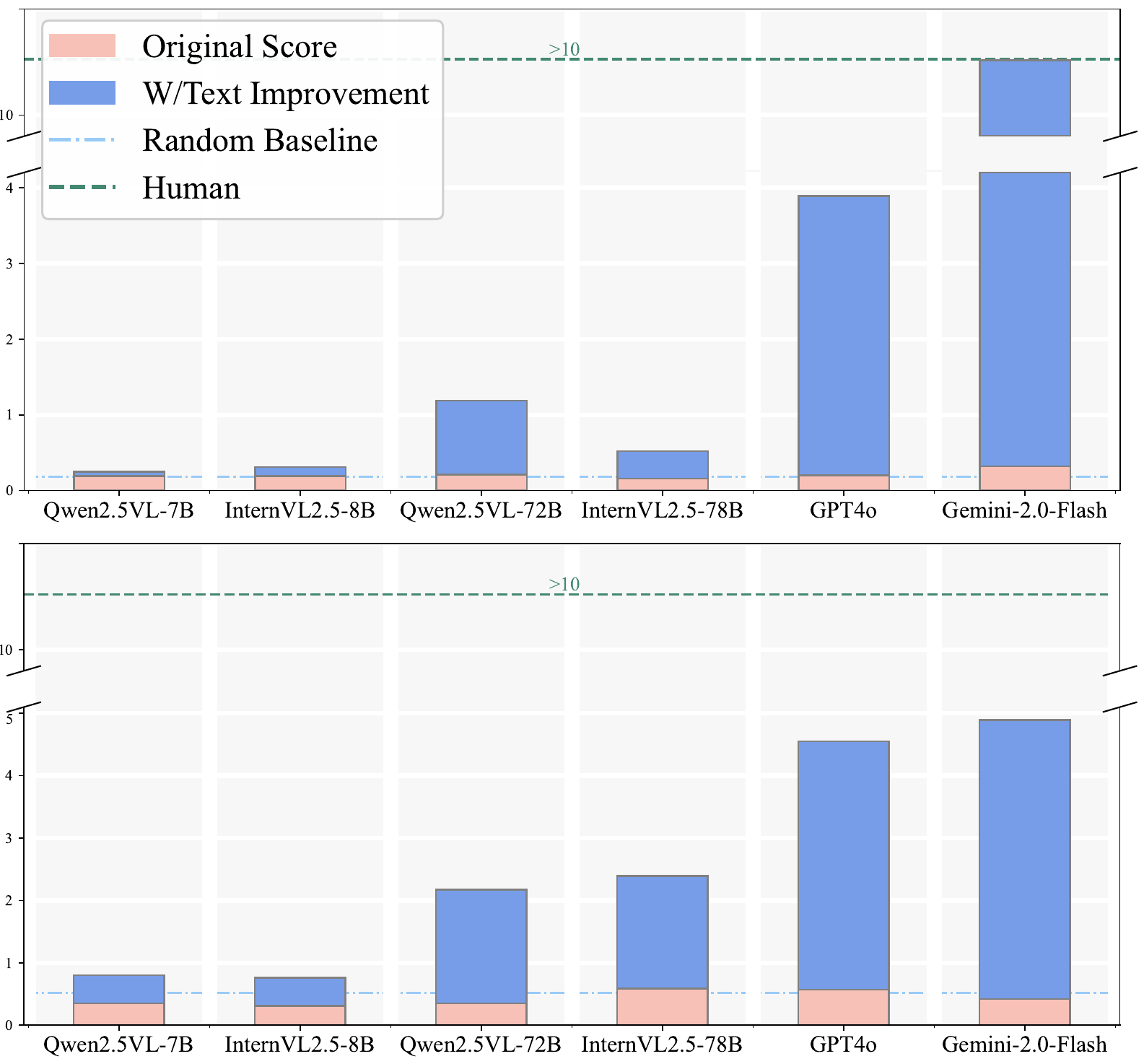}
        \caption{Model performance with vs. without text information on Pong Level~2 (top) and Flappy Bird Level~3 (bottom).}
        \label{fig:perception skip}
    \end{subfigure}
\end{figure}

\noindent
\textbf{Unit Tests for Core Visual Abilities.}  
We devised a unit test for vision-centric abilities by extracting foundational levels from V-MAGE. 
Figure \ref{fig:abilities} presents the capability profiles of various models across four core visual competencies. Scores near or below baseline suggest little effective relevant reasoning, while higher scores indicate a greater likelihood of correct reasoning.
For each capability, effective reasoning was evaluated by calculating the percentage of model scores that exceeded a random baseline score on the corresponding unit test levels (as defined in Appendix \ref{unit tests for core visual abilities appendix}).


As depicted, most models substantially outperform the random baseline in \textbf{Positioning} and \textbf{Visual Grounding}, indicating a degree of proficiency in single-frame image comprehension and basic visual information perception. 
However, performance notably declines in \textbf{Tracking} and \textbf{Timing}, which require processing continuous frame information and executing precise spatiotemporal judgments. For the \textbf{Tracking} task, nearly all models fail to significantly surpass the random baseline. 




\begin{figure*}[t]
\vspace{-10pt}
\begin{center}
\centerline{\includegraphics[width= \textwidth]{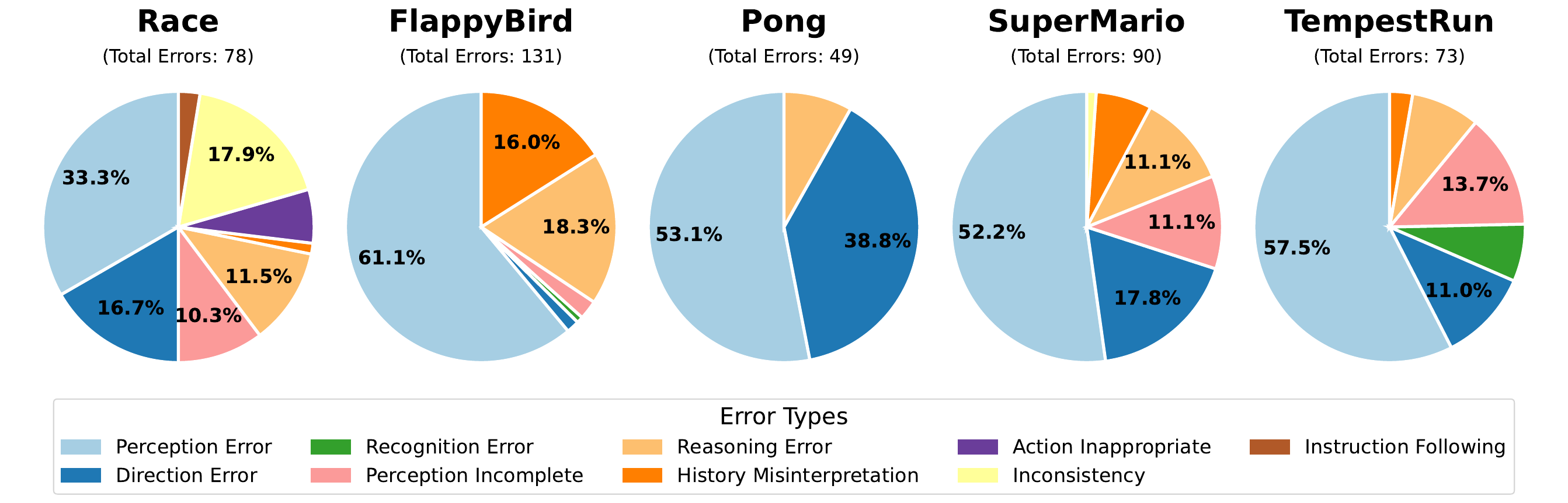}}
\caption{
Error type probability distribution for GPT4o across 494 samples.
}
\label{fig:error analysis}
\end{center}
\vspace{-10pt}
\end{figure*}

\begin{table*}[b]
\centering
\small
\caption{Average number of rounds for each model to generate different responses.}
\label{tab:avearge rounds for changing responses}
\renewcommand{\arraystretch}{1.0}
\begin{adjustbox}{width=\textwidth}

\begin{tabular}
{
l >{\centering\arraybackslash}p{1.5cm} >{\centering\arraybackslash}p{1.5cm} >{\centering\arraybackslash}p{1.5cm} >{\centering\arraybackslash}p{1.5cm} >{\centering\arraybackslash}p{1.5cm}}
\toprule
\textbf{Model} & \textbf{Race} & \textbf{FlappyBird} & \textbf{Pong} & \textbf{TempestRun} & \textbf{Avg.} \\
\midrule
Qwen2VL 7B & 4.3 & 25.9 & 13.7 & 7.3 & 12.8 \\
Qwen2.5VL 72B & 2.3 & 19.3 & 2.6 & 5.3 & 7.4 \\
InternVL2.5 8B & 2.0 & 6.9 & 6.7 & 8.0 & 5.9 \\
InternVL2.5 78B & 6.8 & 16.0 & 2.0 & 3.0 & 7.0 \\
GPT4o & \textbf{1.0} & \textbf{1.6} & \textbf{1.0} & \textbf{1.0} & \textbf{1.1} \\
\midrule
\textbf{PCC $r$} & \multirow{2}{*}{-0.57} & \multirow{2}{*}{-0.71} & \multirow{2}{*}{-0.87} & \multirow{2}{*}{-0.72} & \multirow{2}{*}{-0.72} \\
(Avg. Rounds vs. ELO) & & & & & \\
\bottomrule
\end{tabular}
\end{adjustbox}
\end{table*}

\noindent
\textbf{Limitations Beyond Visual Perception.} 
While visual perception constitutes a critical assessment dimension in V-MAGE's game tasks, our experiments revealed additional limitations and deficiencies in other aspects. 
To validate this, we conducted supplementary experiments in simple levels providing textual game state descriptions, thereby bypassing the perception process.

As shown in Figure \ref{fig:perception skip}, providing textual descriptions of the game state significantly improved the performance of most evaluated models, with this gain being particularly prominent in larger models such as Gemini and in games requiring precise state understanding like Pong. This notable performance increase when perception is bypassed strongly suggests that limitations in processing visual information are indeed a significant bottleneck for current MLLMs. 

However, despite this substantial performance gain, the models' scores still remained considerably lower than the human baseline in most cases. This persistent gap indicates that while visual perception challenges are critical, the models' limitations extend beyond merely ``seeing'' the state accurately. It highlights that significant bottlenecks also exist in the downstream processes responsible for robust interpretation of information (even when provided textually or perceived imperfectly), strategic planning, and effective action generation in complex and dynamic environments.
Furthermore, the less pronounced performance improvement observed in smaller models(like Qwen2.5VL 7B) suggests that inherent limitations in their core reasoning capabilities may also act as a performance bottleneck.
Check Appendix \ref{perceptual skipping exp appendix} for more experimental details.

\noindent
\textbf{Anchoring Bias in Model Inference Processes.} 
When provided with historical information to aid reasoning in dynamic game progression, MLLMs often exhibit anchoring bias, particularly when processing similar consecutive frames. 
This bias manifests as an undue influence of prior inferences on current reasoning, hindering the accurate identification of subtle visual changes and unique frame details. 
Models tend to favor relying on historical textual descriptions over nuanced visual input, making them less sensitive to fine-grained visual updates, consequently leading to unchanged reasoning content over extended game sequences.

\begin{figure*}[t]
\begin{center}
\centerline{\includegraphics[width= \textwidth]{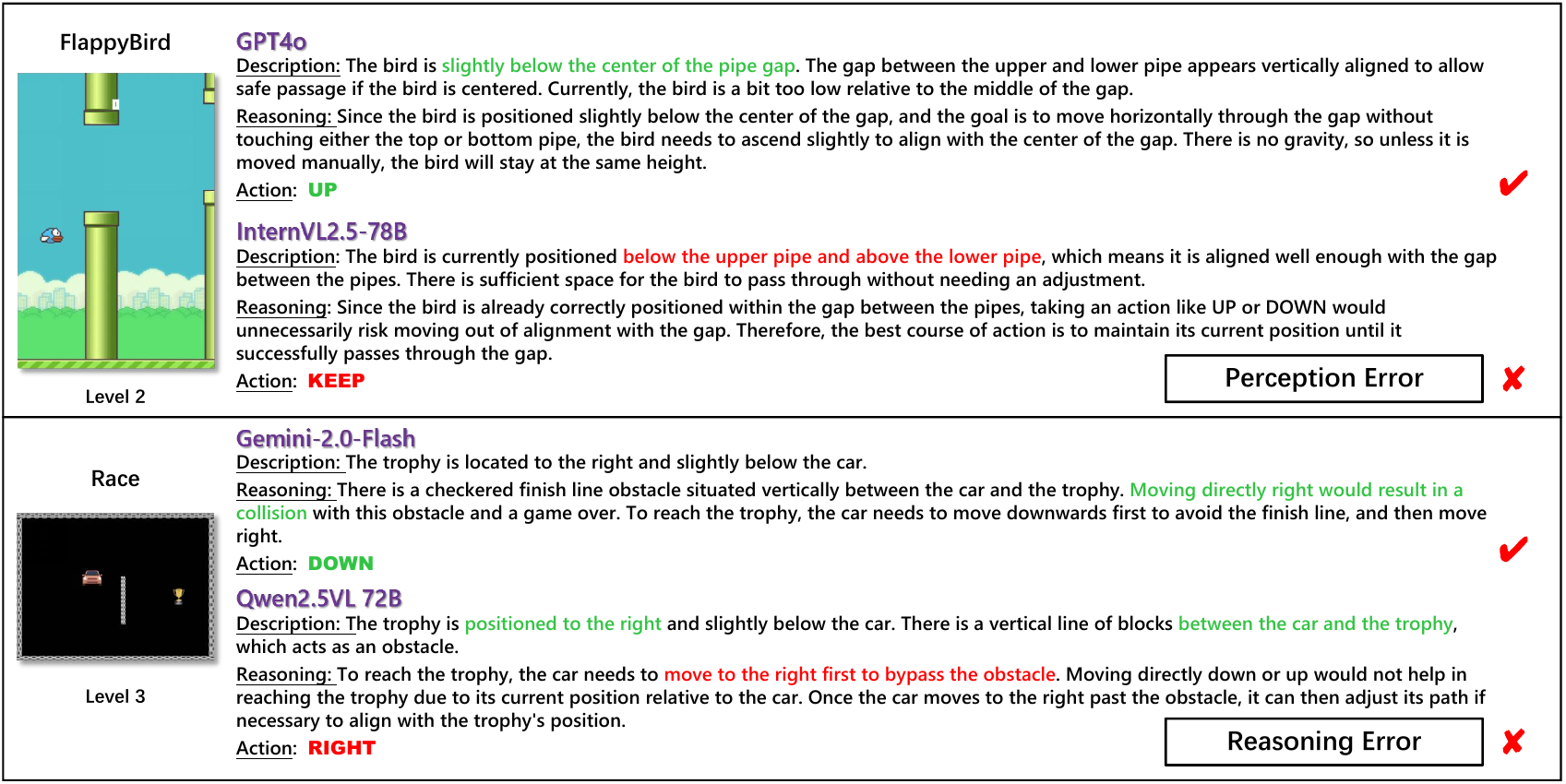}}
\caption{
Case examples illustrating Perception Error and Reasoning Error in FlappyBird and Race. 
The FlappyBird example shows a \textit{Perception Error} where the model misjudges the bird's vertical position relative to the pipe gap. 
The Race example illustrates a \textit{Reasoning Error} where the model fails to plan a path around an obstacle between the car and the trophy, resulting in a suboptimal action.
}
\label{fig: perception and reasoning error case}
\end{center}
\end{figure*}

As shown in Table \ref{tab:avearge rounds for changing responses}, models vary significantly in their responsiveness; for instance, in FlappyBird, Qwen2.5VL 72B altered its reasoning only once every 19.3 responses on average, significantly less frequently than GPT-4o (1.6 responses). The Pearson correlation coefficients (\textbf{PCC} $r$) reveal a consistent negative correlation between the average rounds to change response and ELO score across games, with an average $r$ of \textbf{-0.72}. 
This highlights a critical challenge in maintaining responsiveness to dynamic visual input and its direct impact on task success. 
To investigate the impact of pipeline settings (e.g., frame sampling and decision frequency) on anchoring bias, we conduct experiments, with results presented in Appendix \ref{anchoring bias appendix}.

 \noindent
\textbf{Analysis of Model Errors in V-MAGE.}  
For GPT-4o's complete inputs and responses across all game levels after one to two rounds of gameplay, we uniformly sampled 494 interaction sets for manual annotation and categorized the primary error types. 
The visualization results depicting the distribution of these errors are presented in Figure \ref{fig:error analysis}. The annotation method for error distribution and statistical details are provided in Appendix \ref{Apeendix Analysis of GPT4o Errors in V-MAGE}.

The predominant error type was \textit{perception error}, with \textit{direction error} being particularly prevalent. 
In such cases, the model frequently misidentified directional cues in visual content, leading to incorrect inferences. 
Another major category was \textit{reasoning error}, primarily involving logical flaws and decision-making failures, including misinterpretation of historical information (\textit{history misinterpretation}) and suboptimal action selection (\textit{action inappropriate}). Figure \ref{fig: perception and reasoning error case} presents examples illustrating perception and reasoning errors.

Additionally, \textit{perception incomplete} errors were commonly observed, where the model failed to fully extract useful information from visual inputs, resulting in partial information acquisition. 
\textit{Inconsistency} errors mainly occurred in scenarios permitting multiple valid solutions, where the model exhibited unstable decision-making by frequently revising its choices, ultimately leading to timeout failures due to excessive deliberation. 
Notably, \textit{instruction following} errors were virtually absent, as GPT-4o consistently adhered to the provided prompts. Additional case study analyses are documented in Appendix \ref{sec: Case Study}.

\noindent
\textbf{V-MAGE Poses Significant Challenges to MLLMs.}
Unlike conventional static VQA or text-reducible grid-based benchmarks, V-MAGE evaluates interactive frame-by-frame control in dynamic, vision-centric game environments under a paused continuous-time protocol. 
The framework exposes persistent limitations in current MLLMs. 
Models struggle to integrate information across sequences of frames, which affects tracking, temporal reasoning, and trajectory understanding. 
This difficulty may contribute to anchoring bias, as models can over-rely on prior inferences and under-react to subtle visual changes. 
Furthermore, MLLMs exhibit fundamental deficiencies in planning, strategic decision-making, and action generation. 
These limitations persist even when initial visual processing challenges are mitigated, highlighting that deficiencies in the core reasoning process itself extend beyond perception.

\vspace{-5pt}

\section{Conclusion}
\label{sec: conclusion}


This paper introduces V-MAGE, a pioneering game-based evaluation framework designed to assess the vision-centric capabilities of MLLMs in dynamic, interactive environments. Using over 30 levels across 5 games, we reveal significant limitations: models exhibit insufficient multi-image perception, leading to issues like anchoring bias, and demonstrate fundamental deficiencies in complex reasoning and strategic planning that persist even when perceptual challenges are mitigated.  Highlighting the need for enhanced multi-frame processing and strategic planning, V-MAGE establishes a rigorous standard to drive the development of robust, human-like visual intelligence.

\section*{Limitations}

While V-MAGE represents a significant step towards evaluating MLLMs in interactive, dynamic, and visually complex environments that closer resemble real-world tasks, the current benchmark is still constrained by the inherent scale and complexity of the included games.
This is a current boundary imposed by balancing complexity with controllability for systematic evaluation. 
As MLLM capabilities continue to advance and hardware performance improves, we anticipate being able to incorporate larger and more complex game environments in future iterations.
These future environments will be designed to offer a wider array of challenges, further pushing the limits of MLLM evaluation and narrowing the gap between simulated and real-world performance assessment.

\section*{Ethical Considerations}

This research contributes to the field of multimodal models by providing a novel and challenging benchmark for evaluating vision-centric capabilities in dynamic environments. The primary positive impact is facilitating the diagnosis of limitations in current MLLMs and guiding future research towards developing more capable, robust, and potentially safer AI systems for real-world interaction. As our work focuses on foundational evaluation in simulated environments and does not involve the deployment of high-risk models or the collection of sensitive personal data, the potential for negative societal impacts is considered minimal and indirect at this stage. We believe that developing better evaluation tools is a crucial step towards building more reliable and trustworthy AI.

\section*{Acknowledgment}
This work is supported by the Frontier Technologies R\&D Program of Jiangsu (BF2024059), the National Natural Science Foundation of China (Grant \#62572229 and \#62502544), and the Information Technology Service Center of Nanjing University.

\bibliography{custom}

\appendix

\clearpage
\section*{Appendix Overview}

\renewcommand{\arraystretch}{1.8} 

In the appendix, we provide the following contents:


Sec \ref{sec: experiment details}: Introduces experimental settings and provides detailed information on the experiments, models, and their performance.

Sec \ref{sec: Games in V-MAGE}: Delineates the game selection methodology and sources, including all level designs and prompts.

Sec \ref{sec: elo details}: Presents specific details of ELO-based ranking system in V-MAGE.

Sec \ref{sec: ablation study}: Provides ablation studies on pipeline settings(e.g., frame sampling strategy, resolution, etc.)

Sec \ref{sec: additional experiments}: Provides details of supplemental experimental analyses.

Sec \ref{sec: miscellaneous material}: Provides details on miscellaneous material, including a \textbf{statement of LLM usage}(Sec. \ref{LLM usage}) and a discussion about broader impacts.

Sec \ref{sec: Case Study}: Presents some case studies.

\section{Experiments Details}
\label{sec: experiment details}

\subsection{Evaluation Pipeline Details}

V-MAGE employs a three-module architecture, as illustrated in Figure \ref{fig:pipeline}. The specific configurations and details within each module are as follows:
\vskip -0.08in

\subsubsection{Game Module}
\vskip -0.08in

The Game module encompasses the game simulators and their operational parameters. 
In V-MAGE, game simulators, representing environments developed with Pygame, are configured to present tasks that test specific visual and reasoning capabilities. 
To address potential confounding factors such as API latency and computational constraints, V-MAGE employs a frame-pausing mechanism during model inference. 
This ensures that the game environment remains static while the model processes visual inputs and generates actions, effectively decoupling timing evaluation from raw inference speed. 

Regarding real-time execution and inference delays, the frame-pausing mechanism ensures fairness across models. 
While current models may not consistently achieve real-time inference due to API or GPU limitations, V-MAGE automatically pauses the game environment during model processing. 
This approach isolates the model’s temporal reasoning ability (strategic "when to act") from infrastructure-related delays, enabling a focused assessment of capabilities such as Timing.
In V-MAGE, Timing is explicitly designed to evaluate the model’s ability to choose the optimal timing of actions, not the system’s response speed.

The frames per second (FPS) for all our games is 30. 
In our standard benchmark setting, we use a sample rate of 3. 
This corresponds to the model making a decision approximately every 100 milliseconds (30 FPS / 3) in the game.

\vskip -0.08in
\subsubsection{Agent Module}
\label{appedix agent module details}
\vskip -0.08in


Researchers can modify the agent's operational mode by adjusting the configuration file. This includes altering historical strategies, such as employing a longer history of steps or sparsely sampling more distant historical records to provide different contextual information to the models. However, for fairness in evaluation and due to resource constraints, we present results obtained using only the naive configuration as a baseline in our main findings. This allows for standardized comparisons across different MLLMs.

In the baseline agent strategy, we utilize historical information comprising three preceding screenshots and the model's reasoning outputs associated with these screenshots, combined with the current game screenshot, to form the \texttt{<history>} and \texttt{<image\_current\_step>} components (as shown in Sec. \ref{subsec: Games and Prompts}) that are incorporated into the model's input framework.

The \texttt{<history>} component is structured as follows:

\noindent
\fbox{%
  \parbox{\dimexpr\linewidth-2\fboxsep-2\fboxrule\relax}{%
    This screenshot is \texttt{<step>} steps before the current step of the game.
    After this frame, your reasoning message was \texttt{<history\_reasoning>}.
    After the action was executed, the game info was \texttt{<history\_action\_info>}.
 }%
}

Upon obtaining the model's output, the system parses the structured response format specified in the prompt to systematically extract both the cognitive reasoning log and action parameters. These extracted elements are subsequently stored and fed back to the game module through a standardized data pipeline. This process ensures consistent interpretation of the model's decision-making trajectory while maintaining synchronization with the game state.

Taking an input from the actual evaluation process in the \textbf{Tempest Run} as an example:

\noindent
\fbox{%
  \parbox{\dimexpr\linewidth-2\fboxsep-2\fboxrule\relax}{%
        Assume you are playing a PC game called 'Tempest Run'. 

You need to control a character who moves through a three-dimensional space inside a futuristic tunnel filled with various obstacles and enemies. Your goal is to navigate through the tunnel, avoid or overcome obstacles, and run as far as possible. Avoid colliding with red spikes, purple walls, or failing to deal with green enemies.  


Use the optimal combination of movements to progress through the tunnel smoothly and efficiently. Monitor the character's position relative to obstacles and react appropriately to avoid losing progress.  
Now, I will give you some history screenshots in the game for decision making.

\texttt{<image>}

This screenshot is three steps before the current step of the game. After this frame, your reasoning message was "{'observation': 'The character is running forward in a narrow tunnel with blue lines forming the walls and ceiling. Red spikes are visible on the ground ahead, and green enemies are further down the tunnel. The path appears clear with no immediate obstacles or enemies to avoid.', 'reasoning': 'Maintaining the current trajectory with no action is optimal to continue moving straight towards the green enemy and prepare for a potential slide to eliminate it.', 'action': 'NONE'}". After the action was executed, the game info was "Game is running."

...

\texttt{<image>}

This screenshot represents the current step of the game.  

The last frame shows the current state of the game, while the previous frames show the character's previous movements.  

Important notes:
1. Use JUMP to jump over red spikes on the ground.  
2. Use SLIDE to duck and kick green enemies to eliminate them.

...

You should think step by step and respond with the following format, remember to respond with plain text without any special characters or symbols, DO NOT respond in markdown or Latex or any other format.  

Response:  

Observation: ... (Describe the character's current position and nearby obstacles or enemies.) 

}
}

\noindent
\fbox{%
  \parbox{\dimexpr\linewidth-2\fboxsep-2\fboxrule\relax}{%
Reasoning: ... (Think step by step and explain how you choose the action.)  

Action: ... (Choose ONE of the six actions to control the character. Do NOT add any other words.)

  }%
}

\vskip -0.08in
\subsubsection{Model Module}
\label{subsubsec: model module}
\vskip -0.08in

The Model module is primarily responsible for model deployment and parameter control. In addition to closed-source models accessed via APIs, we deployed open-source models on an \textbf{8×V100 GPU Azure cluster} and an \textbf{8×A100 GPU Azure cluster} for our experiments, utilizing the \textbf{vLLM} library for efficient serving. For text output generation across all models, we standardized the decoding parameters by setting \textbf{top\_p}=0.9 and \textbf{temperature}=0.8.

\subsection{Detailed Statistics}
\label{detailed statistics}

\subsubsection{Visualization}

\begin{figure*}[htbp]
\begin{center}
\centerline{\includegraphics[width= \textwidth]{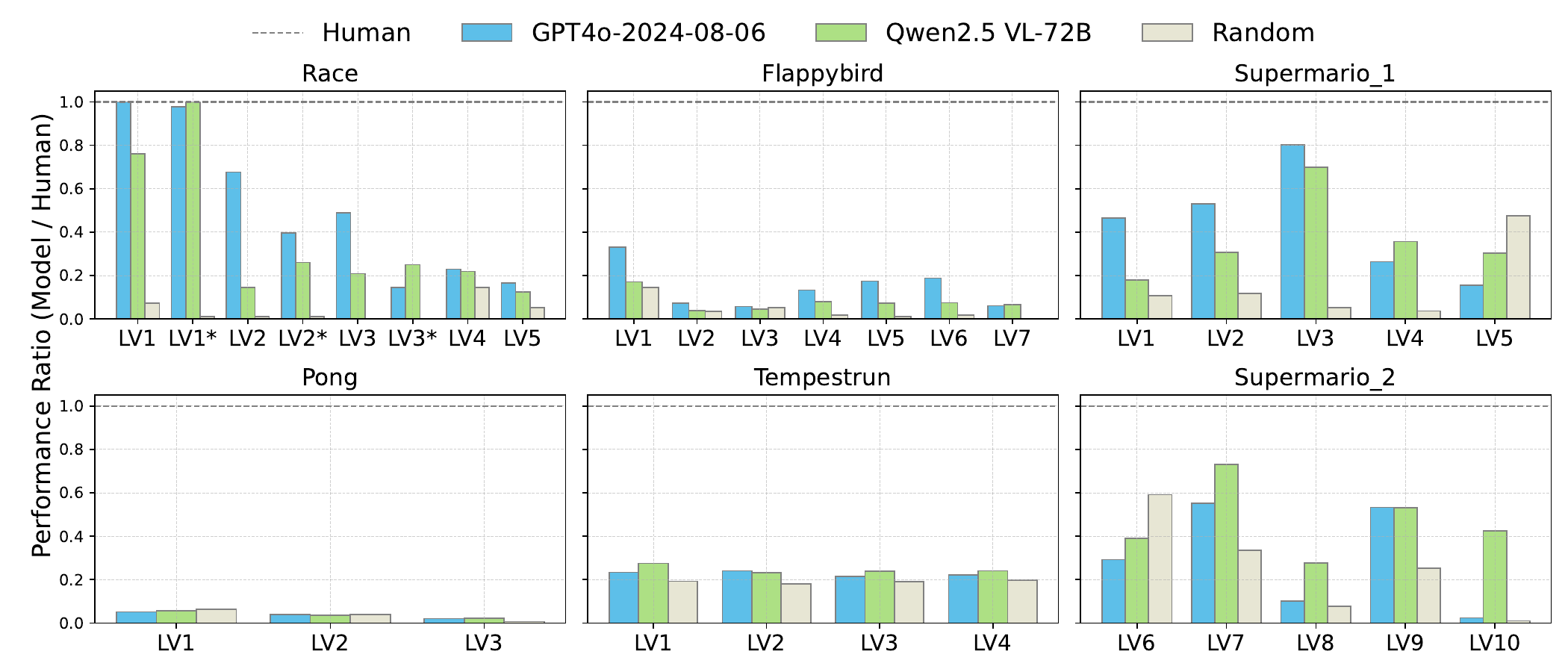}}
\caption{\small
\textbf{The MLLM trails humans by a large margin in all six games. } 
The levels with an asterisk (*) represent `no history'. Detailed performance metrics for each model across individual game levels are provided in Appendix \ref{detailed statistics} (Tables \ref{tab:race_performance}-\ref{tab:tempestrun_performance}).
}
\label{fig:model analysis (human)}
\end{center}
\vspace{-10pt}
\end{figure*}

Figure \ref{fig:model analysis (human)} compares the performance of leading MLLMs and human players across different game levels. 

\subsubsection{Score-based Performance} 

Cross-task result analysis reveals the limitations of parameter scaling: In Race Level 1 (with historical frame input), Qwen2VL showed a 429\% improvement in score when scaling from 7B to 72B (from 10.43 to 55.19), reaching about 55\% of the human baseline score. However, in more complex tasks such as Tempestrun Level 4, InternVL2.5-78B (199.78 points) only improved by 14.4\% compared to its 8B version (174.58 points), still achieving only 25\% of the human score (800 points). This suggests that parameter scaling cannot compensate for the inherent ability gap in complex dynamic tasks. 
The detailed scores are presented in 
Tables~\ref{tab:race_performance}, 
\ref{tab:pong_performance}, 
\ref{tab:supermario_performance}, 
\ref{tab:flappybird_performance}, and 
\ref{tab:tempestrun_performance}.

\begin{table*}[htbp]
    \centering
    \small
    \caption{Performance analysis based on average scores in Race}
    \label{tab:race_performance}
    \renewcommand{\arraystretch}{1.5}
    \resizebox{\textwidth}{!}{
    \begin{tabular}{l|c|c|c|c|c|c|c|c|c}
    \toprule
    \multirow{2}{*}{Level} & \multirow{2}{*}{GPT-4o} & Gemini & Qwen2.5VL & Qwen2VL & Qwen2VL & InternVL2.5 & InternVL2.5 & \multirow{2}{*}{Random} & \multirow{2}{*}{Human} \\
    & & 2.0-flash & 72B & 7B & 72B & 8B & 78B & & \\
        \midrule
        Level1 & \textbf{99.99} & 35.41 & 76.01 & 10.43 & 55.19 & 28.12 & 64.56 & 7.30 & 100.00 \\
        Level1 No History & 97.87 & 98.91 & 99.95 & 87.46 & 97.87 & 89.54 & \textbf{99.99} & 1.06 & 100.00 \\
        Level2 & \textbf{67.68} & 7.30 & 14.59 & 0.00 & 1.06 & 3.14 & 15.63 & 1.06 & 100.00 \\
        Level2 No History & \textbf{39.57} & 22.92 & 26.04 & 1.06 & 23.96 & 5.22 & 26.04 & 1.06 & 100.00 \\
        Level3 & \textbf{48.94} & 5.22 & 20.84 & 4.18 & 7.30 & 6.26 & 11.47 & 0.02 & 100.00 \\
        Level3 No History & 14.59 & 4.18 & \textbf{25.00} & 4.18 & 16.67 & 11.47 & 22.92 & 0.00 & 100.00 \\
        Level4 & \textbf{22.92} & 6.26 & 21.88 & 0.02 & 2.10 & 0.00 & 2.10 & 14.59 & 100.00 \\
        Level5 & 16.67 & 8.34 & 12.51 & 13.55 & 4.18 & \textbf{20.84} & 7.30 & 5.22 & 100.00 \\
        \bottomrule
    \end{tabular}
    }
\end{table*}

\begin{table*}[htbp]
    \centering
    \small
    \caption{Performance analysis based on average scores in Pong}
    \label{tab:pong_performance}
    \renewcommand{\arraystretch}{1.8}
    \resizebox{\textwidth}{!}{
    \begin{tabular}{l|c|c|c|c|c|c|c|c|c}
    \toprule
    \multirow{2}{*}{Level} & \multirow{2}{*}{GPT-4o} & Gemini & Qwen2.5VL & Qwen2VL & Qwen2VL & InternVL2.5 & InternVL2.5 & \multirow{2}{*}{Random} & \multirow{2}{*}{Human} \\
    & & 2.0-flash & 72B & 7B & 72B & 8B & 78B & & \\
        \midrule
        Level1 & 0.51 & 0.54 & 0.56 & 0.59 & 0.54 & 0.68 & \textbf{0.77} & 0.63 & 10.00 \\
        Level2 & 0.39 & \textbf{0.41} & 0.35 & 0.31 & 0.33 & 0.31 & 0.38 & 0.39 & 10.00 \\
        Level3 & 0.19 & \textbf{0.32} & 0.21 & 0.18 & 0.20 & 0.18 & 0.15 & 0.06 & 10.00 \\
        \bottomrule
    \end{tabular}
    }
\end{table*}

\begin{table*}[htbp]
    \centering
    \small
    \caption{Performance analysis based on average scores in Supermario}
    \label{tab:supermario_performance}
    \renewcommand{\arraystretch}{1.8}
    \resizebox{\textwidth}{!}{
    \begin{tabular}{l|c|c|c|c|c|c|c|c|c}
    \toprule
    \multirow{2}{*}{Level} & \multirow{2}{*}{GPT-4o} & Gemini & Qwen2.5VL & Qwen2VL & Qwen2VL & InternVL2.5 & InternVL2.5 & \multirow{2}{*}{Random} & \multirow{2}{*}{Human} \\
    & & 2.0-flash & 72B & 7B & 72B & 8B & 78B & & \\
        \midrule
        Level10 & 18.77 & 108.30 & \textbf{339.57} & 12.51 & 29.16 & 14.61 & 80.19 & 8.36 & 800.00 \\
        Level1 & \textbf{372.85} & 109.41 & 142.76 & 33.41 & 216.67 & 69.83 & 203.12 & 86.50 & 800.00 \\
        Level2 & \textbf{424.92} & 127.17 & 244.78 & 102.12 & 338.47 & 102.12 & 186.48 & 94.83 & 800.00 \\
        Level3 & \textbf{802.99} & 429.10 & 697.91 & 188.54 & 565.46 & 286.44 & 610.26 & 53.19 & 1000.00 \\
        Level4 & 369.76 & 251.07 & \textbf{499.89} & 112.53 & 346.84 & 151.09 & 447.84 & 52.15 & 1400.00 \\
        Level5 & 125.08 & 258.33 & 242.72 & 232.29 & 192.75 & 209.41 & \textbf{433.23} & 380.13 & 800.00 \\
        Level6 & 233.36 & 325.96 & 311.36 & 324.96 & 296.79 & 267.70 & 344.74 & \textbf{472.78} & 800.00 \\
        Level7 & 440.66 & 527.96 & \textbf{585.21} & 161.48 & 490.48 & 220.86 & 491.52 & 268.74 & 800.00 \\
        Level8 & 91.75 & 211.43 & \textbf{248.96} & 51.13 & 179.20 & 76.09 & 168.74 & 68.79 & 900.00 \\
        Level9 & 693.56 & 594.67 & 690.46 & 162.62 & 508.24 & 201.12 & \textbf{756.02} & 329.19 & 1300.00 \\
        \bottomrule
    \end{tabular}
    }
\end{table*}

\begin{table*}[htbp]
    \centering
    \small
    \caption{Performance analysis based on average scores in Flappybird}
    \label{tab:flappybird_performance}
    \renewcommand{\arraystretch}{1.8}
    \resizebox{\textwidth}{!}{
    \begin{tabular}{l|c|c|c|c|c|c|c|c|c}
    \toprule
    \multirow{2}{*}{Level} & \multirow{2}{*}{GPT-4o} & Gemini & Qwen2.5VL & Qwen2VL & Qwen2VL & InternVL2.5 & InternVL2.5 & \multirow{2}{*}{Random} & \multirow{2}{*}{Human} \\
    & & 2.0-flash & 72B & 7B & 72B & 8B & 78B & & \\
        \midrule
        Level1 & \textbf{3.30} & 2.38 & 1.70 & 0.76 & 0.47 & 1.20 & 1.54 & 1.45 & 10.00 \\
        Level2 & \textbf{0.71} & 0.47 & 0.38 & 0.20 & 0.12 & 0.36 & 0.39 & 0.34 & 10.00 \\
        Level3 & \textbf{0.57} & 0.41 & 0.45 & 0.20 & 0.35 & 0.33 & 0.43 & 0.52 & 10.00 \\
        Level4 & 1.33 & 1.50 & 0.79 & \textbf{1.52} & 0.38 & 1.43 & 0.64 & 0.16 & 10.00 \\
        Level5 & \textbf{1.74} & 1.38 & 0.71 & 1.44 & 0.51 & 1.20 & 0.49 & 0.10 & 10.00 \\
        Level6 & \textbf{1.88} & 1.05 & 0.73 & 1.62 & 0.56 & 1.14 & 0.66 & 0.17 & 10.00 \\
        Level7 & 0.60 & 0.07 & \textbf{0.66} & 0.03 & 0.14 & 0.00 & 0.13 & 0.00 & 10.00 \\
        \bottomrule
    \end{tabular}
    }
\end{table*}

\begin{table*}[htbp]
    \centering
    \small
    \caption{Performance analysis based on average scores in Tempestrun}
    \label{tab:tempestrun_performance}
    \renewcommand{\arraystretch}{1.8}
    \resizebox{\textwidth}{!}{
    \begin{tabular}{l|c|c|c|c|c|c|c|c|c}
    \toprule
    \multirow{2}{*}{Level} & \multirow{2}{*}{GPT-4o} & Gemini & Qwen2.5VL & Qwen2VL & Qwen2VL & InternVL2.5 & InternVL2.5 & \multirow{2}{*}{Random} & \multirow{2}{*}{Human} \\
    & & 2.0-flash & 72B & 7B & 72B & 8B & 78B & & \\
        \midrule
        Level1 & 466.25 & 478.35 & \textbf{549.98} & 446.92 & 519.22 & 444.71 & 475.22 & 385.72 & 2000.00 \\
        Level2 & 361.44 & 356.05 & 349.06 & 352.76 & \textbf{370.13} & 327.38 & 333.37 & 271.65 & 1500.00 \\
        Level3 & 213.73 & 197.91 & \textbf{238.74} & 208.75 & 220.21 & 197.71 & 216.64 & 190.71 & 1000.00 \\
        Level4 & 177.60 & \textbf{201.67} & 192.79 & 182.91 & 195.19 & 174.58 & 199.78 & 157.17 & 800.00 \\
        \bottomrule
    \end{tabular}
    }
\end{table*}




\subsubsection{Additional Indicators}
\label{subsubsec: additional indicators}

Due to the dynamic game environments inherent in the V-MAGE evaluation process, certain levels may necessitate a considerable number of tokens during assessment. In this section, using the \textbf{Qwen2.5VL-72B} model deployed with \textbf{vLLM} as an illustrative example, we provide the statistically averaged frame counts (equal to the number of frames between two neighboring interactions multiplied by the number of interactions) and the average input and output token consumption per game round, serving as a reference. 
The detailed statistics are presented in Table~\ref{tab:SuperMario_avg_tokens}, \ref{tab:Race_avg_tokens}, \ref{tab:FlappyBird_avg_tokens}, and \ref{tab:TempestRun&Pong_avg_tokens}.

Depending on the differences in the models and the randomness of the games and reasoning, as well as other further experiments, the full research project may require \textbf{more} compute than the experiments reported here. The time of execution of the experiment depends on the network environment and computational power.

\begin{table*}[h!]
\centering
\caption{\textbf{SuperMario} Average Frames and Tokens Consumed}
\label{tab:SuperMario_avg_tokens}
\renewcommand{\arraystretch}{1.8}
\resizebox{\textwidth}{!}{
\begin{tabular}{@{}l | *{11}{c} @{}}
\toprule
Metric & Level 1 & Level 2 & Level 3 & Level 4 & Level 5 & Level 6 & Level 7 & Level 8 & Level 9 & Level 10 & All \\
\midrule
Average Frames & 400 & 655.56 & 1000 & 641.2 & 234.00 & 300 & 300 & 148.06 & 504.35 & 950.45 & 5133.62\\
Average Prompt Tokens & 150004.78 & 253457.38 & 379649.92 & 266125.09 & 100595.30 & 112241.57 & 111288.00 & 54231.22 & 192642.31 & 361208.32 & 1981443.89 \\
Average Completion Tokens & 10054.86 & 18437.53 & 25428.29 & 18242.65 & 7075.53 & 7639.30 & 7314.46 & 3710.36 & 13492.79 & 24197.61 & 135593.38 \\
\bottomrule
\end{tabular}
}
\end{table*}

\begin{table*}[h!]
\centering
\caption{\textbf{Race} Average Frames and Tokens Consumed}
\label{tab:Race_avg_tokens}
\renewcommand{\arraystretch}{1.8}
\resizebox{\textwidth}{!}{
\begin{tabular}{@{}l | *{10}{c} @{}}
\toprule

\multirow{2}{*}{Metric} & Level 1 & Level 2  & Level 3 & \multirow{2}{*}{Level 1} & \multirow{2}{*}{Level 2} & \multirow{2}{*}{Level 3} & \multirow{2}{*}{Level 4} & \multirow{2}{*}{Level 5} & \multirow{2}{*}{Level 6} & \multirow{2}{*}{All} \\

& No History & No History & No History & & & &
\\
\midrule
Average Frames & 12.66 & 15.39 & 16.66 & 29.20 & 30.69 & 32.14 & 58.07 & 98.06 & 32.46 & 325.33\\
Average Prompt Tokens & 1738.83 & 2309.85 & 2562.15 & 12317.22 & 14044.87 & 14934.32 & 31164.03 & 54346.81 & 17399.59 & 255136.23 \\
Average Completion Tokens & 275.35 & 531.41 & 595.91 & 693.11 & 937.66 & 1060.77 & 2243.55 & 3898.42 & 1517.55 & 20798.72\\
\bottomrule
\end{tabular}
}
\end{table*}

\begin{table*}[h!]
\centering
\caption{\textbf{FlappyBird} Average Frames and Tokens Consumed}
\label{tab:FlappyBird_avg_tokens}
\renewcommand{\arraystretch}{1.8}
\resizebox{\textwidth}{!}{
\begin{tabular}{@{}l | *{8}{c} @{}}
\toprule
Metric & Level 1 & Level 2 & Level 3 & Level 4 & Level 5 & Level 6 & Level 7 & All \\
\midrule
Average Frames & 224.73 & 133.34 & 76.49 & 153.11 & 153.87 & 152 & 143.28 & 1036.82 \\
Average Prompt Tokens & 98273.78 & 57332.59 & 32326.91 & 65853.39 & 66500.27 & 65322.27 & 56528.63 & 442137.84 \\
Average Completion Tokens & 9979.17 & 5772.39 & 3319.76 & 7142.32 & 7309.31 & 7082.73 & 6259.62 & 46865.30 \\
\bottomrule
\end{tabular}
}
\vspace{8pt}
\end{table*}

\begin{table*}[h!]
\centering
\caption{\textbf{TempestRun} and \textbf{PongGame} Average Frames and Tokens Consumed}
\label{tab:TempestRun&Pong_avg_tokens}
\renewcommand{\arraystretch}{1.8}
\resizebox{\textwidth}{!}{
\begin{tabular}{@{}l | *{5}{c} | *{4}{c} @{}}
\toprule
Metric & Level 1 & Level 2 & Level 3 & Level 4  & All & Level 1 & Level 2 & Level 3 & All\\
\midrule
Average Frames & 173.58 & 92.70 & 38.98 & 28.72 & 333.98 & 221.79 & 83.98 &  47.00 & 352.77\\
Average Prompt Tokens & 108291.56 & 57096.18 & 33218.84 & 22874.80 & 237820.07 & 136254.76 & 50056.53 & 26981.30 & 213292.59 \\
Average Completion Tokens & 7000.84 & 3799.28 & 2316.48 & 1619.33 &  15942.53 & 10998.40 & 4064.67 & 2208.39 & 17271.46 \\
\bottomrule
\end{tabular}
}
\end{table*}

\subsection{Inconsistency Between ELO and Performance Ratio Rankings}
\label{inconsistency analysis}

As shown in Table \ref{tab:model_performance} in the main text, ELO and Performance Ratio sometimes do not align in rankings.

We examine \textbf{Keye-VL-8B-Preview} and \textbf{Qwen2.5-VL-7B-Instruct}, with \textbf{LLaVA-v1.6-Mistral-7B} as a control. The experimental results are presented in Table~\ref{tab:elo_ratio}.

\begin{table*}[h!]
\centering
\caption{Elo Scores and Average Performance Ratios (E/R) Across Games.}
\label{tab:elo_ratio}
\resizebox{\textwidth}{!}{
\begin{tabular}{l|ccccc}
\hline
 & \textbf{Race(E/R)} & \textbf{SuperMario(E/R)} & \textbf{Pong(E/R)} & \textbf{FlappyBird(E/R)} & \textbf{TempestRun(E/R)} \\
\hline
Qwen2.5-VL-7B-Instruct & 1487/0.120 & 1459/0.239 & \textbf{1503/0.035} & 1431/0.030 & 1485/0.210 \\
Keye-VL-8B-Preview & 1487/0.118 & 1430/0.217 & \textbf{1495/0.039} & 1450/0.044 & 1513/0.239 \\
LLaVA-v1.6-Mistral-7B & 1462/0.051 & 1374/0.127 & \textbf{1494/0.035} & 1489/0.077 & 1379/0.169 \\
\hline
\end{tabular}
}
\end{table*}

In Pong, Qwen shows higher ELO but lower average ratio. We analyzed level-wise scores and variances to explore this. The variance is calculated as:
$ \text{variance} = \frac{\sum_{i=1}^{n}(\text{score}_i - \bar{\text{score}})^2}{n}$.
The detailed level-wise results and variance statistics are reported in Table~\ref{tab:pong_scores}.

\begin{table*}[h!]
\centering
\caption{Pong Scores by Level (Avg: average score, Var: variance).}
\label{tab:pong_scores}
\begin{tabular}{l|cc|cc|cc}
\hline
 & \textbf{L1 Avg} & \textbf{L1 Var} & \textbf{L2 Avg} & \textbf{L2 Var} & \textbf{L3 Avg} & \textbf{L3 Var} \\
\hline
Qwen2.5-VL-7B-Instruct & 0.48 & 0.50 & 0.37 & 0.25 & 0.20 & 0.18 \\
Keye-VL-8B-Preview & 0.68 & \textcolor{red}{0.67} & 0.26 & \textcolor{red}{0.33} & 0.23 & \textcolor{red}{0.36} \\
LLaVA-v1.6-Mistral-7B & 0.48 & 0.58 & 0.29 & 0.26 & 0.29 & 0.34 \\
\hline
\end{tabular}
\end{table*}

Keye's higher variance across all Pong levels indicates unstable performance, where high-scoring outliers mask frequent weak results. In the ELO system, this instability leads to more losses against a consistent opponent, resulting in a lower rating despite a competitive average score.

Additionally, current models perform poorly on Pong, with ratios tightly clustered in the 0–10\% range. When calculating the performance ratio by averaging across games, minor differences in Pong (3.5\% vs. 3.9\%) are overshadowed by larger gaps in other games(21\% vs. 24\%). 
The ELO system, in contrast, is based on the aggregate outcomes of all pairwise matchups. The ELO rating boost from a consistent pattern of wins in Pong is just as significant as from wins in any other game. 
This demonstrates that ELO is more robust in \textbf{fairly} assessing a model's holistic capabilities across tasks with imbalanced performance levels.

We also observed that in terms of Response Format Accuracy, GPT-4o is slightly lower than Gemini model (by 0.04\%), and InternVL2.5-78B is slightly lower than Qwen2-VL-72B (by 0.25\%). This may also be an influencing factor.

\newpage
\section{Games in V-MAGE}
\label{sec: Games in V-MAGE}

\subsection{Principles and Standards for Game Selection}
\label{Principles and Standards for Game Selection}

\textbf{Simplified and unrealistic considerations. } 
While the simplified visuals in these games differ from real-world scenes, empirical evidence demonstrate that MLLMs comprehend core game semantics (objectives, rules, entities) despite stylistic simplifications. 
Performance limitations primarily emerge from perceptual inaccuracies (e.g., dynamic object tracking) and multi-step reasoning deficiencies rather than misinterpretation. 
V-MAGE therefore focuses more on \textbf{precise evaluation} than \textbf{visual realism} to drive targeted improvements in visual reasoning.

\textbf{Selection criteria.}
The five games in V-MAGE share critical characteristics (e.g., non-textual state representation, free-form gameplay, and continuous-space environments) while offering diverse challenges. 

Our current minimal set covers four \textbf{2D} game types, as summarized in Table~\ref{tab:game_taxonomy}.

\begin{table}[ht]
\centering
\caption{2D Game Taxonomy in V-MAGE}
\label{tab:game_taxonomy}
\begin{tabularx}{\columnwidth}{lXX}
\toprule
 & \textbf{XY-axis} & \textbf{XZ-axis} \\
\midrule
\textbf{Linear Process} & PongGame & FlappyBird \\
\textbf{Open Planning} & RaceGame & SuperMario \\
\bottomrule
\end{tabularx}
\end{table}

The Linear Process implies that the game's progression is, to some extent, enforced. In PongGame, the ball's movement direction is determined by the game environment, requiring the model to move paddles on both sides to catch the ball, while in FlappyBird, the forward movement of the bird is compulsory, with the model controlling the height to navigate through pipes. OpenPlanning, in contrast, is relatively more open-ended. In RaceGame, the model can freely control the car's movement and direction on a plane to reach a trophy. In SuperMario, the model can move and jump in a relatively open environment to collect rewards and earn points.

For \textbf{3D} environments, we selected \textbf{Tempest Run} for its streamlined visual elements.

V-MAGE's flexible framework allows seamless integration of new PyGame-based environments. 
For instance, Tempest Run (one of our five games) was sourced from PyWeek[3], a community-driven game jam with thousands of open-source entries. This demonstrates our framework’s capacity to incorporate externally developed, human-designed games. We provide APIs to wrap new games into V-MAGE’s evaluation pipeline. This allows researchers to easily integrate additional games.

We will continue expanding the benchmark with more diverse titles that meet our selection criteria (e.g., Player vs Player (PVP) games) and will open-source both the codebase and detailed documentation to facilitate community contributions.

\subsection{Design and Implementation}
\label{appendix: game design and implementation}

As previously mentioned, V-MAGE enhances the diversity of the evaluation environment by expanding it through level design. Tables in this section detail the settings, rewards, and design objectives for each game's levels. For more comprehensive visual comparisons and prompt information, please refer to Appendix \ref{subsec: Games and Prompts}.

\textbf{Race Game }is a skill-based driving game where the objective is to control a car through a maze-like track to reach the trophy while avoiding obstacles. 
The car is represented by a red or white vehicle with a visible front and back, while the trophy is shown as a golden cup icon.
The surrounding white-lined boundaries represent walls, which the car must avoid.
For the overall observation and action spaces of the game, including the task and reward definitions, please refer to Table \ref{tab:race environment details}.

\begin{table*}[htbp]
\centering
\caption{Race Environment Details 
(* means potentially requires observation).}
\renewcommand{\arraystretch}{1.2}

\begin{tabular}{|l|l|l|l|}
\hline
Observation space & Action Space & Task & Reward \\
\hline
Car Position & UP, DOWN,  & Move the car  & +100 Success  \\
Trophy Position & LEFT, RIGHT & to reach the trophy & +0 Timeout  \\
Obstacle Position* & & &  +0  Destroyed\\
Speed* & & & \\
Acceleration* & & & \\
Facing Angle* & & & \\
\hline
\end{tabular}
\label{tab:race environment details}
\end{table*}

Each level in Race has a different set of rules and challenges. As presented in Table \ref{tab:race level configs}, we manually designed six levels. 
Levels 1–3 use a \textit{map-view perspective}("map" view), where models adjust absolute coordinates. The four types of movement operations directly translate the vehicle on the map according to the direction of action. Conversely, 
Levels 4--6 shift to a \textit{first-person perspective} ("car" view), where the observation is centered on the vehicle and movements are performed from the vehicle's perspective, requiring interpretation of velocity vectors and acceleration constraints.
Furthermore, acceleration is introduced in the high-difficulty levels, which further expands the observation space. This requires the model to extract more information from the visual input, including current speed and acceleration, in order to perform rational reasoning.

\begin{table*}[htbp]
\centering
\caption{Race Level Configurations}
\renewcommand{\arraystretch}{1.2}
\resizebox{\textwidth}{!}{

\begin{tabular}{|c|c|c|c|c|c|c|}
\hline
Level & View & bstacle & Initial Direction & Acceleration & Max Rounds & Sample Frames \\
\hline
1 & Map & No & - & No & 100 & 1 \\
\hline
2 & Map & Yes & - & No & 150 & 1 \\
\hline
3 & Map & Yes & - & No & 150 & 1 \\
\hline
4 & Car & No & Vertical(up) & Low & 150 & 3 \\
\hline
5 & Car & No & Horizontal(random) & Mid & 150 & 3 \\
\hline
6 & Car & Yes & Vertical(up) & Mid & 150 & 1 \\
\hline
\end{tabular}
}
\label{tab:race level configs}
\end{table*}

\textbf{SuperMario} is a two-dimensional side-scrolling platformer where the player controls the character Mario navigating through environments populated with various platforms, enemies, and obstacles. The goal is to traverse the level, collect coins, evade or defeat enemies, and reach the flagpole at the stage's conclusion. Players must avoid falling off platforms, colliding with enemies, or being struck by obstacles. Successful gameplay involves employing optimal movement combinations for smooth and efficient progression, alongside monitoring Mario's position relative to environmental elements. The task and reward definitions are shown in Table~\ref{tab:supermario environment details}.

\begin{table*}[htbp]
\centering
\caption{SuperMario Environment Details.}
\renewcommand{\arraystretch}{1.2}
\resizebox{\textwidth}{!}{

\begin{tabular}{|l|l|l|l|}
\hline
Observation space & Action Space & Task & Reward \\
\hline
Mario Position & UP, UP+LEFT,  & Collect coins and & +100 for collecting a coin \\
Platforms Position & UP+RIGHT, LEFT, & evade or defeat  & +100 for defeating a Goomba \\
Enemies Position & RIGHT, NONE & enemies & Penalties for falling or collisions \\
Obstacles Position & & & \\
\hline
\end{tabular}
}
\label{tab:supermario environment details}
\end{table*}

\begin{table*}[h!]
\centering
\vskip -10pt
\caption{SuperMario Level Configurations}
\renewcommand{\arraystretch}{1.2}
\resizebox{\textwidth}{!}{

\begin{tabular}{|c|c|c|c|c|c|}
\hline
Level & Enemy count & Coin Count & CoinBox Count & Max Rounds & Gameplay \\
\hline
1 & 0 & 6 & 2 & 400 & Common \\
\hline
2 & 2 & 6 & 2 & 1000 & Common \\
\hline
3 & 0 & 17 & 4 & 1000 & Long History (Two ways) \\
\hline
4 & 2 & 17 & 4 & 1000 & Long History (Two ways) \\
\hline
5 & 3 & 8 & 0 & 300 & Left or Right \\
\hline
6 & 0 & 13 & 0 & 300 & Left or Right \\
\hline
7 & 0 & 8 & 0 & 300 & Left or Right \\
\hline
8 & 0 & 12 & 0 & 1000 & Jump Only \\
\hline
9 & 5 & 8 & 0 & 1000 & Jump and Enemy \\
\hline
10 & 12 & 0 & 9 & 5000 & Classic W1-1 \\
\hline
\end{tabular}
}
\label{tab:supermario level configs}
\end{table*}

To provide a comprehensive evaluation of MLLMs' visual reasoning and planning capabilities, SuperMario features ten levels with configurations detailed in Table \ref{tab:supermario level configs}. These levels vary in enemy count, coin and coinbox quantities, maximum allowed rounds, and specific gameplay mechanics or focuses. Of these, level 10 serves as a standard human-difficulty benchmark, providing a 1:1 replica of the original Super Mario game's World 1-1 stage.

\textbf{Flappy Bird }is a widely recognized side-scrolling mobile game serving as a common benchmark in reinforcement learning. The objective is to control a bird's vertical movement to navigate through a continuous series of horizontal gaps within vertically oriented pipes. Successful traversal of a pipe pair increments the player's score, while collision with any pipe or the ground constitutes a terminal state, ending the game. The game mechanic involves a constant downward gravitational pull, counteracted by discrete upward 'flaps' initiated by the player.
The task and reward definitions are shown in Table~\ref{tab:flappybird environment details}.

\begin{table*}[htbp]
\centering
\caption{Flappy Bird Environment Details (* means only available at certain levels).}
\renewcommand{\arraystretch}{1.2}
\resizebox{\textwidth}{!}{

\begin{tabular}{|l|l|l|l|}
\hline
Observation space & Action Space & Task & Reward \\
\hline
Bird Position & UP & Maneuver the bird to avoid & +1 per pipe pair passed \\
Bird Velocity & NONE &  hitting the pipes &  +0 Collision \\
Next Pipe Distance & DOWN* & & \\
Gap VerticalPosition & KEEP* & & \\
\hline
\end{tabular}
}
\label{tab:flappybird environment details}
\end{table*}

Given the high difficulty of human-standard levels for MLLMs, we designed seven levels with progressive difficulty. Specifically, as presented in Table \ref{tab:flappybird level configs}, levels 1-3 constitute a simplified game environment where the gravity factor is removed, and height is controlled via UP and DOWN actions to navigate through the pipes. Levels 4-6 are based on the standard difficulty but incorporate a 'KEEP' option, enabling the model to maintain the bird's altitude through this action. Within the same difficulty tier, levels are differentiated by varying the bird's forward speed and the pipe gap width. Level 7 represents the standard human game difficulty, retaining the original game settings.

\begin{table*}[htbp]
\centering
\caption{FlappyBird Level Configurations}
\renewcommand{\arraystretch}{1.2}
\resizebox{\textwidth}{!}{

\begin{tabular}{|c|c|c|c|c|}
\hline
Level & Gravity & Availability of "DOWN" & Availability of "KEEP" & Others \\
\hline
1 - 3 & No & Yes & Yes & Distinguished by gap\\
\cline{1-4} 
4 - 6 & Yes & No & Yes & clearance and speed  \\
\hline
7 & Yes & No & No & Human Standard \\
\hline
\end{tabular}
}
\label{tab:flappybird level configs}
\end{table*}

\textbf{Pong} Game is a classic two-player adversarial game. The objective is to control the paddles on the left and right sides of the screen to return the ball, preventing it from passing one's own paddle while simultaneously attempting to make the ball pass the opponent's paddle. One point is awarded to the player for each successful return of the ball. The final score is the sum of both players' scores. The task and reward definitions are shown in Table~\ref{tab:pong environment details}.

\begin{table*}[htbp]
\centering
\caption{Pong Game Environment Details.}
\renewcommand{\arraystretch}{1.2}
\resizebox{\textwidth}{!}{
\begin{tabular}{|l|l|l|l|}
\hline
Observation space & Action Space & Task & Reward \\
\hline
Left Paddle Position & LEFTUP & Track the ball's trajectory & +1 per successful  hit \\
Right Paddle Position & LEFTDOWN & and maneuver the left and  & +0 if ball passes paddle \\
Ball Position & RIGHTUP &  right paddles to intercept  & \\
Ball Trajectory &RIGHTDOWN & and return the ball. & \\
 & NONE & &  \\
\hline
\end{tabular}
}
\label{tab:pong environment details}

\end{table*}

\begin{table*}[h!]
\centering
\caption{Pong Game Level Configurations}
\renewcommand{\arraystretch}{1.2}
\begin{tabular}{|c|c|c|c|c|}
\hline
Level & Paddle Width & Ball Speed & Ball Size & Others \\
\hline
1 & Big & Slow & Big & Ball initial position\\
\cline{1-4} 
2 & Mid & Mid & Mid &  randomly changes. \\
\cline{1-4} 
3 & Small & Fast & Small & \\
\hline
\end{tabular}
\label{tab:pong level configs}
\end{table*}

Considering the challenges MLLMs face in tracking and temporal tasks, we designed levels with varying difficulty. As shown in Table \ref{tab:pong level configs}, difficulty for Levels 1-3 is differentiated by adjusting the paddle width and the speed of the ball. Within the same level, the initial position of the ball is randomized, but the relative difficulty remains consistent.

\begin{table*}[h!]
\centering
\caption{Tempest Run Environment Details.}
\renewcommand{\arraystretch}{1.2}
\resizebox{\textwidth}{!}{
\begin{tabular}{|l|l|l|l|}
\hline
Observation space & Action Space & Task & Reward \\
\hline
Current Character State & JUMP, LEFT, RIGHT, & Perform corresponding & Score increases \\
Nearby Obstacles Position & SLIDE, RISE, NONE & actions to avoid or  &  with distance run. \\
Nearby Obstacles Type & & destroy obstacles. & \\
Visual Information Quantity & & & \\
\hline
\end{tabular}
}
\label{tab:tempestrun environment details}
\end{table*}

\textbf{Tempest Run} is a third-person perspective 3D runner game where the player controls a character moving within a futuristic tunnel filled with various obstacles and enemies. The objective is to navigate through the tunnel, avoiding or overcoming impediments, and to run as far as possible. Players must specifically avoid colliding with red spikes, purple walls, or failing to manage green enemies. Successful gameplay requires employing optimal combinations of movements for smooth and efficient tunnel traversal, alongside monitoring the character's position relative to obstacles and reacting appropriately. The task and reward definitions are shown in Table~\ref{tab:tempestrun environment details}.

To evaluate MLLMs' visual comprehension and reactive capabilities within a dynamic 3D environment, Tempest Run includes four levels of varying difficulty. As outlined in Table \ref{tab:tempestrun level configs}, Levels 1-4 are primarily differentiated by parameters including role speed, cell length (denoting the distance between environmental segments), and random rate (controlling obstacle spawning frequency). These parameters collectively influence the pace of barrier generation and the overall visual complexity of the tunnel environment, thereby varying the level of challenge. Within the same level, the positioning of environmental elements is randomized, while maintaining consistent relative difficulty. 

\begin{table*}[htbp]
\centering
\caption{Tempest Run Level Configurations}
\renewcommand{\arraystretch}{1.2}
\begin{tabular}{|c|c|c|c|l|}
\hline
Level & Role Speed & Cell Length & Random Rate & Others \\
\hline
1 & Slow & Large & Low & Environmental elements initial\\
\cline{1-4}
2 & Medium & Medium & Medium-Low & positions randomly change.\\
\cline{1-4}
3 & Fast & Small & Medium-High & \\
\cline{1-4}
4 & Very Fast & Small & High & \\
\hline
\end{tabular}
\label{tab:tempestrun level configs}
\end{table*}


\subsection{Original Sources}
\label{subsec: sources}

Thanks to the open-source community, we are able to leverage existing game codebases to build our benchmark. 
The codebases used in this work are summarized in Table~\ref{tab:game_sources}.
Most of these projects are licensed under the MIT License; for those without an explicit license, we have obtained formal permission from the original authors for their use in this research.

\begin{table*}[htbp]
\small
\centering
\caption{Game Codebase Sources}
\label{tab:game_sources}
\renewcommand{\arraystretch}{1.2}
\begin{tabular}{l|l}
\toprule
\textbf{Game} & \textbf{Codebase} \\
\midrule
\textbf{Race} & \url{https://github.com/tdostilio/Race_Game} \\
\midrule
\textbf{FlappyBird} & \url{https://github.com/agneay/pygame-projects/tree/master/Flappy\%20Bird} \\
\midrule
\textbf{Pong} & \url{https://github.com/pyGuru123/Python-Games/tree/master/Pong} \\
\midrule
\textbf{SuperMario} & \url{https://github.com/mx0c/super-mario-python} \\
\midrule
\textbf{Tempest Run} & \url{https://github.com/davidpendergast/pygame-summer-team-jam} \\
\bottomrule
\end{tabular}

\end{table*}

In most cases, the original codebases lacked comprehensive difficulty settings and level designs suitable for systematic evaluation. We therefore modified the default human-oriented game configurations to adapt them for benchmarking purposes, while meticulously designing a diverse set of challenging levels to ensure rigorous assessment.

\newpage
\subsection{Games and Prompts}
\label{subsec: Games and Prompts}

\begin{figure*}[htbp]
\vskip 0.2in
\begin{center}
\centerline{\includegraphics[width=0.9 \textwidth]{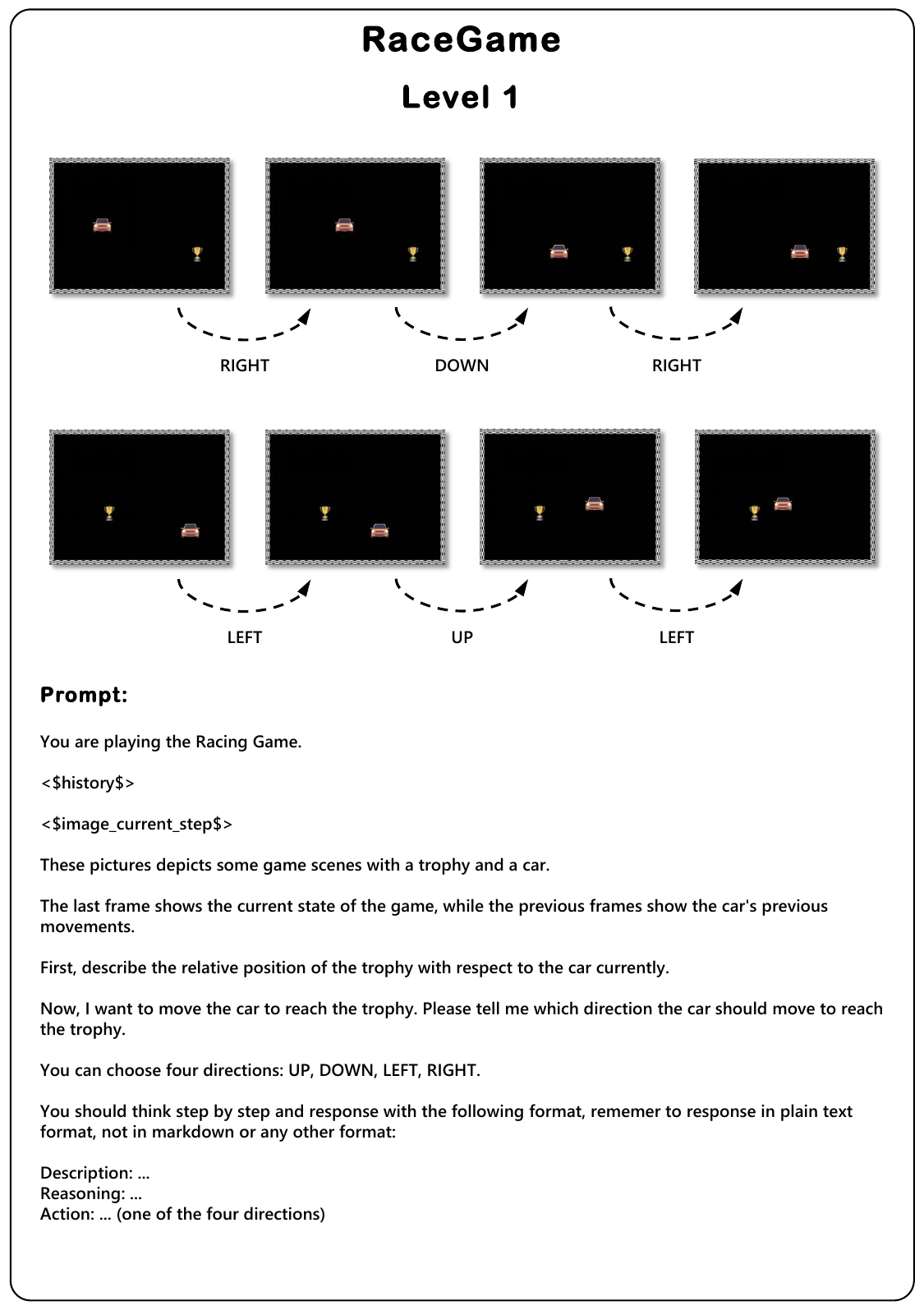}}
\caption{
\textbf{RaceGame Level 1: Level Design and Prompt Overview.} The images showcase the scene from Level 1, illustrating the level design and corresponding prompt. Elements in the same level will randomly change their initial positions while maintaining consistent relative difficulty.}
\label{fig:race1}
\end{center}
\vskip -0.2in
\end{figure*}

\begin{figure*}[htbp]
\vskip 0.2in
\begin{center}
\centerline{\includegraphics[width=0.95 \textwidth]{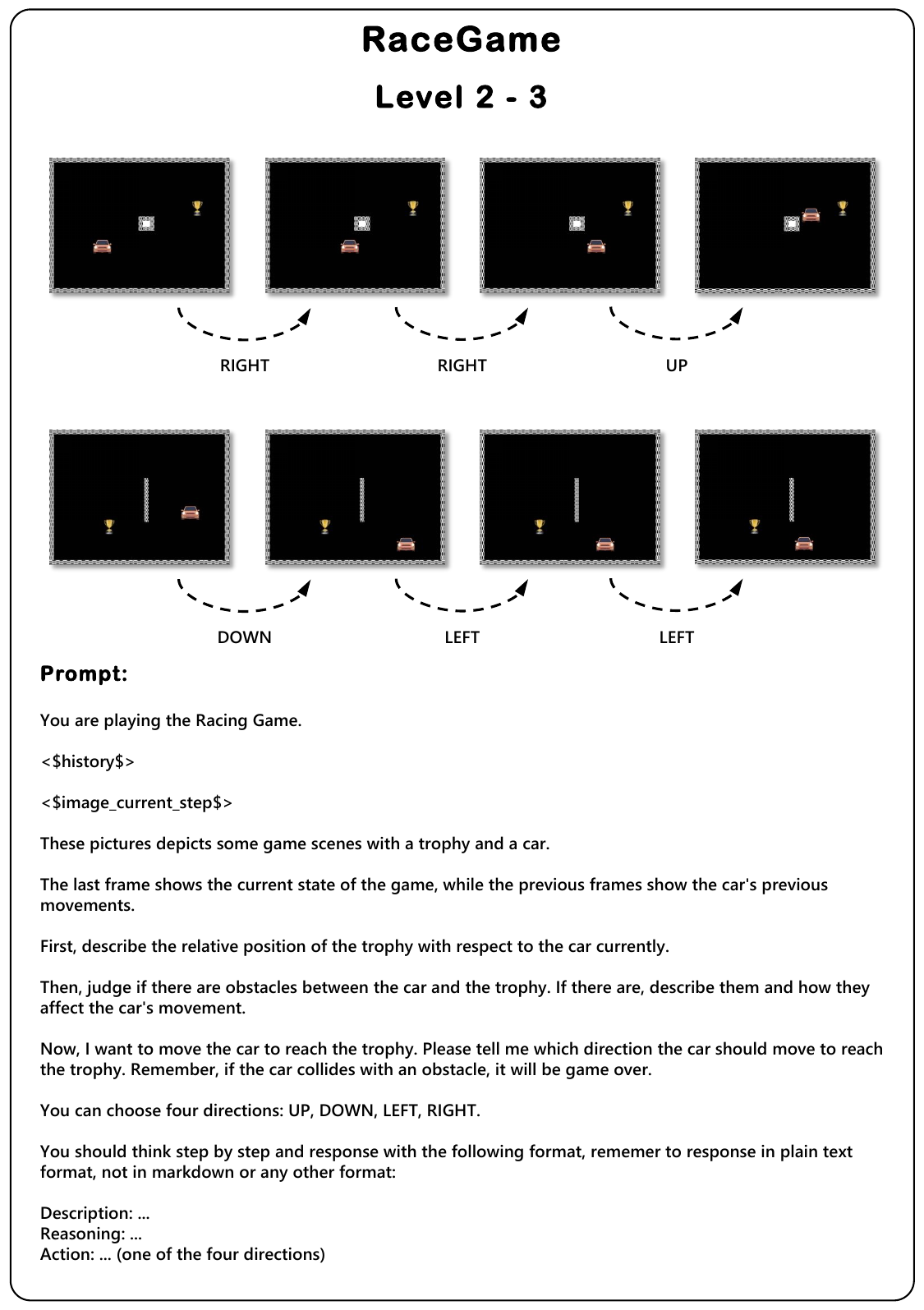}}
\caption{\textbf{RaceGame Level 2-3: Level Design and Prompt Overview.} The images showcase the scene from Level 2-3, illustrating the level design and corresponding prompt. Elements in the scene will randomly change their initial positions while maintaining consistent relative difficulty.}
\label{fig:race2}
\end{center}
\vskip -0.2in
\end{figure*}

\begin{figure*}[htbp]
\vskip 0.2in
\begin{center}
\centerline{\includegraphics[width=0.95 \textwidth]{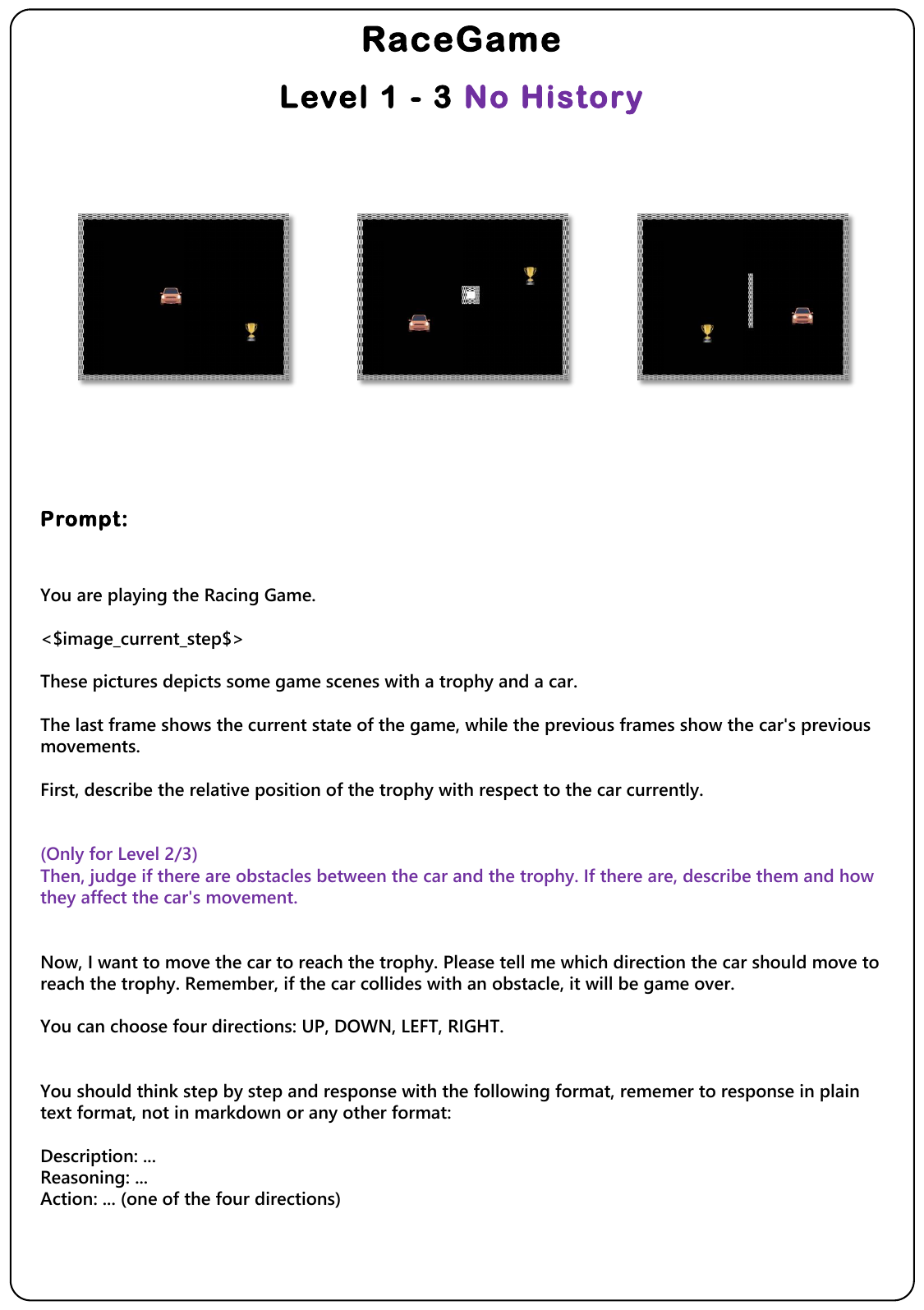}}
\caption{\textbf{RaceGame Level 1-3 No History: Level Design and Prompt Overview.} The images showcase the scene from Level 1-3 No History, illustrating the level design and corresponding prompt. Elements in the scene will randomly change their initial positions while maintaining consistent relative difficulty. Same as the original levels except the input sequence has been changed to the single image.}
\label{fig:race_no_his}
\end{center}
\vskip -0.2in
\end{figure*}

\begin{figure*}[htbp]
\vskip 0.2in
\begin{center}
\centerline{\includegraphics[width=0.95 \textwidth]{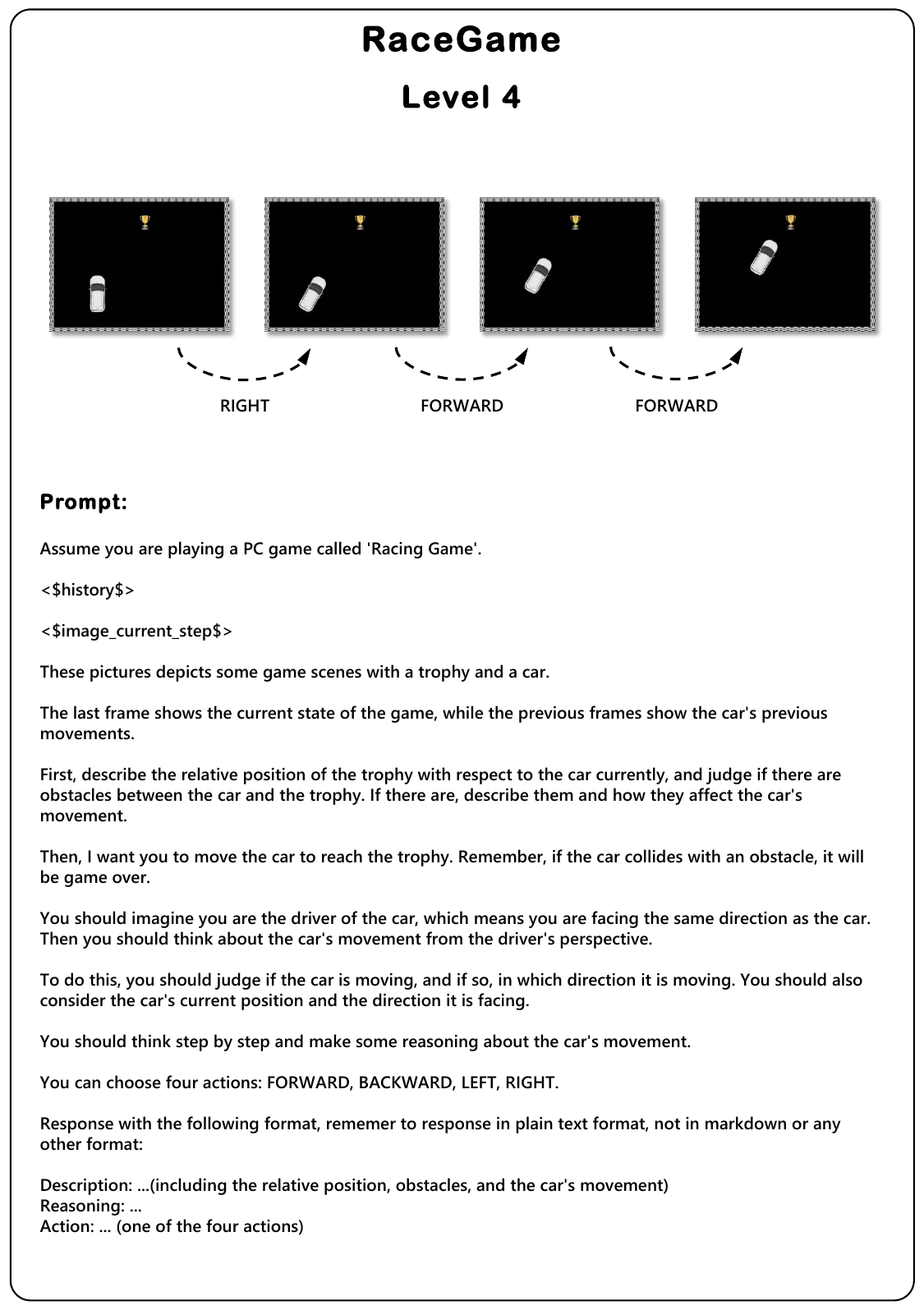}}
\caption{\textbf{RaceGame Level 4: Level Design and Prompt Overview.} The images showcase the scene from Level 4, illustrating the level design and corresponding prompt. Elements in the same level will randomly change their initial positions while maintaining consistent relative difficulty.}
\label{fig:race3}
\end{center}
\vskip -0.2in
\end{figure*}

\begin{figure*}[htbp]
\vskip 0.2in
\begin{center}
\centerline{\includegraphics[width=0.95 \textwidth]{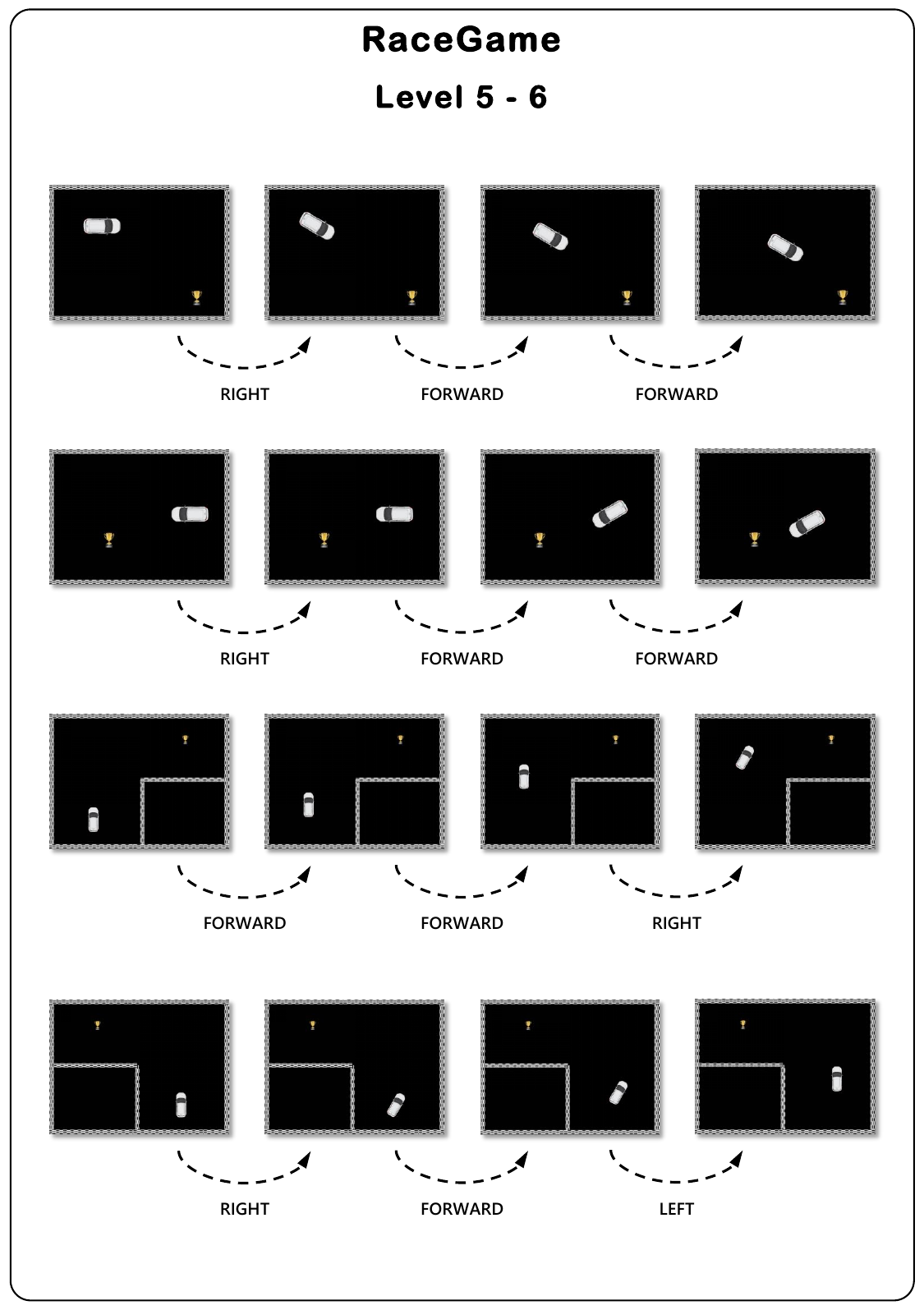}}
\caption{\textbf{RaceGame Level 5-6: Level Design and Prompt Overview.} The images showcase the scene from Level 5-6, illustrating the level design and corresponding prompt. Elements in the same level will randomly change their initial positions while maintaining consistent relative difficulty. \textbf{The prompt is the same as in Level 4.}}
\label{fig:race4}
\end{center}
\vskip -0.2in
\end{figure*}

\begin{figure*}[htbp]
\vskip 0.2in
\begin{center}
\centerline{\includegraphics[width=0.95 \textwidth]{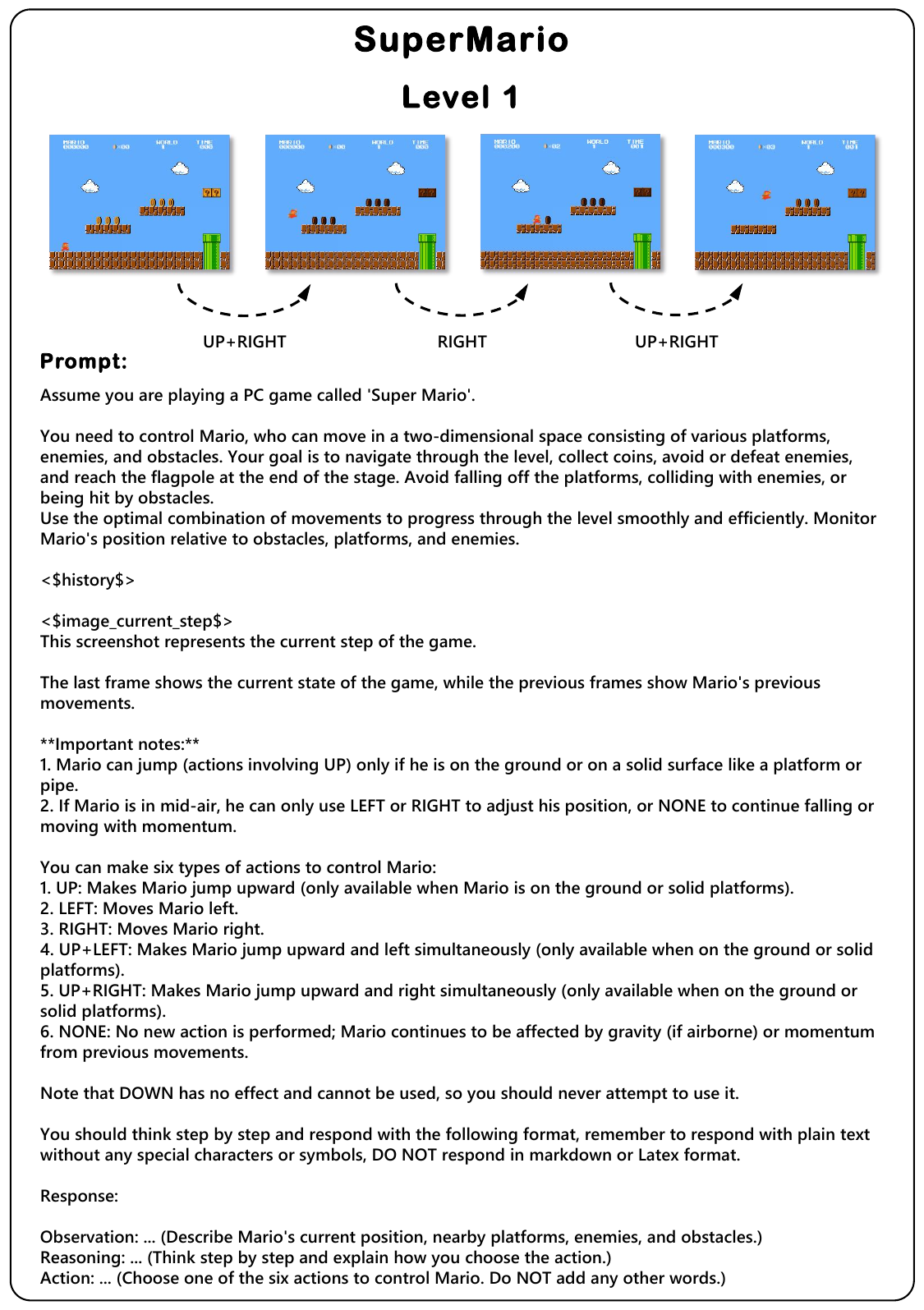}}
\caption{\textbf{SuperMario Level 1: Level Design and Prompt Overview.} The images showcase the scene from Level 1, illustrating the level design and corresponding prompt.}
\label{fig:mario1}
\end{center}
\vskip -0.2in
\end{figure*}

\begin{figure*}[htbp]
\vskip 0.2in
\begin{center}
\centerline{\includegraphics[width=0.95 \textwidth]{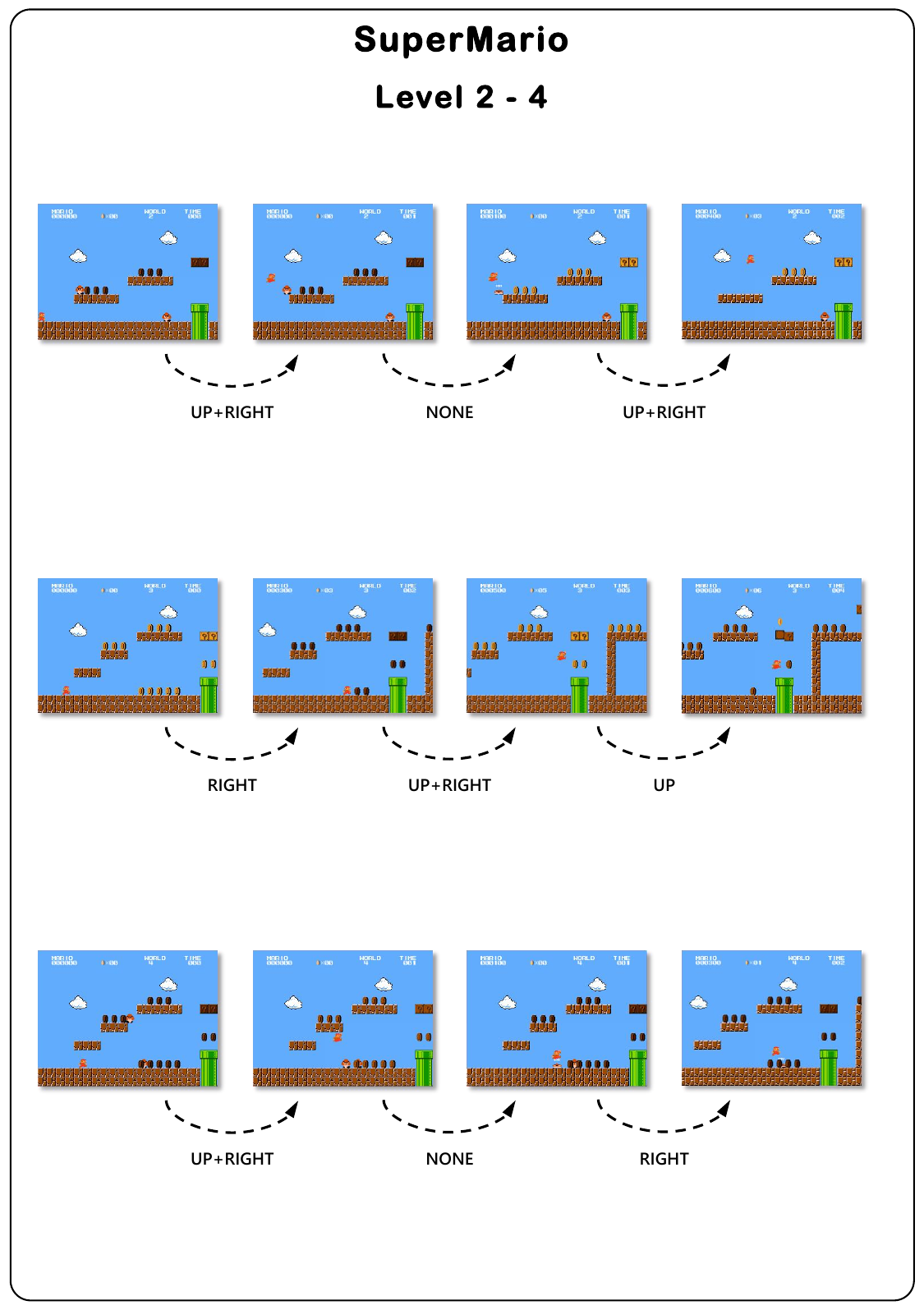}}
\caption{\textbf{SuperMario Level 2-4: Level Design and Prompt Overview.} The images showcase the scene from Level 2-4, illustrating the level design and corresponding prompt.\textbf{The prompt is the same as in Level 4.}}
\label{fig:mario2}
\end{center}
\vskip -0.2in
\end{figure*}

\begin{figure*}[htbp]
\vskip 0.2in
\begin{center}
\centerline{\includegraphics[width=0.95 \textwidth]{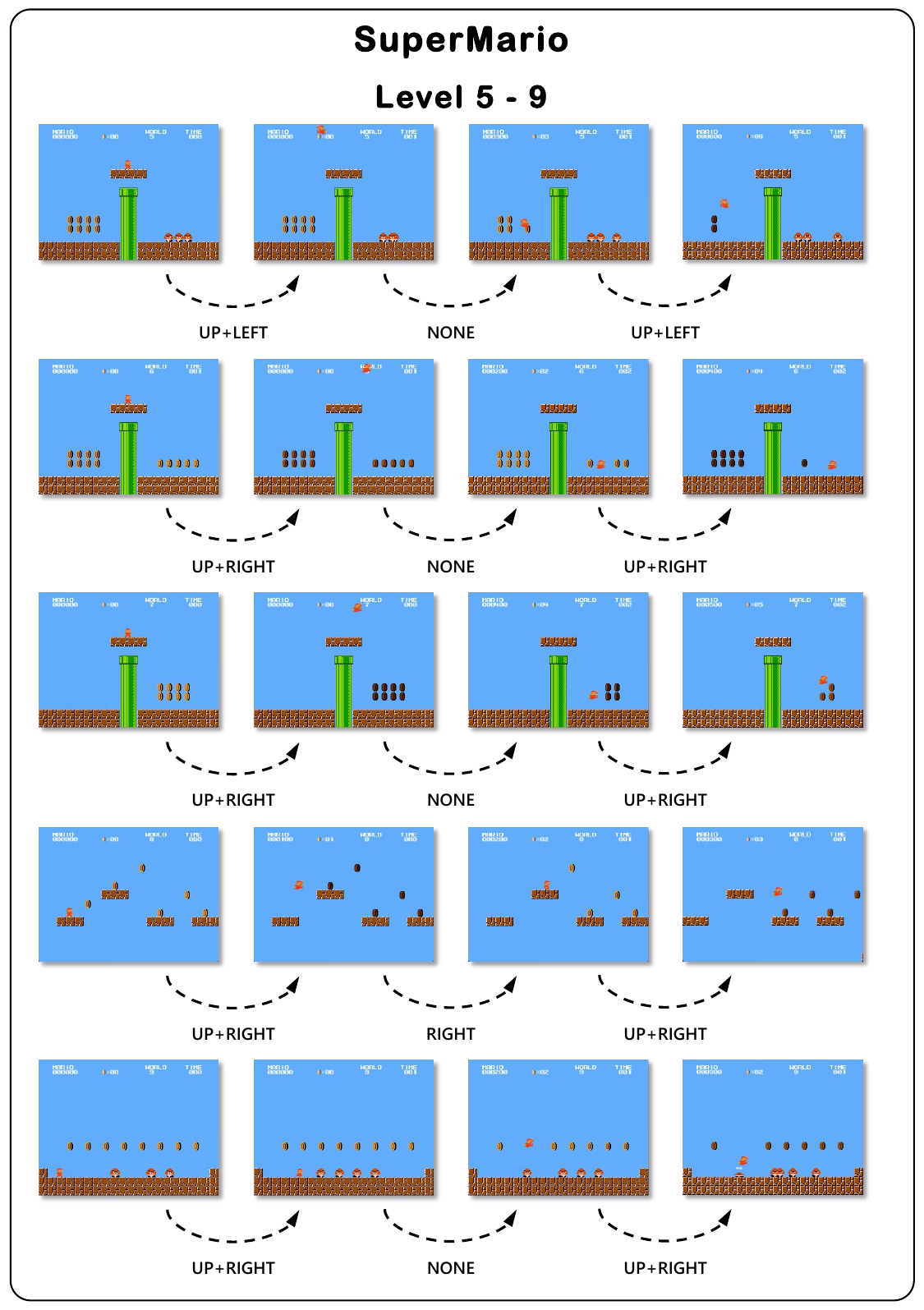}}
\caption{\textbf{SuperMario Level 5-9: Level Design and Prompt Overview.} The images showcase the scene from Level 5-9, illustrating the level design and corresponding prompt.\textbf{The prompt is the same as in Level 4.}}
\label{fig:mario3}
\end{center}
\vskip -0.2in
\end{figure*}

\begin{figure*}[htbp]
\vskip 0.2in
\begin{center}
\centerline{\includegraphics[width=0.95 \textwidth]{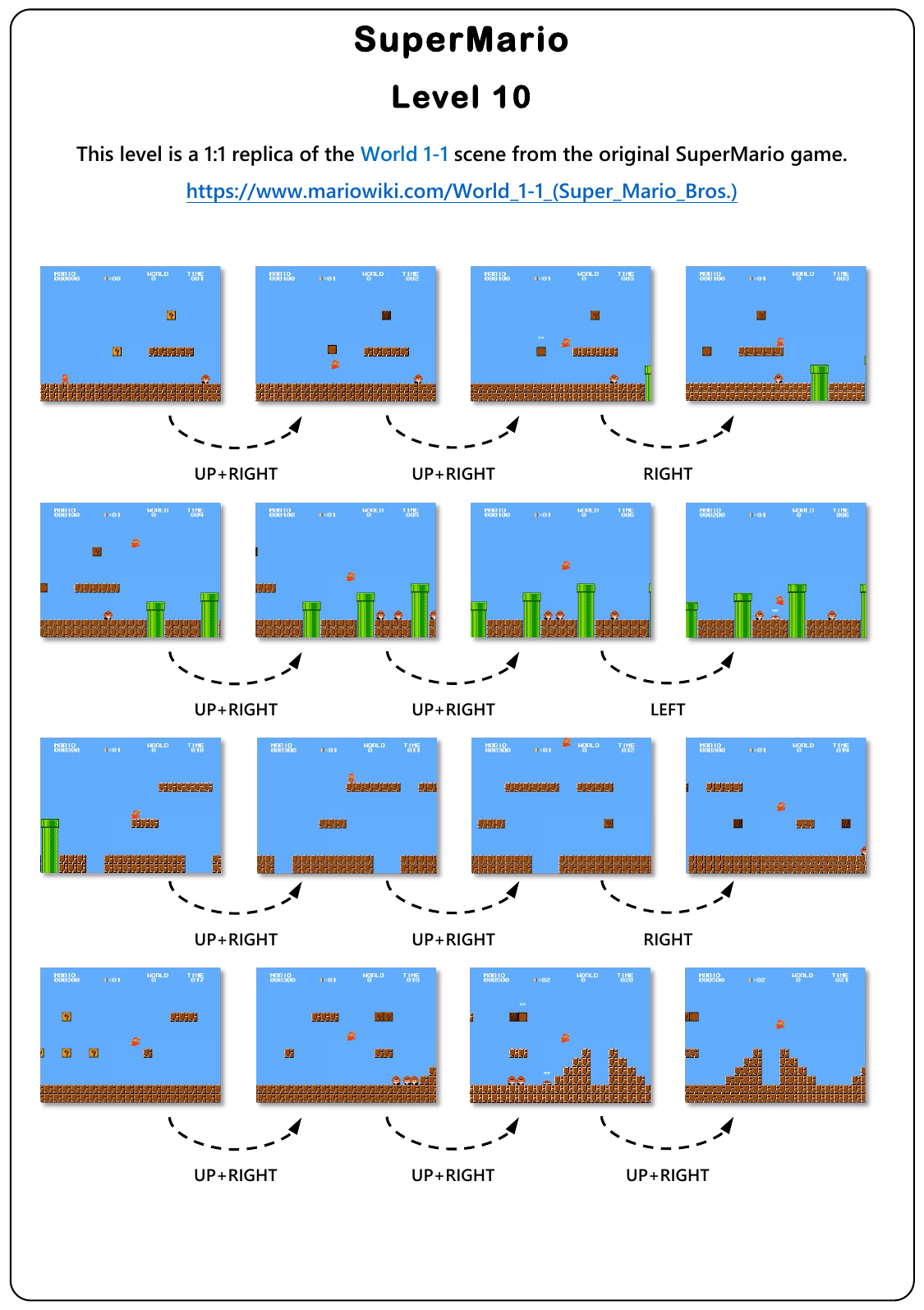}}
\caption{\textbf{SuperMario Level 10 (Standard Level): Level Design and Prompt Overview.} The images showcase the scene from Level 10, illustrating the level design and corresponding prompt. This is The standard level that matches the difficulty of the human game. \textbf{The prompt is the same as in Level 4.}}
\label{fig:mario4}
\end{center}
\vskip -0.2in
\end{figure*}

\begin{figure*}[htbp]
\vskip 0.2in
\begin{center}
\centerline{\includegraphics[width=0.95 \textwidth]{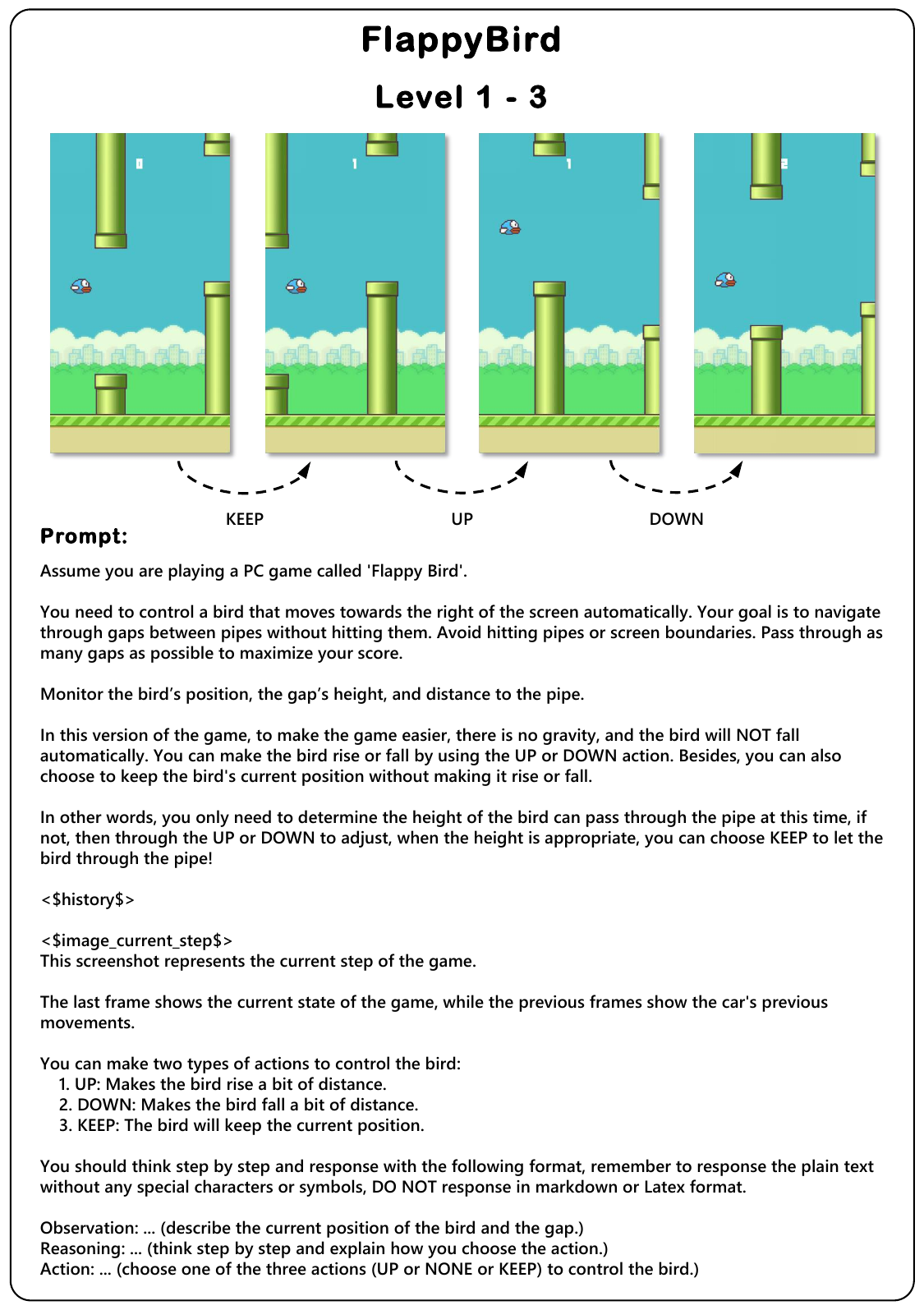}}
\caption{\textbf{FlappyBird Level 1-3: Level Design and Prompt Overview.} The images showcase the scene from Level 1, illustrating the level design and corresponding prompt. Levels are differentiated by the pipe gap width and the bird's forward speed. Elements in the same level will randomly change their initial positions while maintaining consistent relative difficulty.}
\label{fig:flappybird1}
\end{center}
\vskip -0.2in
\end{figure*}

\begin{figure*}[htbp]
\vskip 0.2in
\begin{center}
\centerline{\includegraphics[width=0.95 \textwidth]{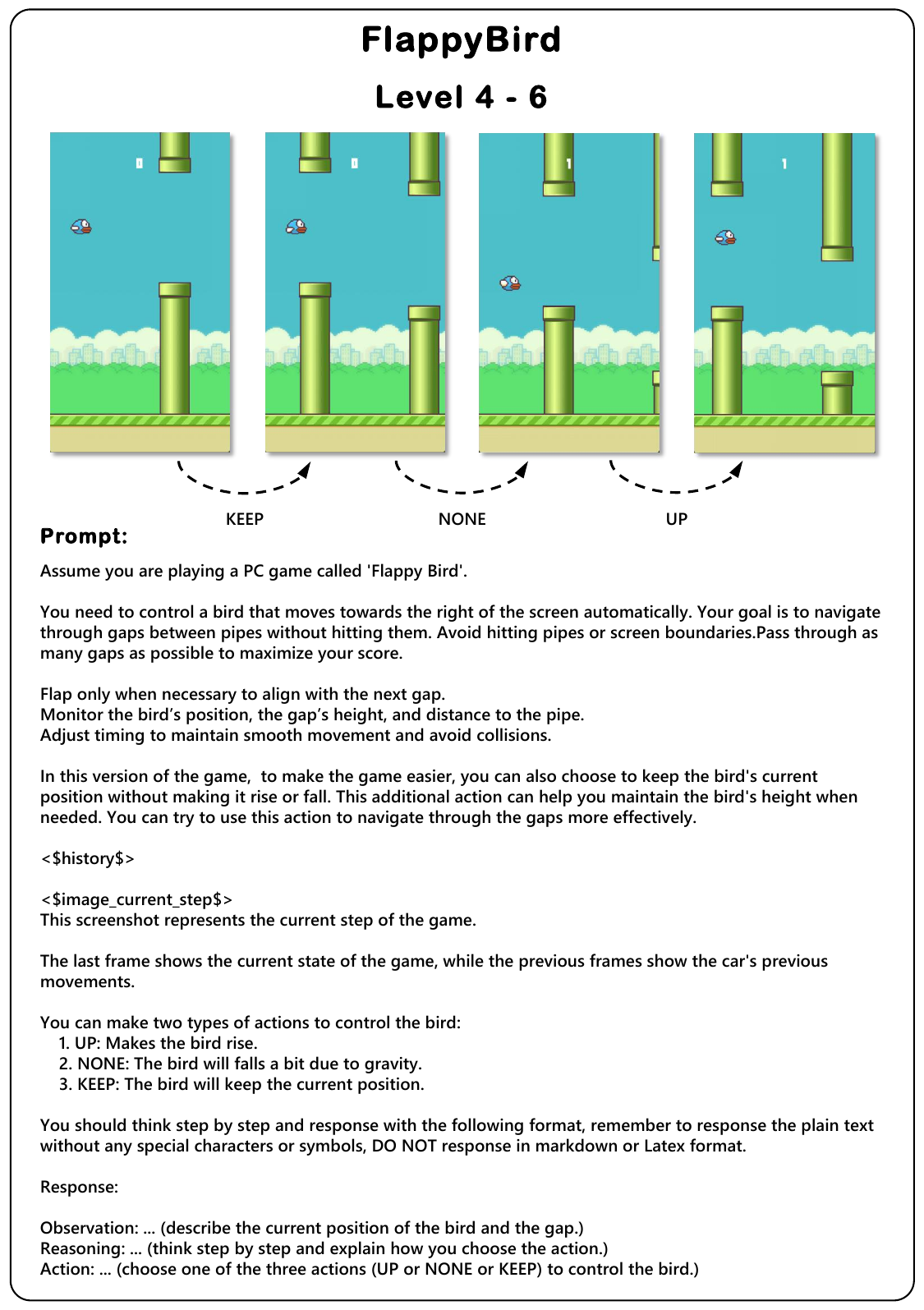}}
\caption{\textbf{FlappyBird Level 4-6: Level Design and Prompt Overview.} The images showcase the scene from Level 4, illustrating the level design and corresponding prompt. Levels are differentiated by the pipe gap width and the bird's forward speed. Elements in the same level will randomly change their initial positions while maintaining consistent relative difficulty.}
\label{fig:flappybird2}
\end{center}
\vskip -0.2in
\end{figure*}

\begin{figure*}[htbp]
\vskip 0.2in
\begin{center}
\centerline{\includegraphics[width=0.95 \textwidth]{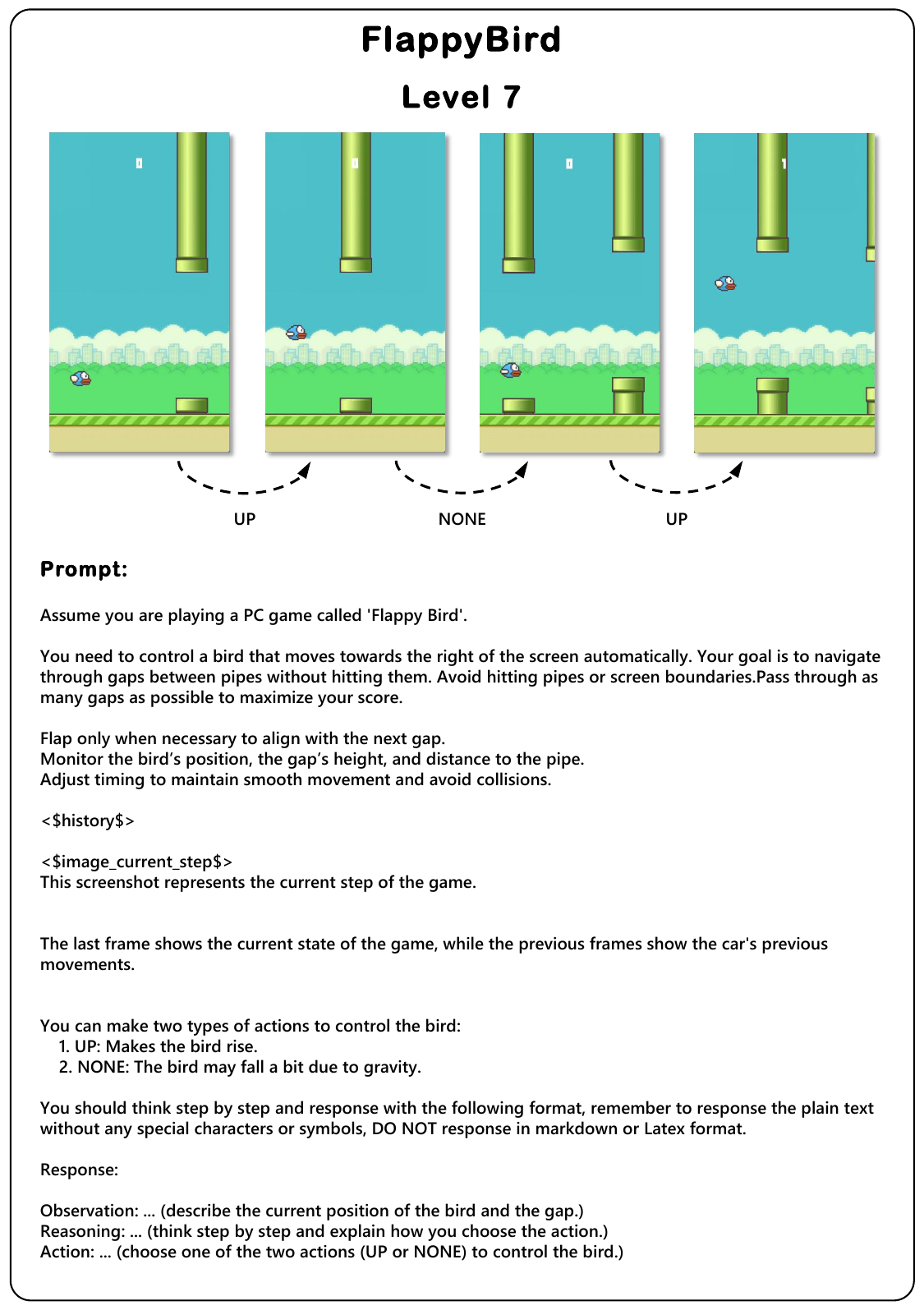}}
\caption{\textbf{FlappyBird Level 7 (Standard Level): Level Design and Prompt Overview.} The images showcase the scene from Level 7, illustrating the level design and corresponding prompt. Elements in the same level will randomly change their initial positions while maintaining consistent relative difficulty. This is The standard level that matches the difficulty of the human game.}
\label{fig:flappybird3}
\end{center}
\vskip -0.2in
\end{figure*}

\begin{figure*}[htbp]
\vskip 0.2in
\begin{center}
\centerline{\includegraphics[width=0.95 \textwidth]{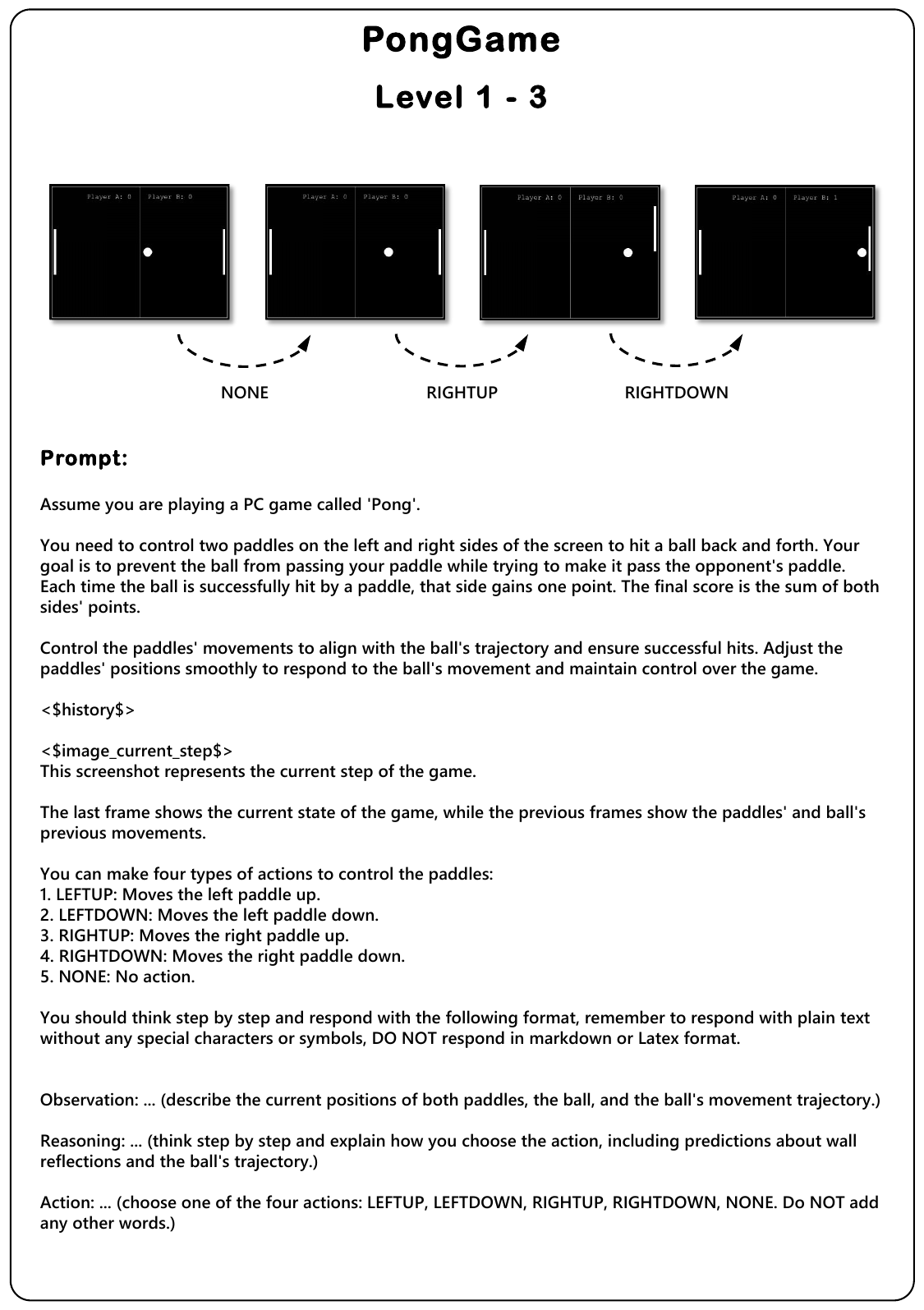}}
\caption{\textbf{PongGame Level 1-3: Level Design and Prompt Overview.} The images showcase the scene from Level 1, illustrating the level design and corresponding prompt. Levels are differentiated by the paddle width and the ping pong ball's speed. The ping pong ball in the same level will randomly change its initial position while maintaining consistent relative difficulty.}
\label{fig:pong1}
\end{center}
\vskip -0.2in
\end{figure*}

\begin{figure*}[htbp]
\vskip 0.2in
\begin{center}
\centerline{\includegraphics[width=0.95 \textwidth]{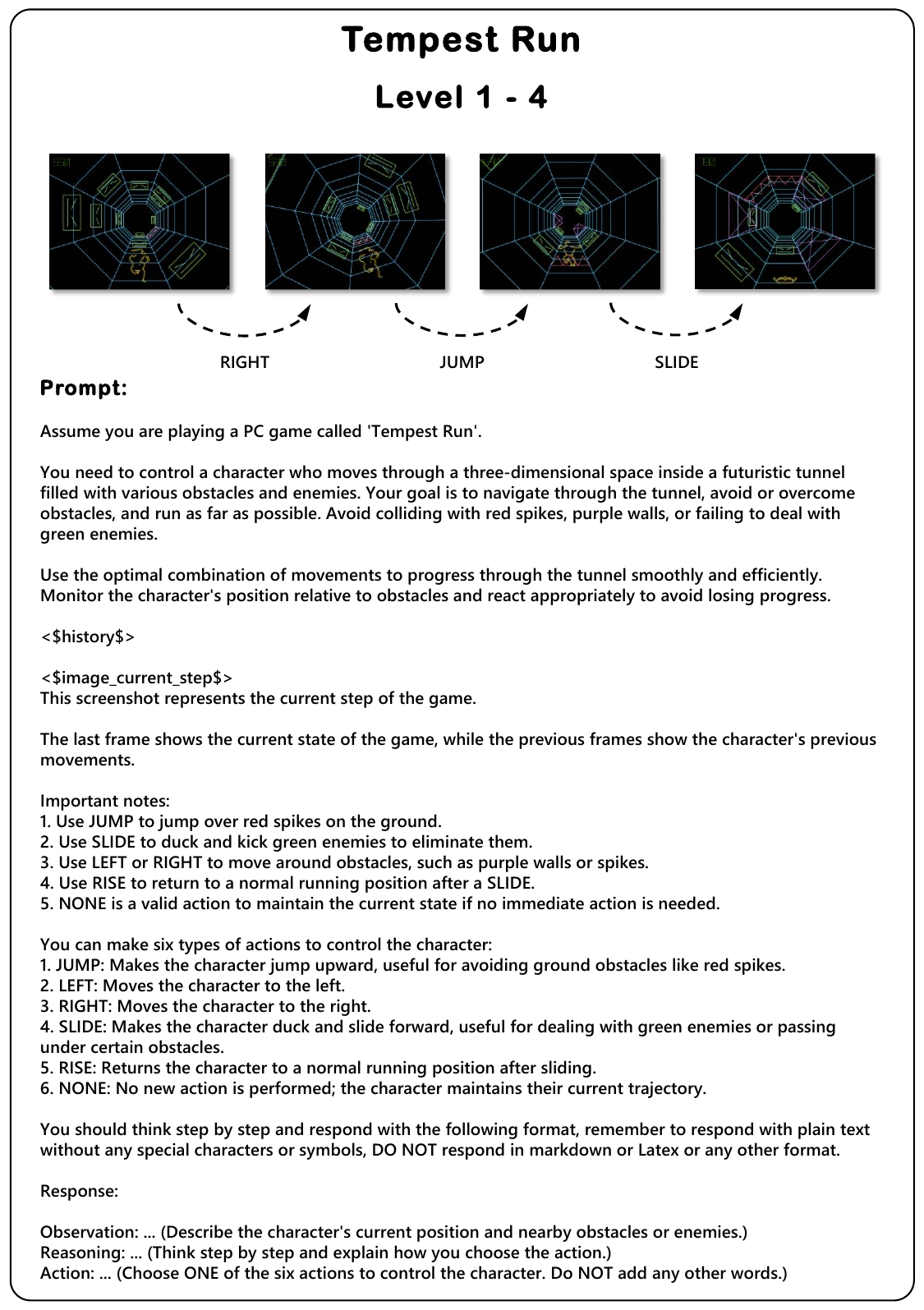}}
\caption{\textbf{Tempest Run Level 1-4: Level Design and Prompt Overview.} The images showcase the scene from Level 1, illustrating the level design and corresponding prompt. Levels are differentiated by the speed of barrier generation and the amount of visual information available. Elements in the same level will randomly change their initial positions while maintaining consistent relative difficulty.}
\label{fig:tem1}
\end{center}
\vskip -0.2in
\end{figure*}

All the games have been modified based on publicly available code. 
The detailed level designs and corresponding prompts for all games are illustrated in 
Figures~\ref{fig:race1}--\ref{fig:race4}, 
\ref{fig:mario1}--\ref{fig:mario4}, 
\ref{fig:flappybird1}--\ref{fig:flappybird3}, 
\ref{fig:pong1}, and 
\ref{fig:tem1}.



\newpage

\clearpage
\section{Elo Performance Comparison Protocol Details}
\label{sec: elo details}

The core methodology for metrics evaluation in V-MAGE comprises two interconnected components: \textit{performance comparison} and \textit{statistical stabilization}. 
\newline

\textbf{Performance Comparison Protocol.}  

Each model begins with an initial Elo rating $R_m^{(0)} = 1500$, where $m \in \mathcal{M}$ represents the set of candidate models. 
We conducted 100 evaluation rounds for each game level $\ell$ where models were randomly paired in each round through a shuffle operation:

\begin{equation}
\mathcal{P}_t = \{(A_t,B_t) \,|\, A_t, B_t \stackrel{\text{rand}}{\sim} \mathcal{M}^\ell, A_t \neq B_t\}
\end{equation}

where $A_t$ and $B_t$ denote the paired models in round $t$. 

For paired models $(A,B)$, their game scores $\text{score}_A$ and $\text{score}_B$ are compared next. We first construct 
\begin{equation}
f(m) = (\text{score}_m, \text{valid\_rate}_m)
\end{equation}
where $\text{valid\_rate}_m$ represents the proportion of actions output by model $m$ in this game round that conform to the formatting requirements. 
The outcome $(S_A, S_B) \in \{(1,0), (0,1), (0.5,0.5)\}$ is determined by:

\begin{equation}
(S_A, S_B) = 
\begin{cases}
(1, 0) & \text{if } f(A) \succ f(B) \\
(0, 1) & \text{if } f(A) \prec f(B) \\
(0.5, 0.5) & \text{otherwise}
\end{cases}
\end{equation}

The rating update mechanism follows the classical Elo formulation with enhanced stability controls. For models $A$ and $B$ with pre-match ratings $R_A$ and $R_B$, their expected performance is calculated as:
{\small
\begin{equation}
E_A = \frac{1}{1 + 10^{(R_B - R_A)/400}}, \quad E_B = \frac{1}{1 + 10^{(R_A - R_B)/400}}
\end{equation}
}
where the denominator base 10 and scaling factor 400 establish a logarithmic relationship between rating differences and expected outcomes. The post-match ratings become:

\begin{equation}
\begin{aligned}
R'_A &= R_A + K(S_A - E_A) \\
R'_B &= R_B + K(S_B - E_B)
\end{aligned}
\end{equation}

where \(K\) is a constant determining the sensitivity of the rating system (typically set to 32),
\newline

\textbf{Stabilization through Randomized Iteration.}  

To ensure the robustness of rating updates, V-MAGE implements multi-pass stabilization protocol. All historical comparisons are aggregated into a win/loss pool:

\begin{equation}
\mathcal{W} = \bigcup_{g\in\mathcal{G}} \bigcup_{\ell\in\mathcal{L}_g} \bigcup_{t=1}^N (A_{g\ell t}, B_{g\ell t}, S_A^{g\ell t}, S_B^{g\ell t})
\end{equation}

which undergoes $T=10^4$ independent shuffles. For each permutation $\pi_i(\mathcal{W})$, complete rating recalculation yields $R^{(i)}_m$. The final stabilized rating combines these trials:

\begin{equation}
    \bar{R}_m = \frac{1}{T}\sum_{i=1}^T R^{(i)}_m
\end{equation}

\section{Ablation Study On Pipeline Settings}
\label{sec: ablation study}

\subsection{Impact of History Sampling Configuration}

We conducted supplementary experiments on the \textbf{Qwen2.5-VL-7B} and \textbf{Qwen2.5-VL-72B} models using various history strategies (including increasing the number of history steps and altering sampling methods). 
The results are presented in Tables~\ref{tab:history_strategies} and~\ref{tab:frame_sampling_strategies}. 
These scores were calculated as a percentage of model scores versus human performance in a manner similar to Figure~\ref{fig:model analysis (human)}.  

In the default setting of our main experiments, the history sampling configuration is one where decisions are made using information from the most recent \textbf{\textcolor{blue}{3steps}}, and the game screen is sampled every \textcolor{blue}{3frames}.

\begin{table*}[h!]
\centering
\caption{Performance comparison of different history strategies for Qwen2.5-VL 7B and 72B models.}
\label{tab:history_strategies}
\resizebox{\textwidth}{!}{%
\begin{tabular}{l|ccccc|ccccc}
\hline
& \multicolumn{5}{c|}{\textbf{Qwen2.5-VL-7B}} & \multicolumn{5}{c}{\textbf{Qwen2.5-VL-72B}} \\
\textbf{Game} & \textbf{3steps\_2sample} & \textbf{3steps\_5sample} & \textbf{\textcolor{blue}{3steps}} & \textbf{5steps} & \textbf{8steps} & \textbf{3steps\_2sample} & \textbf{3steps\_5sample} & \textbf{\textcolor{blue}{3steps}} & \textbf{5steps} & \textbf{8steps} \\
\hline
\textbf{race} & 11.20 & 11.20 & 12.60 & 11.20 & 12.40 & 30.00 & 29.00 & 29.60 & 32.60 & 33.60 \\
\textbf{supermario} & 20.10 & 22.10 & 22.60 & 22.80 & 21.20 & 34.50 & 33.90 & 42.10 & 36.40 & 39.80 \\
\textbf{pong} & 3.30 & 4.30 & 3.70 & 4.00 & 4.30 & 4.50 & 3.90 & 4.10 & 4.80 & 5.00 \\
\textbf{flappybird} & 6.70 & 11.20 & 3.40 & 2.10 & 5.00 & 17.70 & 13.10 & 8.10 & 13.30 & 13.60 \\
\textbf{tempestrun} & 18.80 & 17.80 & 21.10 & 18.80 & 17.80 & 22.00 & 21.10 & 24.80 & 22.70 & 23.70 \\
\hline
\textbf{average} & \textbf{12.02} & \textbf{13.32} & \textbf{12.68} & \textbf{11.78} & \textbf{12.14} & \textbf{21.74} & \textbf{20.20} & \textbf{21.74} & \textbf{21.96} & \textbf{23.14} \\
\hline
\end{tabular}%
}
\end{table*}

\begin{table*}[h!]
\centering
\caption{Performance comparison of different frame sampling strategies for Qwen2.5-VL 7B and 72B models.}
\label{tab:frame_sampling_strategies}
\resizebox{\textwidth}{!}{%
\begin{tabular}{l|cccc|cccc}
\hline
& \multicolumn{4}{c|}{\textbf{Qwen2.5-VL-7B}} & \multicolumn{4}{c}{\textbf{Qwen2.5-VL-72B}} \\
\textbf{Game} & \textbf{8frames} & \textbf{5frames} & \textbf{\textcolor{blue}{3frames}} & \textbf{1frames} & \textbf{8frames} & \textbf{5frames} & \textbf{\textcolor{blue}{3frames}} & \textbf{1frames} \\
\hline
\textbf{race} & 13.80 & 11.60 & 12.60 & 10.40 & 19.60 & 27.20 & 29.60 & 26.00 \\
\textbf{pong} & 4.00 & 3.90 & 3.70 & 4.50 & 4.80 & 5.90 & 4.10 & 7.60 \\
\textbf{flappybird} & 3.40 & 5.70 & 3.40 & 7.60 & 10.80 & 14.00 & 8.10 & 14.10 \\
\textbf{tempestrun} & 19.00 & 19.30 & 21.10 & 16.50 & 23.40 & 18.50 & 24.80 & 26.60 \\
\hline
\textbf{average} & \textbf{10.05} & \textbf{10.13} & \textbf{10.20} & \textbf{9.75} & \textbf{14.65} & \textbf{16.40} & \textbf{16.65} & \textbf{18.58} \\
\hline
\end{tabular}%
}
\end{table*}

The experimental results show that simply increasing the length of the history window (e.g., from 3 to 8 steps) does not yield significant performance gains. 
This finding supports our core argument: the bottleneck for current MLLMs lies \textbf{not in the quantity} of historical information they receive, but in their ability to \textbf{understand and utilize} this dynamic visual information. 

Therefore, we chose a 3-frame history as our baseline configuration. 
This provides the necessary temporal context while establishing a fair, simple, and effective standard for exposing the models' core deficiencies, without confounding the evaluation with complex agent strategies.

As mentioned in the main text, to investigate the impact of settings within the sampling strategies on anchoring bias, we also conducted relevant experiments, with the results presented in Appendix \ref{anchoring bias ablation}.

\subsection{Impact of Input Resolution}
To systematically investigate the impact of input resolution on model performance, we conducted a new set of experiments, testing the \textbf{Qwen2.5-VL 7B} and \textbf{72B} models on four different resolutions.

The resolutions from 120 to 480 refer to images with heights of 120 to 480 pixels, respectively, with the width scaled according to the original aspect ratio. We selected these four resolutions to cover different levels of visual detail, from low to high. In the \textcolor{blue}{default} setting of our main experiments, the model's input resolution was \textcolor{blue}{360} pixels height. The scores were calculated as a percentage of model scores versus human performance in a manner similar to Figure 4. The results are presented in Table~\ref{tab:resolution_impact}.

\begin{table*}[h!]
\centering
\caption{Performance comparison of Qwen2.5-VL 7B and 72B models across different input resolutions (height in pixels). Scores are percentages relative to human performance.}
\label{tab:resolution_impact}
\resizebox{\textwidth}{!}{%
\begin{tabular}{l|cccc|cccc}
\hline
& \multicolumn{4}{c|}{\textbf{Qwen2.5-VL 7B}} & \multicolumn{4}{c}{\textbf{Qwen2.5-VL 72B}} \\
\textbf{Game} & \textbf{120 (7B)} & \textbf{240 (7B)} & \textbf{\textcolor{blue}{360} (7B)} & \textbf{480 (7B)} & \textbf{120 (72B)} & \textbf{240 (72B)} & \textbf{\textcolor{blue}{360} (72B)} & \textbf{480 (72B)} \\
\hline
\textbf{race} & 9.80 & 10.60 & 12.60 & 11.00 & 15.80 & 23.60 & 29.60 & 28.20 \\
\textbf{supermario} & 17.50 & 21.70 & 22.60 & 17.20 & 38.90 & 44.90 & 42.10 & 47.90 \\
\textbf{pong} & 4.00 & 3.90 & 3.70 & 3.60 & 3.50 & 3.50 & 4.10 & 3.60 \\
\textbf{flappybird} & 4.40 & 5.40 & 3.40 & 9.80 & 7.70 & 12.70 & 8.10 & 12.00 \\
\textbf{tempestrun} & 19.60 & 19.10 & 21.10 & 18.50 & 19.60 & 24.30 & 24.80 & 22.50 \\
\hline
\textbf{average} & \textit{11.06} & \textit{12.14} & \textbf{12.68} & \textit{12.02} & \textit{17.10} & \textit{21.80} & \textit{21.74} & \textbf{22.84} \\
\hline
\end{tabular}%
}
\end{table*}

This data reveals a nuanced relationship: for the more capable 72B model, the overall performance trend improves with higher resolution, peaking at 480px. This suggests it can benefit from the finer details in higher-resolution images. However, for the smaller 7B model, performance peaks at our default setting of 360px and declines at the higher 480px resolution.

This indicates that the relationship between model performance and input resolution is \textbf{not simply linear}. For less capable models, excessive resolution might introduce `noise' that they struggle to filter effectively, thereby interfering with their decision-making process.

\newpage
\section{Additional Experimental Details}
\label{sec: additional experiments}


\subsection{Unit Tests For Core Visual Abilities Experiment}
\label{unit tests for core visual abilities appendix}

\begin{table*}[ht]
\centering
\caption{Basic visual capabilities and their corresponding simple game levels.}
\label{tab:unit test levels}
\renewcommand{\arraystretch}{1.2}
\begin{tabular}{lll}
\toprule
Visual Abilities & Game & Levels \\
\midrule
Tracking & Pong &  1, 2, 3 \\
Positioning & Race &  1, 1\_no\_history \\
Visual Grounding & TempestRun &   1 \\
Timing & FlappyBird &  1, 2, 3 \\
\bottomrule
\end{tabular}
\end{table*}

The unit testing framework conducts a systematic assessment of fundamental visual capabilities by drawing from the comprehensive V-MAGE benchmark. 
In each carefully designed level of a game, a random baseline score is first determined by averaging scores from random actions. 
Following this, the performance of each evaluated model on the said level is quantified by calculating the percentage of rounds where the model's score outperforms this established random baseline. The specific game levels used for assessing each ability are listed in Table \ref{tab:unit test levels}.

As illustrated in Figure \ref{fig:abilities}, model performances across representative levels for four fundamental visual competencies reveal critical insights: In tracking tasks requiring cross-frame analysis, nearly all models underperform random baselines. This indicates that while current models achieve reasoning through caption-based approaches in single-frame tasks, they struggle to extract discriminative features in multi-frame scenarios requiring fine-grained spatiotemporal comparisons. The quantitative results for each model across the four core visual abilities are presented in Table \ref{tab: appendix unit test performance}.

\begin{table*}[htbp]
\small
    \centering
    \caption{Performance of MLLMs on Core Visual Ability Unit Tests (\% Exceeding Random Baseline)}
    \label{tab: appendix unit test performance} 
    \renewcommand{\arraystretch}{1.5}
    \begin{tabular}{lcccc}
        \toprule
        Model & Positioning & Tracking & Visual Grounding & Timing \\
        \midrule
        Qwen2VL 7B & 0.50 & 0.27 & 0.56 & 0.36 \\
        Qwen2VL 72B & 0.76 & 0.26 & 0.70 & 0.43 \\
        Qwen2.5VL 72B & 0.88 & 0.25 & 0.68 & \textbf{0.51} \\
        InternVL2.5 78B & 0.82 & \textbf{0.33} & 0.66 & 0.49 \\
        InternVL2.5 8B & 0.60 & 0.28 & 0.55 & 0.39 \\
        Gemini-2.0-Flash & 0.68 & 0.32 & \textbf{0.70} & \textbf{0.51} \\
        GPT4o & \textbf{0.98} & 0.29 & 0.66 & 0.58 \\
        \bottomrule
    \end{tabular}
\end{table*}

It is important to interpret the results of these unit tests within their intended scope. Designed to assess fundamental visual competencies, these tests utilize a random baseline score as the primary reference point. While a model significantly outperforming this random baseline indicates a degree of relevant reasoning ability in that specific task dimension, it does not necessarily imply a high level of overall competence. The random baseline represents minimal performance, and even achieving scores far exceeding it on these foundational tests serves primarily to diagnose basic capabilities rather than validate advanced mastery required for complex gameplay.

\subsection{Perceptual Skipping Experiment}
\label{perceptual skipping exp appendix}


\begin{table*}[ht]

\centering
\caption{Model performance on simple levels with and without textual state information.}
\label{tab:perception_skip}
\renewcommand{\arraystretch}{1}
\begin{tabular}{lcccc}
\toprule
\multirow{2}{*}{Model} & \multicolumn{2}{c}{Flappy Bird} & \multicolumn{2}{c}{Pong} \\
\cmidrule(lr){2-3} \cmidrule(lr){4-5}
 & {w/o Text} & {w/ Text} & {w/o Text} & {w/ Text} \\
\midrule
Qwen2.5VL 7B    & 0.8  & 0.35 & 0.19 & 0.25 \\
InternVL2.5 8B  & 0.31 & 0.76 & 0.19 & 0.31 \\
Qwen2.5VL 72B   & 0.35 & 2.17 & 0.21 & 1.19 \\
InternVL2.5 78B & 0.59 & 2.39 & 0.16 & 0.52 \\
GPT4o           & 0.57 & 4.55 & 0.20 & 3.89 \\
Gemini-2.0-Flash& 0.42 & 4.89 & 0.32 & {>10} \\ 
random          & 0.52 &  & 0.18 &  \\ 
human           & \multicolumn{1}{c}{$>10$} & \multicolumn{1}{c}{} & \multicolumn{1}{c}{$>10$} & \multicolumn{1}{c}{} \\
\bottomrule
\end{tabular}
\end{table*}


To further investigate the interplay between visual perception and reasoning, we conducted supplementary experiments where textual descriptions of the game state were provided, effectively bypassing the visual perception module (see Table \ref{tab:perception_skip} for detailed results on Flappy Bird Level 3 and Pong Level 2). 

The results indicate that alleviating the perceptual challenge generally improves performance, particularly for larger models like GPT-4o and the 72B/78B parameter models, supporting the hypothesis that visual perception is a significant bottleneck. However, even with this intervention, model scores remained substantially lower than the human baseline (>10), underscoring the presence of critical reasoning and planning deficiencies beyond visual perception, as discussed earlier.

Notably, the performance gains from text input were more pronounced for larger models, suggesting their enhanced capacity to leverage structured textual information for reasoning, whereas smaller models exhibited less consistent benefits or even performance degradation in some cases. This finding further highlights that while perception is a challenge, fundamental reasoning limitations persist across models and are not fully overcome even when provided with simplified, textual state representations.

\subsection{Action Efficiency as a Fine-Grained Diagnostic}
\label{appendix:action-efficiency}

To better differentiate models that achieve similar success rates on easy or medium-difficulty levels, we introduce \textbf{Action Efficiency}, defined as the average number of decision steps required to reach a successful outcome (lower is better). 
We evaluate four Qwen-family models on Race Levels~1--3, where score ties are common.

\begin{table*}[ht]
\centering
\caption{Action Efficiency on Race Levels~1--3. Lower Action Efficiency is better.}
\label{tab:action_efficiency}
\renewcommand{\arraystretch}{1.1}
\begin{tabular}{lcc}
\toprule
Model & Success Rate (L1 / L2 / L3) & Action Efficiency (L1 / L2 / L3) \\
\midrule
Qwen3-VL-235B-A22B & 85\% / 37\% / 40\% & 4.5 / 8.9 / 8.4 \\
Qwen2.5-VL-72B & 100\% / 27\% / 27\% & 4.6 / 7.9 / 8.1 \\
Qwen2.5-VL-7B & 55\% / 9\% / 7\% & 6.0 / 13.1 / 6.3 \\
Qwen2-VL-7B & 7\% / 0\% / 2\% & 3.0$^{*}$ / -- / 5.0$^{*}$ \\
\midrule
\multicolumn{3}{l}{$^{*}$Low success rates introduce survivorship bias: only the easiest successful seeds are counted.} \\
\bottomrule
\end{tabular}
\end{table*}

We also find that incorporating Action Efficiency into ELO updates preserves the ranking order while increasing separation between efficient and inefficient policies. 
This supports using Action Efficiency as a secondary diagnostic when raw success rates alone are insufficiently discriminative.

\subsection{Anchoring Bias Experiments}
\label{anchoring bias appendix}
\subsubsection{Details and Examples}
\begin{table*}[ht]
\centering
\caption{Average number of rounds for each model to generate different responses.}
\label{tab: appendix avearge rounds for changing responses}
\renewcommand{\arraystretch}{1}
\begin{tabular}{lccccc}
\toprule
\textbf{Model} & \textbf{Race} & \textbf{FlappyBird} & \textbf{Pong} & \textbf{TempestRun} & \textbf{Avg.} \\
\midrule
Qwen2VL 7B & 4.3 & 25.9 & 13.7 & 7.3 & 12.8 \\
Qwen2.5VL 72B & 2.3 & 19.3 & 2.6 & 5.3 & 7.4 \\
InternVL2.5 8B & 2.0 & 6.9 & 6.7 & 8.0 & 5.9 \\
InternVL2.5 78B & 6.8 & 16.0 & 2.0 & 3.0 & 7.0 \\
GPT4o & \textbf{1.0} & \textbf{1.6} & \textbf{1.0} & \textbf{1.0} & \textbf{1.1} \\
\midrule
\textbf{PCC $r$} & \multirow{2}{*}{-0.63} & \multirow{2}{*}{-0.86} & \multirow{2}{*}{-0.88} & \multirow{2}{*}{-0.64} & \multirow{2}{*}{-0.75} \\
(Avg. Rounds vs. ELO) & & & & & \\
\bottomrule
\end{tabular}
\end{table*}

Due to the possibility of models receiving identical visual inputs over multiple rounds in Super Mario (e.g., being stuck in a corner), we conducted a statistical analysis using lots of rounds of responses from each model across the other four games. This was done by iterating through the recorded interactions for each level and measuring the number of sequential rounds where the model's output response remained unchanged. The average of these durations across interactions within a game provides the metric presented in Table \ref{tab: appendix avearge rounds for changing responses}.

The results indicate that GPT-4o updates its responses more actively and frequently when the visuals change, while other models do so less frequently. This may suggest that GPT-4o is more sensitive to subtle visual updates, enabling it to make timely inferences and more accurately track game progress.

\begin{figure*}[htbp]
\vskip 0.2in
\begin{center}
\centerline{\includegraphics[width= \textwidth]{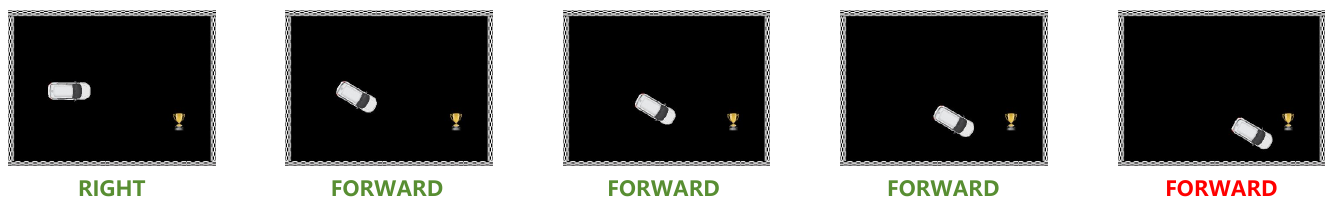}}
\caption{In the prior reasoning processes, GPT-4o accurately identified the car's orientation relative to the trophy's position. However, in the final reasoning instance, due to anchoring bias, the model misinterpreted previous historical information and incorrectly assumed that the car's orientation was already directed toward the trophy. Consequently, despite correctly determining the direction, the model executed an erroneous action.}
\label{fig:anchoring bias}
\end{center}
\vskip -0.2in
\end{figure*}

In the example of Figure \ref{fig:anchoring bias}, GPT-4o's response during the final interaction is as follows:

\noindent
\fbox{
\parbox{\dimexpr\columnwidth-2\fboxsep-2\fboxrule\relax}{%
    Description: The car is currently facing diagonally towards the \textbf{bottom-right} of the screen. The trophy is positioned near the car, slightly to the \textbf{right}. There are no visible obstacles between the car and the trophy, indicating a clear path. The car has been moving consistently forward in the previous steps, maintaining its alignment towards the trophy.
    
    Reasoning: Given the car's orientation towards the trophy and the absence of obstacles, the car's current position and direction remain favorable for reaching the trophy. Continuing to move forward is the most efficient strategy to maintain the direct path to the trophy. The car's current trajectory should allow it to reach the trophy without any adjustments.
    
    Action: FORWARD
    }
}

\subsubsection{Impact of Frame Sampling and Decision Frequency}
\label{anchoring bias ablation}

To further quantify the relationship between perceptual sensitivity and anchoring bias, we have conducted a deeper quantitative exploration of the relationship between anchoring bias and model performance. 
To more objectively measure a model's reaction to dynamic changes in the game world, we introduced a new metric: \textbf{Average Response Game Frames} (abbreviated as \textbf{`avg frames'} in the results). This metric is calculated by:
\text{Avg. Response Frames} = \text{sampling interval} * \text{avg. rounds to generate a different response}, 
and represents \textbf{how many game frames, on average, have elapsed before a model makes a substantive change in its reasoning}.

We performed a series of experiments with different sampling strategies, first testing the Qwen2.5VL-72B model. The results are shown in Table~\ref{tab:sampling_72b}:

\begin{table*}[h!]
\centering
\caption{Performance of Qwen2.5VL-72B under different frame sampling strategies. The top section shows response frequency metrics, while the bottom shows game scores.}
\label{tab:sampling_72b}
\renewcommand{\arraystretch}{1}
\begin{tabular}{l|cccc}
\hline
\textbf{Game} & \textbf{8frames} & \textbf{5frames} & \textbf{3frames(default)} & \textbf{1frames} \\
\hline
\multicolumn{5}{l}{\textit{Average Response Game Frames}} \\
\textbf{race} & 5 & 1.8 & 2.3 & 10.8 \\
\textbf{pong} & 1.7 & 1.7 & 2.6 & 19.7 \\
\textbf{flappybird} & 1.5 & 2.3 & 19.3 & 64.2 \\
\textbf{tempestrun} & 1.8 & 14.2 & 5.3 & 23.2 \\
\textbf{avg request} & 2.5 & 5 & 7.4 & 29.5 \\
\textbf{avg frames} & \textbf{20} & \textbf{25} & \textbf{22.2} & \textbf{29.5} \\
\hline
\multicolumn{5}{l}{\textit{Game Score}} \\
\textbf{race} & 19.60 & 27.20 & 29.60 & 26.00 \\
\textbf{pong} & 4.80 & 5.90 & 4.10 & 7.60 \\
\textbf{flappybird} & 10.80 & 14.00 & 8.10 & 11.90 \\
\textbf{tempestrun} & 23.40 & 18.50 & 24.80 & 26.60 \\
\textbf{avg score} & \textbf{14.65 \textcolor{myred}{$\downarrow$2.00}} & \textbf{16.40 \textcolor{myred}{$\downarrow$0.25}} & \textbf{16.65 \textcolor{mygrey}{$\uparrow$0.00}} & \textbf{18.03 \textcolor{mygreen}{$\uparrow$1.38}} \\
\hline
\end{tabular}%
\end{table*}

\textbf{Stable `visual reaction threshold' in strong models:} From the avg frames metric, the Qwen2.5VL-72B model demonstrates remarkable consistency across different sampling strategies, with its average response time stabilizing within a narrow range of \textbf{20-30} game frames. This suggests that the model possesses a relatively constant intrinsic reaction threshold, where a certain amount of accumulated visual change triggers a shift in its reasoning.

\textbf{Regarding task score (\textbf{avg score}):} The 72B model's performance clearly improves as the sampling interval decreases, with the highest score achieved at the highest decision frequency (1-frame interval). Under such high-frequency decision-making, the model can capture crucial task timings with the highest precision. As the decision frequency decreases, the opportunities for the model to take appropriate action at the right moment are reduced, thus may lead to a drop in performance.

Next, we compared the \textbf{Qwen2.5VL-7B} and \textbf{72B} models under the same sampling strategies. 
The results are presented in Table~\ref{tab:sampling_7b_vs_72b}.

\begin{table*}[h!]
\centering
\caption{Comparison of Qwen2.5VL-7B and 72B models across sampling strategies. The 72B model shows a consistently lower reaction threshold (avg frames) and higher scores.}
\renewcommand{\arraystretch}{1}
\label{tab:sampling_7b_vs_72b}
\begin{tabular}{l|cccc}
\hline
\textbf{Game} & \textbf{8frames} & \textbf{5frames} & \textbf{3frames} & \textbf{1frames} \\
\hline
\multicolumn{5}{l}{\textbf{Qwen2.5VL-7B}} \\
\hline
avg request & 13.5 & 12.5 & 34.7 & 97.8 \\
avg frames & \textbf{108.0} & \textbf{62.5} & \textbf{104.1} & \textbf{97.8} \\
avg score & \textbf{10.1} & \textbf{10.1} & \textbf{10.2} & \textbf{9.8} \\
\hline
\multicolumn{5}{l}{\textbf{Qwen2.5VL-72B}} \\
\hline
avg request & 2.5 & 5 & 7.4 & 29.5 \\
avg frames & \textbf{20.0 \textcolor{mygreen}{$\downarrow$88.0}} & \textbf{25.0 \textcolor{mygreen}{$\downarrow$37.5}} & \textbf{22.2 \textcolor{mygreen}{$\downarrow$81.9}} & \textbf{29.5 \textcolor{mygreen}{$\downarrow$68.3}} \\
avg score & \textbf{14.7 \textcolor{mygreen}{$\uparrow$4.6}} & \textbf{16.4 \textcolor{mygreen}{$\uparrow$6.3}} & \textbf{16.7 \textcolor{mygreen}{$\uparrow$6.5}} & \textbf{18.6 \textcolor{mygreen}{$\uparrow$8.8}} \\
\hline
\end{tabular}%
\end{table*}

The correlation between the "visual reaction threshold" and "task score" remains clear when comparing across models. The 72B model exhibits a \textbf{lower} Average Response Game Frames (indicating higher perceptual sensitivity) and a \textbf{higher} task score, while the 7B model shows the opposite. This is consistent with the conclusions about anchoring bias: a more powerful model possesses greater sensitivity to dynamic visual perception, which forms the basis for more accurate decision-making in interactive tasks.

\subsection{Analysis of Model Errors in V-MAGE}
\label{Apeendix Analysis of GPT4o Errors in V-MAGE}

We have collected \textbf{2,351} prompt-response pairs generated by GPT-4o while completing all levels for 1 to 5 rounds. From these, \textbf{494} examples were randomly and uniformly sampled for manual error annotation. The frequency of occurrence for various error types is shown in Tables~\ref{tab:error_analysis_transposed_gpt4o} and~\ref{tab:error_analysis_transposed_qwen3}, corresponding to GPT-4o and Qwen3-VL-235B-A22B-Instruct, respectively.

\begin{table*}[htbp]
\centering
\caption{Error count by error type and game environment in GPT4o}
\label{tab:error_analysis_transposed_gpt4o}
\renewcommand{\arraystretch}{1.1}

\begin{tabular}{l | ccccc}
\toprule
Error Type & FlappyBird & Pong & Race & SuperMario & TempestRun \\
\midrule
no error & 30 & 18 & 54 & 88 & 21 \\
perception error & 80 & 26 & 26 & 47 & 42 \\
direction error & 2 & 19 & 13 & 16 & 8 \\
recognition error & 1 & 0 & 0 & 0 & 5 \\
perception incomplete & 3 & 0 & 8 & 10 & 10 \\
reasoning error & 24 & 4 & 9 & 10 & 6 \\
history misinterpretation & 21 & 0 & 1 & 6 & 2 \\
action inappropriate & 0 & 0 & 5 & 0 & 0 \\
inconsistency & 0 & 0 & 14 & 1 & 0 \\
instruction following & 0 & 0 & 2 & 0 & 0 \\
\bottomrule
\end{tabular}
\end{table*}

In addition, complementing the analysis of GPT-4o, we also investigated the best-performing open-source model under our evaluation framework, namely Qwen3-VL-235B-A22B-Instruct. The results are detailed as follows.

\begin{table*}[t]
\centering
\caption{Error count by error type and game environment in Qwen3-VL-235B-A22B-Instruct}
\label{tab:error_analysis_transposed_qwen3}
\renewcommand{\arraystretch}{1.1}

\begin{tabular}{l | ccccc}
\toprule
Error Type & FlappyBird & Pong & Race & SuperMario & TempestRun \\
\midrule
no error & 44 & 20 & 52 & 90 & 22 \\
perception error & 35 & 15 & 14 & 25 & 77 \\
direction error & 3 & 18 & 12 & 17 & 9 \\
recognition error & 0 & 0 & 0 & 0 & 0 \\
perception incomplete & 10 & 0 & 6 & 8 & 12 \\
reasoning error & 3 & 5 & 10 & 0 & 5 \\
history misinterpretation & 41 & 1 & 11 & 15 & 3 \\
action inappropriate & 0 & 0 & 6 & 0 & 0 \\
inconsistency & 10 & 5 & 15 & 0 & 0 \\
instruction following & 0 & 0 & 0 & 0 & 0 \\
\bottomrule
\end{tabular}
\end{table*}

\newpage
The definitions of each error type are presented as follows:

\begin{itemize}

    \item \textbf{no error}: There is no error in the response.
    \item \textbf{perception error}: Description misinterpreted elements.
    \item \textbf{direction error}: A type of perception error. Confused directions (e.g., LEFT/RIGHT)
    \item \textbf{recognition error}: A type of perception error. Failed to identify key objects/elements.
    \item \textbf{perception incomplete}: Description missed important elements in the scene.
    \item \textbf{reasoning error}: Flawed logic in the reasoning section for the chosen action.
    \item \textbf{history misinterpretation}: A type of reasoning error. Misunderstood the game history.
    \item \textbf{action inappropriate}: A type of reasoning error. The chosen Action is clearly wrong given the Observation/Reasoning.
    \item \textbf{inconsistency}: Inconsistent action plans in multiple response processes.
    \item \textbf{instruction following}: Failed to follow instructions in the prompt.
    
\end{itemize}

\newpage

\clearpage
\clearpage

\section{Miscellaneous Material}
\label{sec: miscellaneous material}

\subsection{LLM Usage Statement}
\label{LLM usage}
Our research methodology centered on the evaluation of various MLLMs. Models such as GPT-4o served as the subjects within our V-MAGE pipeline, generating the outputs that form the basis of our analysis and conclusions on MLLM performance. The role of these MLLMs was strictly limited to this evaluation phase. The conceptualization and implementation of the V-MAGE framework and its software were carried out entirely by the authors.


For the manuscript preparation, we employed LLMs for the sole purpose of improving grammar and polishing the language. All scientific contributions, including the research ideas, experimental design, and results interpretation, originate exclusively from the authors.





\subsection{Crowdsourcing and Research with Human Subjects}
\label{subsec: appendix crowdsourcing}

Research involving human subjects in this paper was limited to inviting a small number of participants to perform tasks within the V-MAGE game environments for the purpose of establishing a human performance baseline. Participants were provided with the standard game rules and objectives, identical to the instructions given to the evaluated models (see Appendix \ref{subsec: Games and Prompts} for details on setup and prompts). Participation was entirely voluntary, and no compensation was provided. No sensitive or personally identifiable information was collected from participants.

\subsection{Institutional Review Board (IRB) Approvals}
\label{subsec: appendix IRB approvals}

The research involving human subjects in this project has always been conducted under the guidance and supervision of our institution's Institutional Review Board (IRB) and in full compliance with its policies. To formally document this compliance for publication, our research protocol was reviewed by the IRB committee. The committee confirmed the study's classification as 'minimal risk' and has approved our research protocol. 

Furthermore, to address the concern about reproducibility from a practical standpoint, the game tasks we selected have clear objectives and straightforward rules. For human players who understand these rules, their scores are highly stable and approach the theoretical performance ceiling. We are therefore confident that the human baseline can be easily reproduced and verified by other researchers.

\clearpage
\section{Case Study}
\label{sec: Case Study}

\begin{figure*}[b]
\vskip 0.2in
\begin{center}
\centerline{\includegraphics[width= \textwidth]{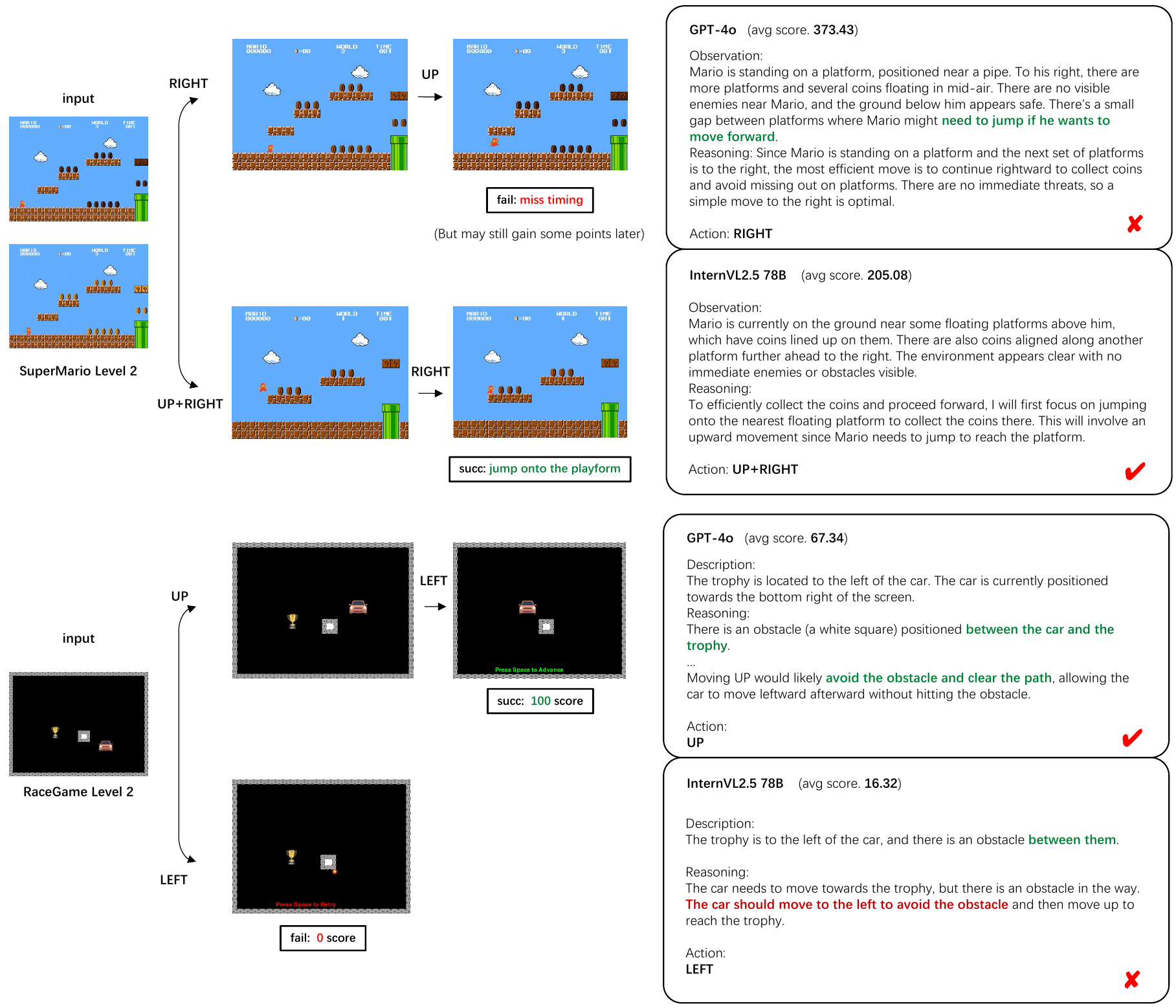}}
\caption{case studies group 1.}
\label{fig:case_studies_group_1}
\end{center}
\vskip -0.2in
\end{figure*}

\begin{figure*}[htbp]
\vskip 0.2in
\begin{center}
\centerline{\includegraphics[width=  \textwidth]{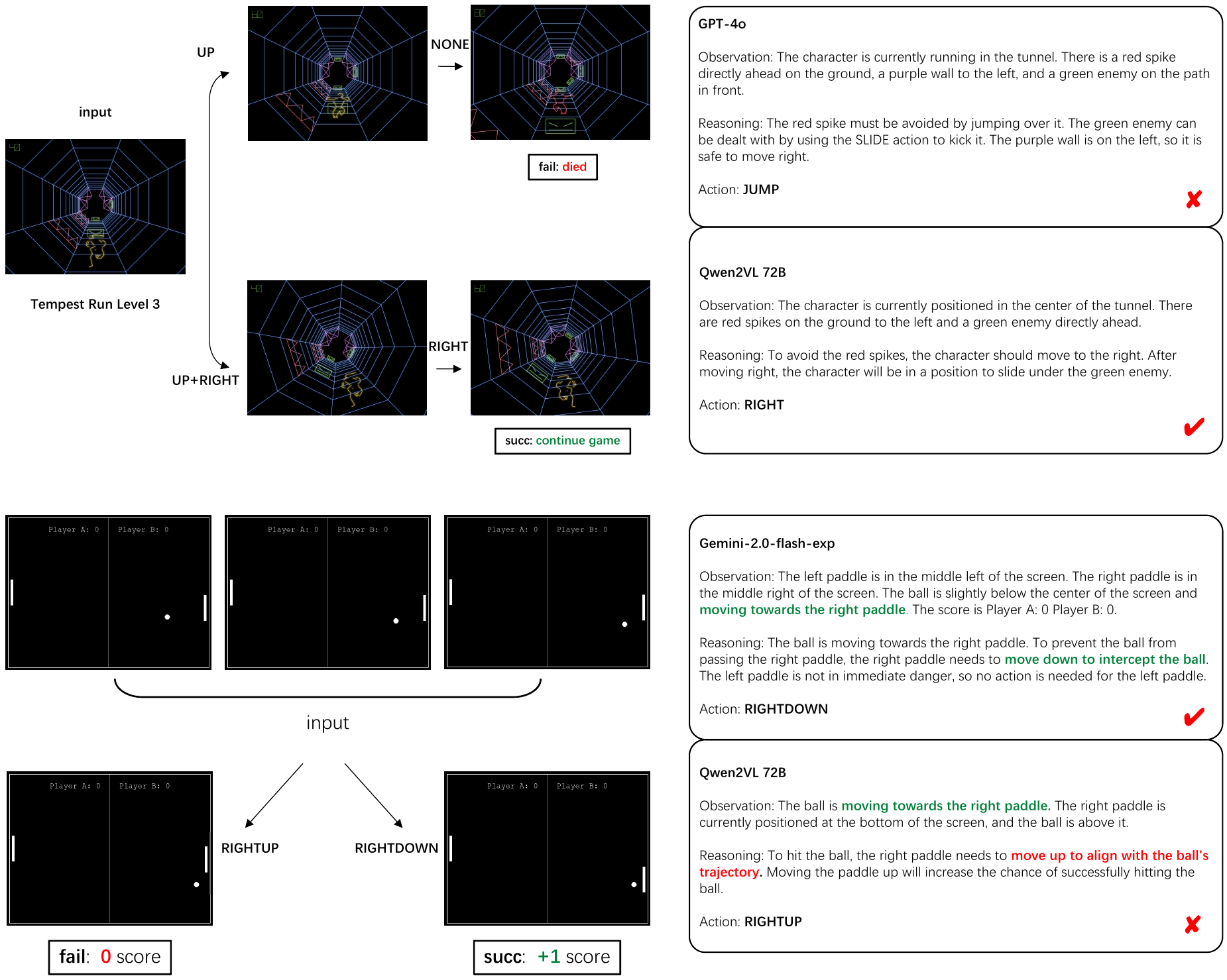}}
\caption{case studies group 2.}
\label{fig:case_studies_group_2}
\end{center}
\vskip -0.2in
\end{figure*}

\begin{figure*}[htbp]
\vskip 0.2in
\begin{center}
\centerline{\includegraphics[width= \textwidth]{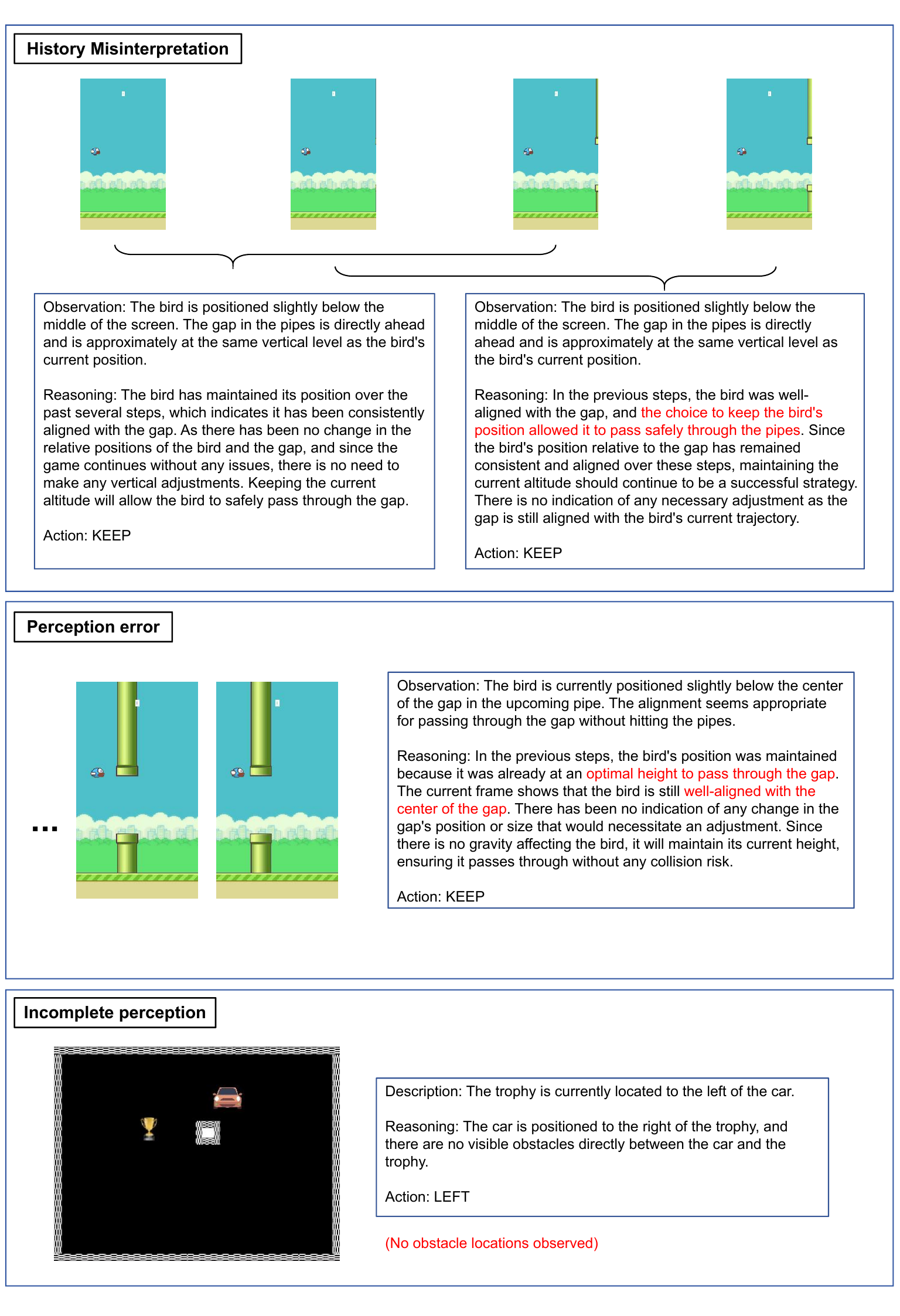}}
\caption{Error analysis in GPT4o cases.}
\label{fig:error_distribution_cases_1}
\end{center}
\vskip -0.2in
\end{figure*}

\begin{figure*}[htbp]
\vskip 0.2in
\begin{center}
\centerline{\includegraphics[width=  \textwidth]{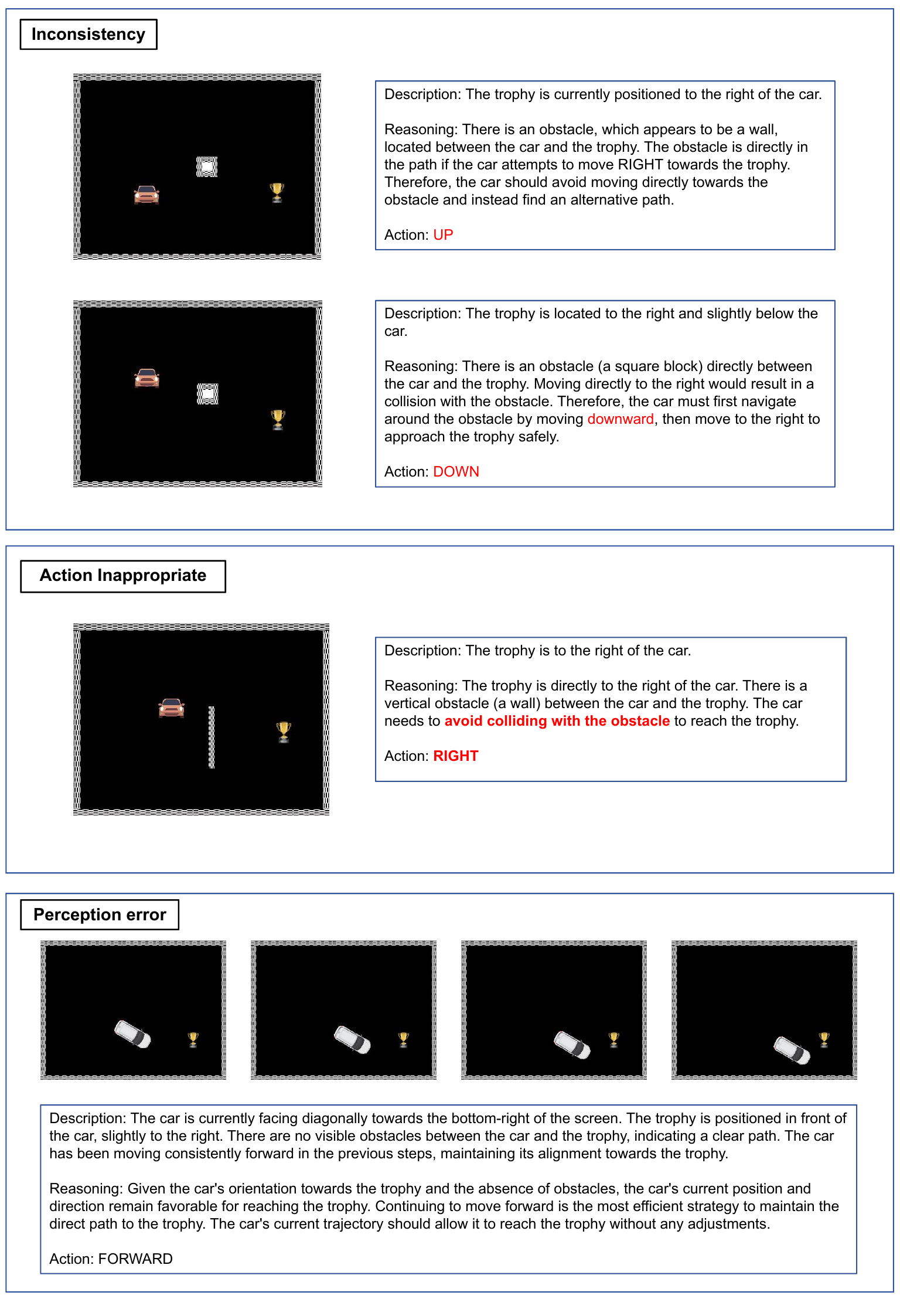}}
\caption{Error analysis in GPT4o cases.}
\label{fig:error_distribution_cases_2}
\end{center}
\vskip -0.2in
\end{figure*}

\begin{figure*}[htbp]
\vskip 0.2in
\begin{center}
\centerline{\includegraphics[width=  \textwidth]{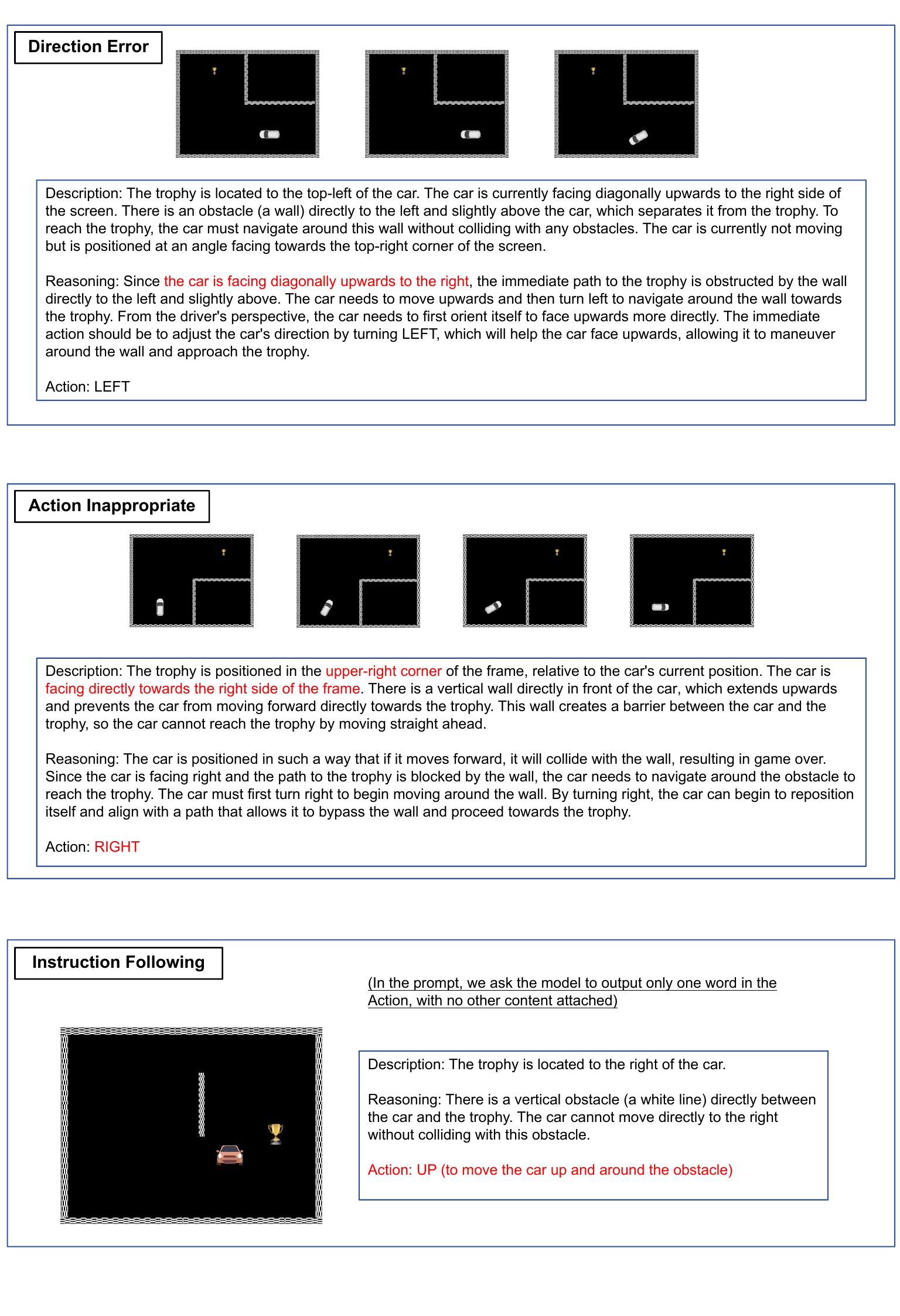}}
\caption{Error analysis in GPT4o cases.}
\label{fig:error_distribution_cases_3}
\end{center}
\vskip -0.2in
\end{figure*}

\begin{figure*}[htbp]
\vskip 0.2in
\begin{center}
\centerline{\includegraphics[width= \textwidth]{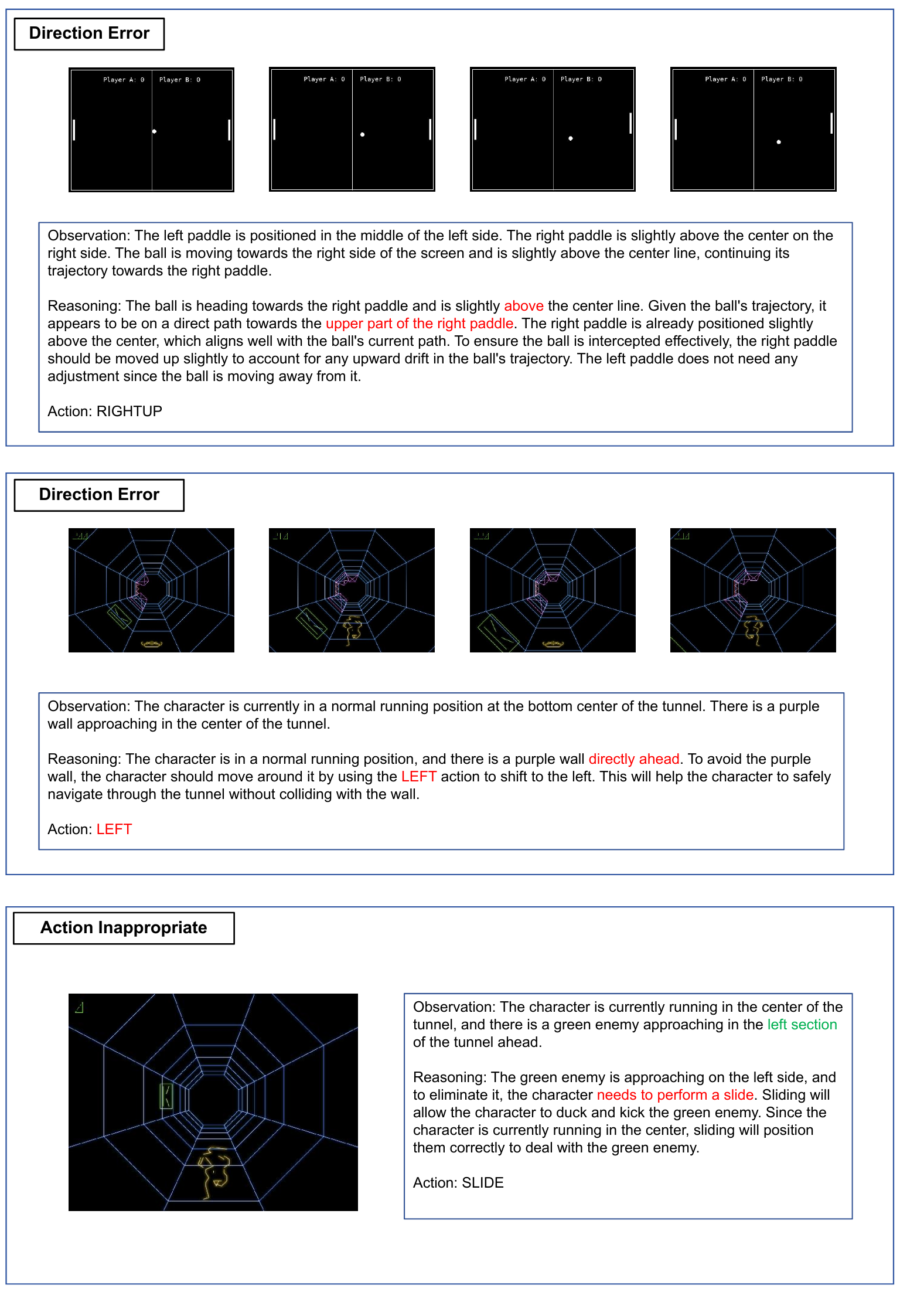}}
\caption{Error analysis in GPT4o cases.}
\label{fig:error_distribution_cases_4}
\end{center}
\vskip -0.2in
\end{figure*}

We present additional case studies and error analyses, as shown in 
Figures~\ref{fig:case_studies_group_1}, 
\ref{fig:case_studies_group_2}, 
\ref{fig:error_distribution_cases_1}, 
\ref{fig:error_distribution_cases_2}, 
\ref{fig:error_distribution_cases_3}, 
and \ref{fig:error_distribution_cases_4}.


\end{document}